\documentclass[]{fairmeta}
\microtypesetup{expansion=false}
\usepackage[utf8]{inputenc}
\usepackage{url,array,makecell,colortbl,amsfonts,amsmath,amssymb}
\usepackage{algorithm,algorithmic,newfloat,listings,paracol,titletoc,enumitem,chngcntr,tabularx}
\usepackage{pifont}
\DeclareRobustCommand{\Envelope}{\text{\ding{41}}}

\DeclareCaptionStyle{ruled}{labelfont=normalfont,labelsep=colon,strut=off}
\floatstyle{ruled}\newfloat{listing}{tb}{lst}{}\floatname{listing}{Listing}
\newtcblisting[auto counter,number within=section]{spatialcliprompt}[2][]{enhanced standard,breakable,height fixed for=first and middle,listing only,colback=blue!5!white,colframe=blue!75!black,title={Box~\thetcbcounter: #2},fonttitle=\bfseries,boxrule=0.8pt,arc=2mm,left=1mm,right=1mm,top=1mm,bottom=1mm,listing options={basicstyle=\footnotesize\ttfamily,breaklines=true,breakatwhitespace=true,breakindent=0pt,breakautoindent=false,columns=fullflexible,keepspaces=true,showstringspaces=false,frame=none,aboveskip=0pt,belowskip=0pt},#1}
\newtcolorbox{casestep}[2]{enhanced standard,breakable,height fixed for=first and middle,colback=#1!5!white,colframe=#1!65!black,title={#2},fonttitle=\bfseries\footnotesize,fontupper=\scriptsize,boxrule=0.7pt,arc=1.5mm,left=1.2mm,right=1.2mm,top=0.8mm,bottom=0.8mm,before skip=3pt,after skip=3pt}
\title{SpatialCLI: Learning to Reason With Spatial Tools, Then Without Them}
\author[1,2,\dagger]{Yang Zhou}
\author[2,3,\dagger]{Zixuan Huang}
\author[2,*]{Sunzhu Li}
\author[4]{Zhuo Yang}
\author[1]{Chen Zhang}
\author[2,6]{Shunian Chen}
\author[1]{Caijun Yan}
\author[2]{Jianyao Xu}
\author[5]{Shunyu Liu}
\author[2]{Weijie Fu}
\author[2]{Peiliang Li}
\author[2]{Xiaozhi Chen}
\author[1,\Envelope]{Yuxiang Cai}
\renewcommand{\affiliationlist}{%
\affiliationformat[1]{Zhejiang University},
\affiliationformat[2]{Zhuoyu Technology, Shenzhen, China},
\affiliationformat[3]{Beihang University}\\
\affiliationformat[4]{University of Electronic Science and Technology of China},
\affiliationformat[5]{Nanyang Technological University}\\
\affiliationformat[6]{The Chinese University of Hong Kong, Shenzhen}%
}
\contribution[\dagger]{Equal contribution}
\contribution[*]{Project lead}
\contribution[\Envelope]{Corresponding author}
\date{2026.7.31}
\metadata[Code]{\url{https://github.com/IANNXANG/SpatialCLI}}
\metadata[Model]{\url{https://huggingface.co/ZYT-MFM/SpatialCLI-8B}}
\metadata[Dataset]{\url{https://huggingface.co/datasets/ZYT-MFM/SpatialCLI-Data}}
\abstract{Vision-language models (VLMs) are increasingly used in embodied agents to interpret visual inputs, reason about spatial relationships, and make task-level decisions based on that reasoning.
However, a fundamental capability mismatch remains: general VLMs can reason about the overall task but often miss the visual details that determine success, while specialist vision models can capture those details but cannot translate them into task-level decisions.
In this work, we propose SpatialCLI, a framework that teaches VLMs to reason with spatial tools and progressively internalize the specialist perceptual capabilities they provide.
SpatialCLI proceeds in three stages: (1) \textbf{\underline{C}all} exposes specialist vision models as spatial tools to augment the VLM's perception; (2) \textbf{\underline{L}earn} uses Cold-Start SFT and agentic RL to improve tool use; and (3) \textbf{\underline{I}nternalize} verbalizes successful tool-use trajectories to internalize specialist perceptual capabilities.
We further introduce SpatialCLI-Bench, a 516-example benchmark for compositional perception across localization, segmentation, depth, and pose.
On MindCube, SpatialCLI raises Qwen3-VL-8B-Instruct from 29.3\% to 84.6\% with tools, surpassing GPT-5.6 Sol with tools (72.1\%), while retaining 73.8\% without tools after internalization.}
\correspondence{\email{imzhouyang@zju.edu.cn}, \email{caiyuxiang@zju.edu.cn}}
\begin{document}
\maketitle

\section{Introduction}

\begin{figure}[t]
    \centering
    \includegraphics[width=1.0\columnwidth]{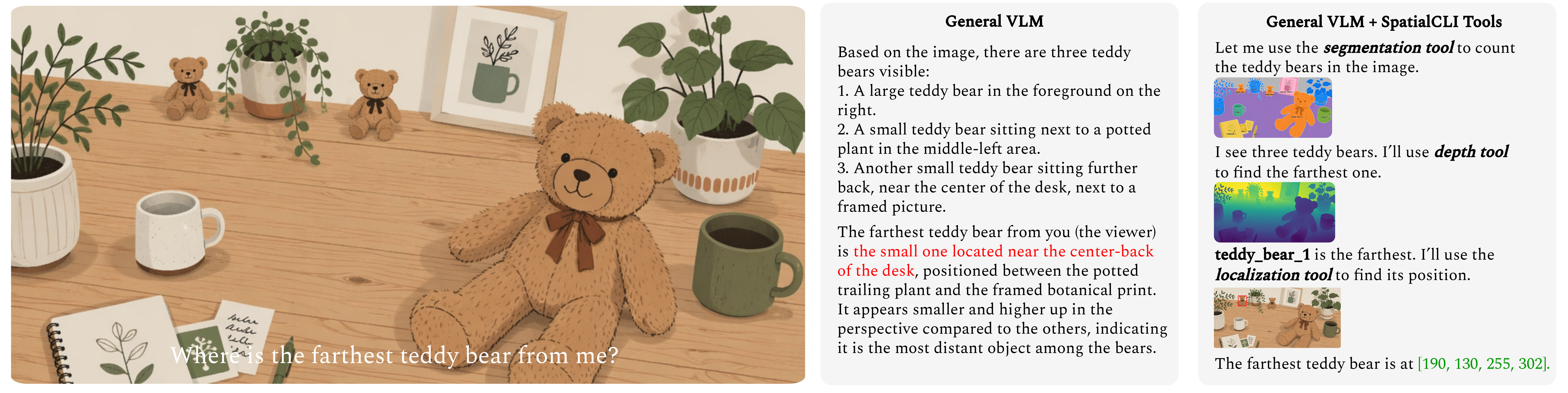}
    \caption{A conceptual comparison of a general VLM with and without SpatialCLI Tools.}
    \label{fig:intro}
\end{figure}

As physical AI moves into open-ended real-world environments, task success increasingly depends on composing multiple spatial perceptual capabilities on demand rather than improving any single capability in isolation \citep{brohan2023rt2,kim2024openvla}.
As shown in Figure~\ref{fig:intro}, finding the farthest teddy bear requires first using segmentation to identify all candidate bears, then comparing their distances using depth information, and finally localizing the selected bear.
Different real-world tasks require different combinations of spatial perceptual capabilities, requiring general physical intelligence to select and coordinate complementary capabilities according to the task objective.

However, general VLMs and specialist vision models exhibit a fundamental capability mismatch under this requirement.
General VLMs can interpret instructions, decompose tasks, and organize multi-step reasoning, but become unreliable when an answer hinges on precise localization, object boundaries, metric depth, or pose.
Specialist vision models such as SAM 3~\citep{sam3}, DA3~\citep{depthanything3}, and VGGT~\citep{wang2025vggt} provide more reliable local perceptual evidence, but each primarily handles a specific perceptual operation and cannot understand which capabilities a task requires, compose outputs from multiple specialists, or translate those outputs into task-level decisions.
Recent systems such as SpaceTools~\citep{chen2025spacetools}, AlloSpatial~\citep{ruan2026allospatial}, and S-Agent~\citep{dai2026s} take important steps by organizing external tools and spatial priors into agentic reasoning processes and training VLMs on the resulting interactions.
An open question is whether the perceptual evidence accumulated during tool interaction can be converted into supervision that enables VLMs to internalize specialist perceptual capabilities and reason without invoking spatial tools.
This would bring VLMs closer to becoming \textit{foundation models native to the physical world}.

In this work, we introduce SpatialCLI, a three-stage framework that enables VLMs to reason with spatial tools and then internalize the specialist perceptual capabilities they provide:
(1) \textit{Call: Inference-Time Tool Augmentation.} The Call stage exposes specialist vision models for localization, segmentation, depth, and pose as spatial tools, augmenting the VLM's perception with fine-grained visual evidence.
(2) \textit{Learn: Agentic Fine-Tuning.} The Learn stage uses Cold-Start SFT to establish basic tool-use behaviors and agentic RL to improve tool planning, selection, and result utilization through task feedback.
(3) \textit{Internalize: Trajectory-Guided Capability Internalization.} SpatialCLI converts successful tool-use trajectories collected from SpatialCLI-RL into evidence-grounded perceptual reasoning chains, and then applies \textit{Dual-View Capability Internalization} to train the final SpatialCLI model to internalize specialist perceptual capabilities while preserving its learned tool-use policy.
Existing benchmarks predominantly evaluate spatial abilities in isolation, leaving it unclear whether models can coordinate multiple complementary capabilities to solve complex spatial problems.
To address this evaluation gap, we construct SpatialCLI-Bench, a 516-example six-choice benchmark for compositional perceptual reasoning across localization, segmentation, depth, and pose.
Our main contributions are summarized as follows:

\begin{itemize}
    \item We propose SpatialCLI, a \textit{Call--Learn--Internalize} framework that equips VLMs with specialist vision models as spatial tools. Through agentic fine-tuning, SpatialCLI learns to coordinate these tools and turns successful tool-use trajectories into evidence-grounded supervision for internalizing specialist perceptual capabilities.
    \item We construct SpatialCLI-Bench, a 516-example benchmark that evaluates compositional perceptual reasoning across localization, segmentation, depth, and pose. The benchmark exposes a substantial limitation of current frontier models: GPT-5.6 Sol achieves only 48.8\%.
    \item Extensive experiments show that SpatialCLI effectively improves both tool-enabled and tool-free reasoning. On SpatialCLI-Bench, SpatialCLI-8B reaches 91.3\% with tools and 72.7\% without tools, with consistent gains on other embodied and spatial benchmarks. These results show that external tool use and internalized direct reasoning can coexist in one model, providing a path toward \textit{foundation models native to the physical world}.
\end{itemize}

\begin{figure*}[t]
    \centering
    \includegraphics[width=1.0\textwidth]{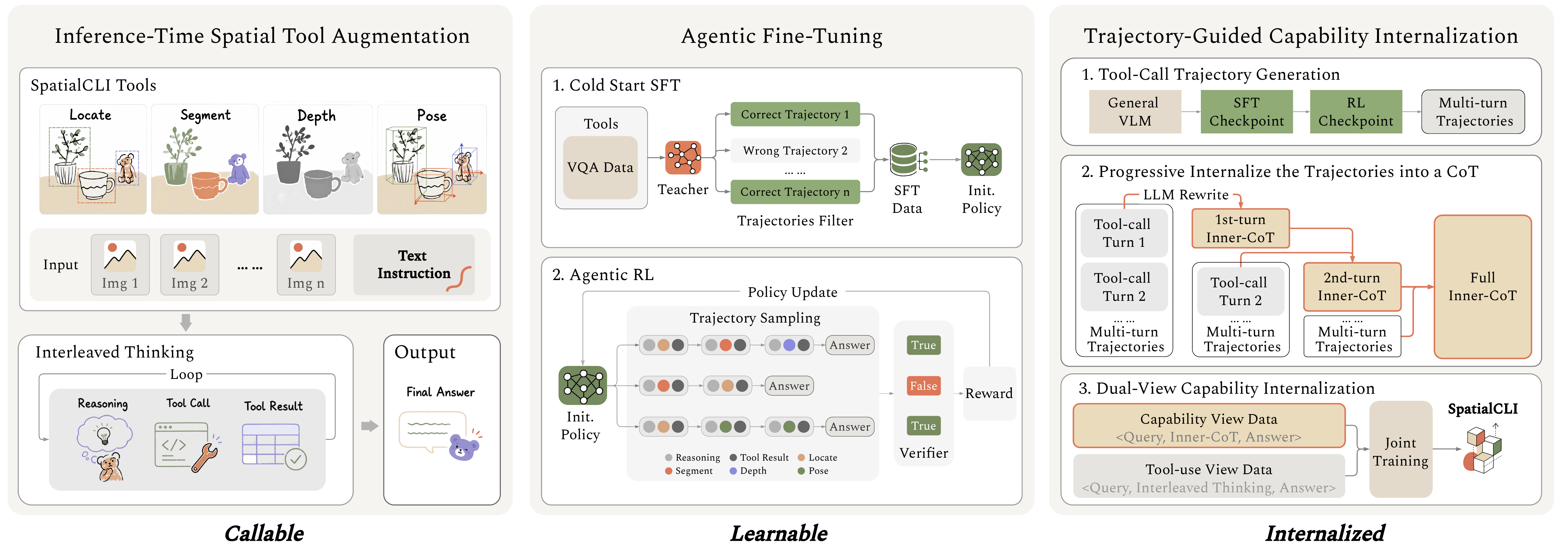}
    \caption{Overview of SpatialCLI. Its \textit{Call}, \textit{Learn}, and \textit{Internalize} stages progressively transform specialist perceptual capabilities into callable spatial tools, learnable tool-use policies, and internalized model capabilities.}
    \label{fig:method}
\end{figure*}

\section{Related Work}

\paragraph{Learning Spatial Capabilities in Vision-Language Models.}
Embodied VLMs and VLAs acquire grounded visual-language knowledge through pretraining and robot demonstrations \citep{brohan2023rt2,kim2024openvla,tencent2026hyembodied05}.
However, current multimodal models still struggle with spatial relations, viewpoint transformation, and 3D understanding \citep{wang2025mindcube,yang2025mmsi,yang2025thinking}.
SpatialVLM, SpatialRGPT, and RoboRefer directly train spatial capabilities with spatial question answering, depth cues, 3D annotations, and task-specific supervision \citep{chen2024spatialvlm,cheng2024spatialrgpt,zhou2025roborefer}.
VLM3 further shows that standard VLM architectures can learn depth, pixel correspondence, camera pose, and object-level 3D understanding from text-based supervision and scaled data mixtures \citep{cai2026vlm3}.
SpatialCLI follows a different route: the VLM invokes external specialist perception tools, then learns direct reasoning from supervision derived from its own successful tool-use trajectories.

\paragraph{Tool-Augmented Agents and Spatial Reasoning.}
Building on advances in LLM reasoning \citep{wei2022cot,deepseekai2025r1,huang2026does}, agents can plan, interact with environments, and invoke tools \citep{yao2023react,schick2023toolformer,shen2023hugginggpt,gupta2023visprog,suris2023vipergpt}, enabling applications in search and information seeking \citep{nakano2021webgpt,jin2025searchr1,liu2025veriweb}, coding \citep{yang2024swe,glm5team2026glm5}, and GUI interaction \citep{qin2025uitars,zheng2024seeact}.
In robotics, hierarchical agents use VLMs as high-level planners and VLAs as low-level executors \citep{hu2026hivla,zhang2026harnessvla,liu2026guava,yang2025agenticrobot,lei2026toolalignedvla,chen2026volo}, but their reliability remains constrained by the spatial perception of the VLM planner.
Spatial reasoning systems use 3D priors or constraints \citep{ma2024spatialpin,chen2025gca}, geometric computation or multi-tool coordination \citep{han2025tiger,chen2025spacetools}, and allocentric representations or spatiotemporal evidence \citep{ruan2026allospatial,dai2026s}.
SpatialCLI not only learns to use multiple specialist perception tools through interaction, but also transforms the VLM's own successful tool-use trajectories into supervision for direct reasoning, enabling reasoning both with external tools and directly without them.

\section{Method}

SpatialCLI aims to turn external specialist perceptual capabilities into abilities that VLMs can \textbf{call, learn, and internalize}.
As shown in Figure~\ref{fig:method}, it proceeds in three stages.
(1) \textit{inference-time tool augmentation} equips VLMs with specialist vision models as spatial tools.
(2) \textit{agentic fine-tuning} fine-tunes the model to learn effective tool-use policies.
(3) \textit{trajectory-guided capability internalization} verbalizes tool-use trajectories to internalize specialist perceptual capabilities while preserving tool-use ability.
The complete end-to-end procedure is summarized in Algorithm~1 of Appendix~A.
In the following, we first introduce the three stages of SpatialCLI and then describe the construction of SpatialCLI-Bench.

\subsection{Inference-Time Tool Augmentation}

Neither VLMs nor specialist vision models can independently solve complex embodied tasks, but they offer complementary strengths.
(1) VLMs excel at semantic understanding and task-level reasoning, but remain unreliable for fine-grained localization, segmentation, depth, and pose perception.
(2) Specialist vision models provide precise perceptual outputs, but lack the task-level understanding needed to determine when they should be invoked or how their outputs should be combined.

To combine these complementary strengths, SpatialCLI builds an inference-time tool-augmented agent framework that equips the VLM with specialist vision models as spatial tools.
Within this framework, the VLM understands the task, selects tools, composes evidence, and completes reasoning, while specialist vision models provide reliable local perceptual evidence.
Following ReAct~\citep{yao2023react}, at each interaction step, the VLM first reasons internally and then either produces a final answer or makes a tool call.
When the VLM makes a tool call, SpatialCLI executes the corresponding spatial tool and returns the tool result to the VLM.
The VLM then continues reasoning with the updated interaction history until it produces a final answer or exhausts the tool-call budget.
Following budget-aware tool-use agents~\citep{liu2025budgetaware}, SpatialCLI exposes the remaining tool-call budget after each tool response, allowing the VLM to decide whether to gather additional evidence or terminate the interaction.
We observe that discarding reasoning content across interaction turns causes redundant reasoning and token inefficiency~\citep{liu2025deepseek}; we therefore retain the VLM's reasoning throughout the entire interaction.

SpatialCLI provides four spatial tools.
(1) \textbf{Locate}, backed by Locate Anything~\citep{wang2026locateanything} and Grounding DINO~\citep{liu2023groundingdino}, takes a language query and returns object bounding boxes.
(2) \textbf{Segment}, backed by SAM 3~\citep{sam3}, takes a language query and returns polygonal object boundaries.
(3) \textbf{Depth}, backed by Depth Anything 3~\citep{depthanything3}, takes queried image points and returns metric depth.
(4) \textbf{Pose}, backed by Orient Anything V2~\citep{wang2026orientanythingv2} and VGGT~\citep{wang2025vggt}, takes an object or camera-motion query and returns object orientation or cross-view camera motion.
The complete interfaces and registrations are provided in Appendix~D.2 and Boxes~D.1--D.4, while the shared agentic tool-use prompt is shown in Box~G.1.
SpatialCLI records the resulting tool calls, returned results, and final answer.
We represent the executed tool-interaction trace and the corresponding complete interaction sample as
\[
\tau = \left((z_t,a_t,o_t)_{t=1}^{T}\right),
\qquad
\xi = (I,q,\tau,y),
\]
where $z_t$ is the reasoning preceding the $t$-th executed tool call, $a_t$ is that tool call, $o_t$ is its returned result, $T$ is the number of executed tool calls, and $y$ is the final answer produced after the trace.

\subsection{Agentic Fine-Tuning}

Within this agent framework, the VLM can interact with spatial tools to produce complete interaction trajectories, but it cannot yet reliably determine when to use which tool, how to specify appropriate arguments, or how to use the returned results to guide subsequent reasoning.
Directly applying multi-tool RL to the initial model requires exploration over a large hybrid action space, where the model is prone to several failure patterns: incorrect tool-call formats, invalid arguments, unnecessary repeated calls, and ignoring tool results when they conflict with its prior beliefs.
SpatialCLI therefore first uses Cold-Start SFT to establish basic tool-interaction behaviors and then applies agentic RL to improve tool-use planning, argument generation, result utilization, and termination through task-level feedback.

\paragraph{Cold-Start SFT.}
Effective RL exploration requires an initial policy capable of selecting appropriate tools and specifying valid arguments.
SpatialCLI therefore uses Qwen3.5-397B-A17B~\citep{qwen2026qwen35} as the teacher model, equips it with the inference-time agent framework, and records the complete multi-turn trajectories it produces while solving the training tasks.
As illustrated by the cases in Appendix~H, these trajectories provide reusable reasoning and common tool-use patterns, such as planning before execution.
We discard samples with invalid tool-call formats, failed tool execution, or incorrect final answers, yielding the SFT dataset $\mathcal{D}_{\mathrm{SFT}}$.
During training, returned tool results serve only as context for subsequent generation, while the loss is computed over model-generated reasoning, tool calls, and final-answer tokens.
This stage transfers these patterns to the initial model and yields the SpatialCLI-SFT checkpoint $\pi_{\mathrm{SFT}}$ for RL.

\paragraph{Agentic RL.}
SFT can only imitate filtered teacher trajectories and cannot use task outcomes to improve tool-use decisions beyond the demonstrations.
Starting from the SpatialCLI-SFT checkpoint $\pi_{\mathrm{SFT}}$, the model follows the same interaction loop as inference-time tool augmentation during rollout, with tool results dynamically appended to the interaction history to guide subsequent decisions.
For each task $x=(I,q,y^*)\sim\mathcal{D}_{\mathrm{RL}}$, GRPO~\citep{shao2024deepseekmath} samples a group of $G$ complete interactions $\{\xi_i\}_{i=1}^{G}$ from the old policy $\pi_{\theta_{\mathrm{old}}}$.
Each interaction receives an outcome reward $R_i=V(y_i,y^*)$ based on its final answer.
Optimizing these rewards yields the SpatialCLI-RL checkpoint $\pi_{\mathrm{RL}}$; the complete objective and training configuration are provided in Appendix~B.1.

\subsection{Trajectory-Guided Capability Internalization}

Moving toward foundation models native to the physical world requires perceptual capabilities to reside in the model itself rather than depend entirely on external tools.
Inspired by learning from visual-program execution traces~\citep{hu2023vpd}, SpatialCLI therefore uses successful SpatialCLI-RL trajectories as the source of supervision for internalizing the specialist perceptual capabilities supplied by spatial tools while preserving the learned tool-use policy.
This process consists of two steps: (1) \textit{Progressive Evidence-Grounded Trajectory Verbalization} converts successful tool-use trajectories into explicit perceptual reasoning chains, and (2) \textit{Dual-View Capability Internalization} jointly trains on capability-internalization and tool-use views to internalize specialist perceptual capabilities without sacrificing tool use.

\paragraph{Progressive Evidence-Grounded Trajectory Verbalization.}
Raw SpatialCLI-RL trajectories contain lengthy model reasoning, tool-call syntax, heterogeneous tool returns, and repeated interaction history, making them unsuitable as direct natural-language supervision.
Verbalizing an entire trajectory in a single pass requires processing a long context and can obscure dependencies between evidence collected across turns.
SpatialCLI therefore processes each successful trajectory ($y=y^*$) in two successive steps.
\textbf{(1) Turn-wise evidence consolidation.}
To preserve cross-turn dependencies without repeatedly processing the full raw history, SpatialCLI consolidates the newly collected evidence after each tool interaction.
Let $e_t$ denote the resulting evidence--reasoning unit at turn $t$, and let $e_{<t}=(e_1,\ldots,e_{t-1})$, with $e_{<1}=\varnothing$.
The extractor combines the visual input and task instruction with the previously consolidated units and the current interaction:
\[
e_t
=
\Psi_{\mathrm{ext}}
\left(I,q,e_{<t},z_t,a_t,o_t\right),
\qquad t=1,\ldots,T.
\]
Each $e_t$ exhaustively verbalizes the current tool result, preserves explicit visual observations from $z_t$, and records how they update the accumulated evidence without repeating unchanged content.
Each perceptual statement must be traceable to the current tool result, an explicit visual observation in $z_t$, or a previous unit; the correct answer $y^*$ is withheld to prevent answer-conditioned evidence reconstruction.
\textbf{(2) Global trajectory verbalization.}
To integrate evidence distributed across turns into a coherent task-level reasoning chain, SpatialCLI passes the visual input, task instruction, consolidated evidence trajectory $E_\tau=(e_1,\ldots,e_T)$, and correct answer to a global verbalizer:
\[
c
=
\Phi_{\mathrm{verb}}
\left(I,q,E_\tau,y^*\right).
\]
The verbalizer removes redundancy and organizes the dependencies from perceptual observations to the correct answer into a concise reasoning chain $c$, without introducing entities, attributes, values, or relations absent from $E_\tau$.
The complete turn-wise consolidation and global verbalization prompts are provided in Appendix~G, Boxes~G.2 and~G.3.

\paragraph{Dual-View Capability Internalization.}
To internalize specialist perceptual capabilities while preserving the model's learned tool-use policy, SpatialCLI constructs two training views from each successful trajectory.
(1) The \textbf{Capability-Internalization View} uses a direct-answer prompt and forms each training sample from the image, task instruction, explicit perceptual reasoning chain, and final answer.
The model learns to generate the reasoning chain and answer without accessing external tools, thereby internalizing the specialist perceptual capabilities originally supplied by spatial tools.
(2) The \textbf{Tool-Use View} uses a prompt containing tool descriptions and preserves the original interaction structure, with returned tool results serving as context for subsequent generation.
Training supervises model-generated reasoning, tool calls, and final-answer tokens, preserving the model's ability to decide when to invoke tools and how to use returned results.
The two views are jointly optimized as:
\[
\mathcal{L}_{\mathrm{CI}}
=
\mathcal{L}_{\mathrm{internal}}
+
\lambda \mathcal{L}_{\mathrm{agentic}}.
\]
Here, $\mathcal{L}_{\mathrm{internal}}$ supervises tool-free explicit perceptual reasoning and the final answer, while $\mathcal{L}_{\mathrm{agentic}}$ supervises model-generated tokens in the tool-interaction trajectory.
Both terms are mean negative log-likelihoods over their respective supervised tokens, and $\lambda$ controls the relative weight of the Tool-Use View.

\subsection{SpatialCLI-Bench Construction}

Existing benchmarks predominantly evaluate spatial capabilities in isolation, making it difficult to assess whether models can coordinate multiple perceptual capabilities.
We therefore construct SpatialCLI-Bench, a 516-example English six-choice visual question answering benchmark for compositional reasoning across localization, segmentation, depth, and pose.

\begin{table}[t]
\centering
\small
\renewcommand{\arraystretch}{1.0}
\setlength{\tabcolsep}{0.9mm}
\newcommand{\scoregain}[3]{#1\hspace{0.15em}%
\ifx#2\uparrow\textcolor{green!50!black}{(\ensuremath{\mathord{#2}}#3)}%
\else\textcolor{red!70!black}{(\ensuremath{\mathord{#2}}#3)}%
\fi}
\resizebox{\textwidth}{!}{%
\begin{tabular}{lllllllll}
\toprule
\multicolumn{1}{c}{\multirow{2.5}{*}{\textbf{Model}}} & \multicolumn{1}{c}{\multirow{2.5}{*}{\shortstack{\textbf{SpatialCLI}\\\textbf{Bench}}}} & \multicolumn{1}{c}{\multirow{2.5}{*}{\textbf{MindCube}}} & \multicolumn{2}{c}{\textbf{MMSI}} & \multicolumn{1}{c}{\multirow{2.5}{*}{\textbf{DA-2K}}} & \multicolumn{2}{c}{\textbf{BOPASK}} & \multicolumn{1}{c}{\multirow{2.5}{*}{\textbf{Avg.}}} \\
\cmidrule(lr){4-5} \cmidrule(lr){7-8}
& & & \textbf{Motion-Cam} & \textbf{Pos-Cam-Cam} & & \textbf{Traj.} & \textbf{ObjRrr} & \\
\midrule
\multicolumn{9}{c}{\textbf{Frontier Models}} \\
\midrule
GPT-5.6 Sol & 48.8 & 70.3 & 52.7 & \textbf{62.4} & 79.2 & 51.5 & \textbf{56.4} & 62.0 \\
\hspace*{1em}+ SpatialCLI Tools & \scoregain{72.9}{\uparrow}{24.1} & \scoregain{72.1}{\uparrow}{1.8} & \scoregain{54.1}{\uparrow}{1.4} & \scoregain{62.4}{\uparrow}{0.0} & \scoregain{86.1}{\uparrow}{6.9} & \scoregain{52.8}{\uparrow}{1.3} & \scoregain{49.5}{\downarrow}{6.9} & \scoregain{68.1}{\uparrow}{6.1} \\
Gemini 3.1 Pro & 52.9 & 74.2 & \textbf{59.5} & 51.6 & 67.7 & 48.6 & 14.2 & 56.4 \\
Qwen3.7-Plus & 43.8 & 62.8 & 43.2 & 45.2 & 76.3 & \textbf{61.3} & 42.4 & 55.8 \\
Qwen3.5-397B-A17B & 40.9 & 49.3 & 43.2 & 46.2 & 70.3 & 56.4 & 31.0 & 49.8 \\
\hspace*{1em}+ SpatialCLI Tools & \scoregain{79.8}{\uparrow}{39.0} & \scoregain{67.6}{\uparrow}{18.3} & \scoregain{46.0}{\uparrow}{2.8} & \scoregain{54.8}{\uparrow}{8.6} & \scoregain{91.9}{\uparrow}{21.6} & \scoregain{57.7}{\uparrow}{1.3} & \scoregain{44.7}{\uparrow}{13.7} & \scoregain{68.2}{\uparrow}{18.4} \\
\midrule
\multicolumn{9}{c}{\textbf{Our Models}} \\
\midrule
Qwen3.6-27B & 46.3 & 56.6 & 40.5 & 40.9 & 75.0 & 50.9 & 36.8 & 52.5 \\
\hspace*{1em}+ SpatialCLI Tools & \scoregain{82.2}{\uparrow}{35.9} & \scoregain{62.4}{\uparrow}{5.8} & \scoregain{51.4}{\uparrow}{10.9} & \scoregain{46.2}{\uparrow}{5.3} & \scoregain{91.5}{\uparrow}{16.5} & \scoregain{60.6}{\uparrow}{9.7} & \scoregain{50.6}{\uparrow}{13.8} & \scoregain{68.1}{\uparrow}{15.6} \\
SpatialCLI-27B & \scoregain{76.3}{\uparrow}{30.0} & \scoregain{80.4}{\uparrow}{23.8} & \scoregain{41.9}{\uparrow}{1.4} & \scoregain{43.0}{\uparrow}{2.1} & \scoregain{85.5}{\uparrow}{10.5} & \scoregain{57.3}{\uparrow}{6.4} & \scoregain{53.1}{\uparrow}{16.3} & \scoregain{68.0}{\uparrow}{15.5} \\
\rowcolor{blue!6}
\hspace*{1em}+ SpatialCLI Tools & \scoregain{91.7}{\uparrow}{45.4} & \scoregain{85.5}{\uparrow}{28.9} & \scoregain{52.7}{\uparrow}{12.2} & \scoregain{53.8}{\uparrow}{12.9} & \scoregain{91.9}{\uparrow}{16.9} & \scoregain{58.2}{\uparrow}{7.3} & \scoregain{55.6}{\uparrow}{18.8} & \scoregain{\textbf{75.9}}{\uparrow}{23.4} \\
\midrule
Qwen3.6-35B-A3B & 44.4 & 54.2 & 36.5 & 44.1 & 71.9 & 52.3 & 39.2 & 51.3 \\
\hspace*{1em}+ SpatialCLI Tools & \scoregain{80.2}{\uparrow}{35.8} & \scoregain{63.3}{\uparrow}{9.1} & \scoregain{48.7}{\uparrow}{12.2} & \scoregain{45.2}{\uparrow}{1.1} & \scoregain{91.3}{\uparrow}{19.4} & \scoregain{58.5}{\uparrow}{6.2} & \scoregain{43.5}{\uparrow}{4.3} & \scoregain{66.6}{\uparrow}{15.2} \\
SpatialCLI-35B-A3B & \scoregain{75.0}{\uparrow}{30.6} & \scoregain{78.8}{\uparrow}{24.6} & \scoregain{43.2}{\uparrow}{6.7} & \scoregain{46.2}{\uparrow}{2.1} & \scoregain{83.9}{\uparrow}{12.0} & \scoregain{57.6}{\uparrow}{5.3} & \scoregain{54.7}{\uparrow}{15.5} & \scoregain{67.7}{\uparrow}{16.4} \\
\rowcolor{blue!6}
\hspace*{1em}+ SpatialCLI Tools & \scoregain{\textbf{91.9}}{\uparrow}{47.5} & \scoregain{\textbf{85.6}}{\uparrow}{31.4} & \scoregain{50.0}{\uparrow}{13.5} & \scoregain{53.8}{\uparrow}{9.7} & \scoregain{91.8}{\uparrow}{19.9} & \scoregain{57.9}{\uparrow}{5.6} & \scoregain{53.7}{\uparrow}{14.5} & \scoregain{75.4}{\uparrow}{24.1} \\
\midrule
Qwen3-VL-8B-Instruct & 35.3 & 29.3 & 27.0 & 25.8 & 68.1 & 25.8 & 13.3 & 35.7 \\
\hspace*{1em}+ AlloSpatial Tools & \scoregain{43.2}{\uparrow}{7.9} & \scoregain{35.2}{\uparrow}{5.9} & \scoregain{24.3}{\downarrow}{2.7} & \scoregain{35.5}{\uparrow}{9.7} & \scoregain{47.4}{\downarrow}{20.7} & \scoregain{25.9}{\uparrow}{0.1} & \scoregain{9.7}{\downarrow}{3.6} & \scoregain{34.7}{\downarrow}{1.0} \\
\hspace*{1em}+ SpaceTools Tools & \scoregain{39.3}{\uparrow}{4.0} & \scoregain{30.5}{\uparrow}{1.2} & \scoregain{21.6}{\downarrow}{5.4} & \scoregain{20.4}{\downarrow}{5.4} & \scoregain{\textbf{95.9}}{\uparrow}{27.8} & \scoregain{44.0}{\uparrow}{18.2} & \scoregain{22.3}{\uparrow}{9.0} & \scoregain{44.0}{\uparrow}{8.2} \\
\hspace*{1em}+ SpatialCLI Tools & \scoregain{66.5}{\uparrow}{31.2} & \scoregain{47.2}{\uparrow}{17.9} & \scoregain{41.9}{\uparrow}{14.9} & \scoregain{39.8}{\uparrow}{14.0} & \scoregain{91.6}{\uparrow}{23.5} & \scoregain{54.3}{\uparrow}{28.5} & \scoregain{20.6}{\uparrow}{7.3} & \scoregain{56.7}{\uparrow}{21.0} \\
SpatialCLI-8B & \scoregain{72.7}{\uparrow}{37.4} & \scoregain{73.8}{\uparrow}{44.5} & \scoregain{35.1}{\uparrow}{8.1} & \scoregain{32.3}{\uparrow}{6.5} & \scoregain{80.8}{\uparrow}{12.7} & \scoregain{54.6}{\uparrow}{28.8} & \scoregain{52.3}{\uparrow}{39.0} & \scoregain{62.9}{\uparrow}{27.2} \\
\rowcolor{blue!6}
\hspace*{1em}+ SpatialCLI Tools & \scoregain{91.3}{\uparrow}{56.0} & \scoregain{84.6}{\uparrow}{55.3} & \scoregain{39.2}{\uparrow}{12.2} & \scoregain{53.8}{\uparrow}{28.0} & \scoregain{91.8}{\uparrow}{23.7} & \scoregain{55.2}{\uparrow}{29.4} & \scoregain{48.5}{\uparrow}{35.2} & \scoregain{73.2}{\uparrow}{37.5} \\
\bottomrule
\end{tabular}%
}
\caption{Overall performance on embodied and spatial benchmarks. Scores in model-name rows are obtained under inference w/o Tools, while $+$ rows are obtained under inference w/ the indicated tools. Arrows denote score changes from the corresponding base model w/o Tools. Avg. macro-averages benchmarks after first averaging subsets within each benchmark. The best result in each column is highlighted in \textbf{bold}.}
\label{tab:main_results}
\end{table}

The construction follows four stages: (1) Gemini 3.1 Pro~\citep{google2026gemini31pro} inventories reliable entities and scene relations; (2) specialist vision models provide localization, segmentation, metric-depth, and pose evidence, which is used to filter candidates with missing, inconsistent, or ambiguous evidence; (3) conditioned on the verified evidence, Gemini 3.1 Pro generates the question, correct answer, and five plausible distractors; and (4) human experts independently answer the questions without seeing Gemini's answers, and only examples with matching answers are retained.
Further details are provided in Appendix~C.

\section{Experiments}

\subsection{Experimental Setup}

\paragraph{Models and Training Settings.}
We instantiate SpatialCLI with three base VLMs: Qwen3-VL-8B-Instruct~\citep{bai2025qwen3vl}, Qwen3.6-35B-A3B~\citep{qwen2026qwen35}, and Qwen3.6-27B~\citep{qwen2026qwen35}.
All SFT and RL experiments are conducted using verl~\citep{sheng2024hybridflow}.
Detailed training configurations are provided in Appendix~B.1.

\paragraph{Evaluation Benchmarks.}
We evaluate SpatialCLI on SpatialCLI-Bench, MindCube~\citep{wang2025mindcube}, MMSI~\citep{yang2025mmsi}, DA-2K~\citep{yang2024depthanythingv2}, and BOPASK~\citep{bhat2026bopask}.
We evaluate each model both w/o Tools and w/ Tools.
Detailed evaluation settings, including decoding parameters, tool-call budgets, and repeated evaluations, are provided in Appendix~B.2.

\paragraph{Baselines.}
We compare SpatialCLI with general-purpose models, including GPT-5.6 Sol~\citep{openai2026gpt56sol}, Gemini 3.1 Pro~\citep{google2026gemini31pro}, Qwen3.7-Plus~\citep{qwen2026qwen37}, and Qwen3.5-397B-A17B~\citep{qwen2026qwen35}.
We further compare with SpaceTools~\citep{chen2025spacetools} and AlloSpatial~\citep{ruan2026allospatial}, two agentic methods for improving model spatial capabilities.

\subsection{Overall Performance}

\paragraph{SpatialCLI Tools provide effective runtime spatial augmentation.}
Table~\ref{tab:main_results} shows that SpatialCLI Tools provide broad gains across both frontier and smaller models.
The gains are generally larger for less capable base models: averaged over benchmarks, SpatialCLI Tools improve GPT-5.6 Sol by 6.1 points, compared with 21.0 points for Qwen3-VL-8B-Instruct.
This suggests that external specialist perception is particularly valuable when the base model's native spatial capabilities are limited.
Under the controlled Qwen3-VL-8B-Instruct comparison, SpatialCLI Tools outperform the other tool frameworks on most subsets and AlloSpatial on every subset, showing that effective tool interfaces and tool-use policies are critical for translating specialist perception into task performance.

\paragraph{Training jointly improves direct answering and practical tool use.}
After training, every SpatialCLI variant outperforms its corresponding initial model w/o Tools on every reported evaluation subset, demonstrating consistent capability internalization.
Training also strengthens practical tool use: compared with applying SpatialCLI Tools directly to the corresponding initial models, the trained variants achieve higher w/ Tools performance on most evaluation subsets.
For example, SpatialCLI-8B improves from 35.3 to 72.7 on SpatialCLI-Bench w/o Tools and further reaches 91.3 w/ Tools.
These results indicate that capability internalization need not trade off against tool use: training strengthens native spatial reasoning while preserving, and often further improving, the model's ability to benefit from external tools.

\begin{figure}[t]
    \centering
    \includegraphics[width=0.98\textwidth]{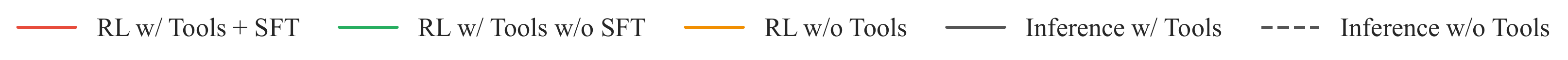}
    \begin{subfigure}[t]{0.23\textwidth}
        \centering
        \includegraphics[width=\textwidth]{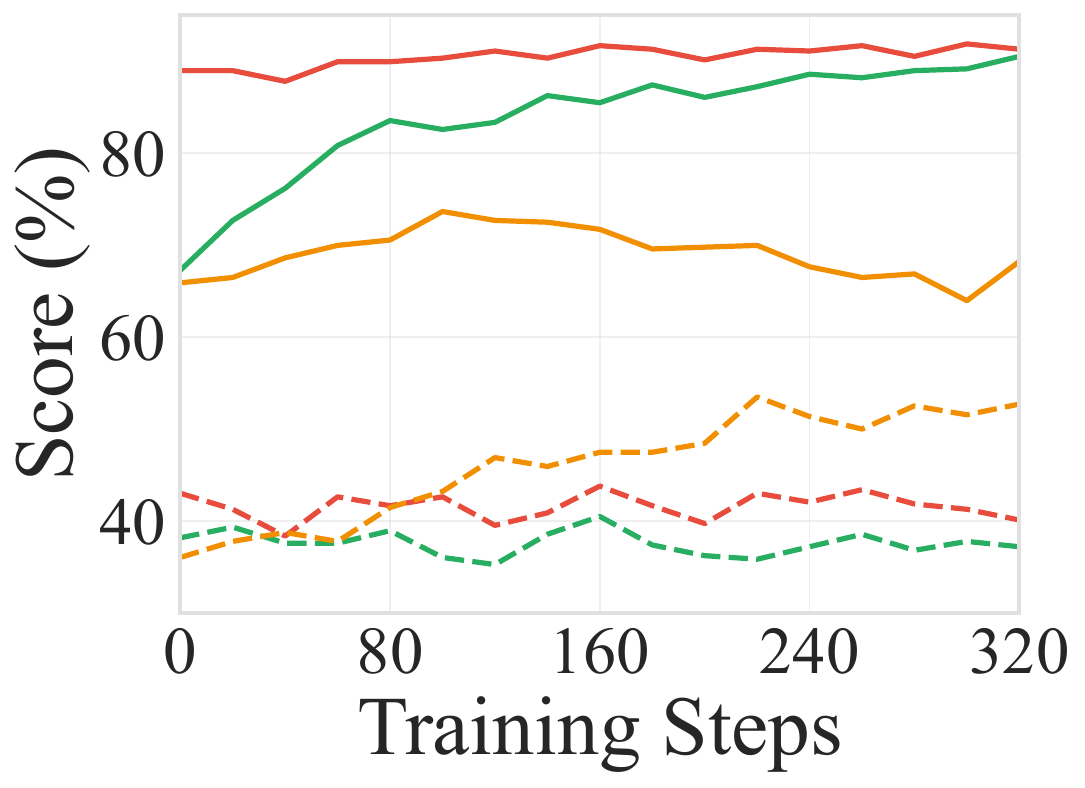}
        \caption{SpatialCLI-Bench Score}
        \label{fig:spatialcli_bench}
    \end{subfigure}
    \hfill
    \begin{subfigure}[t]{0.23\textwidth}
        \centering
        \includegraphics[width=\textwidth]{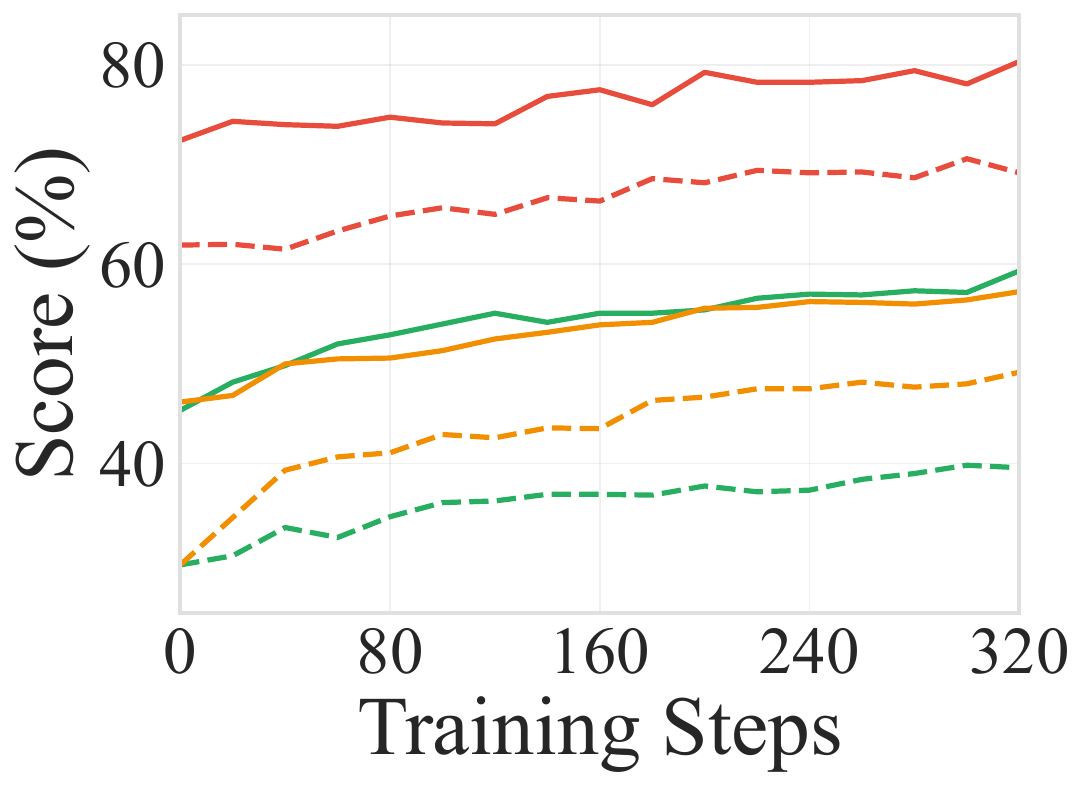}
        \caption{MindCube Score}
        \label{fig:mindcube}
    \end{subfigure}
    \hfill
    \begin{subfigure}[t]{0.23\textwidth}
        \centering
        \includegraphics[width=\textwidth]{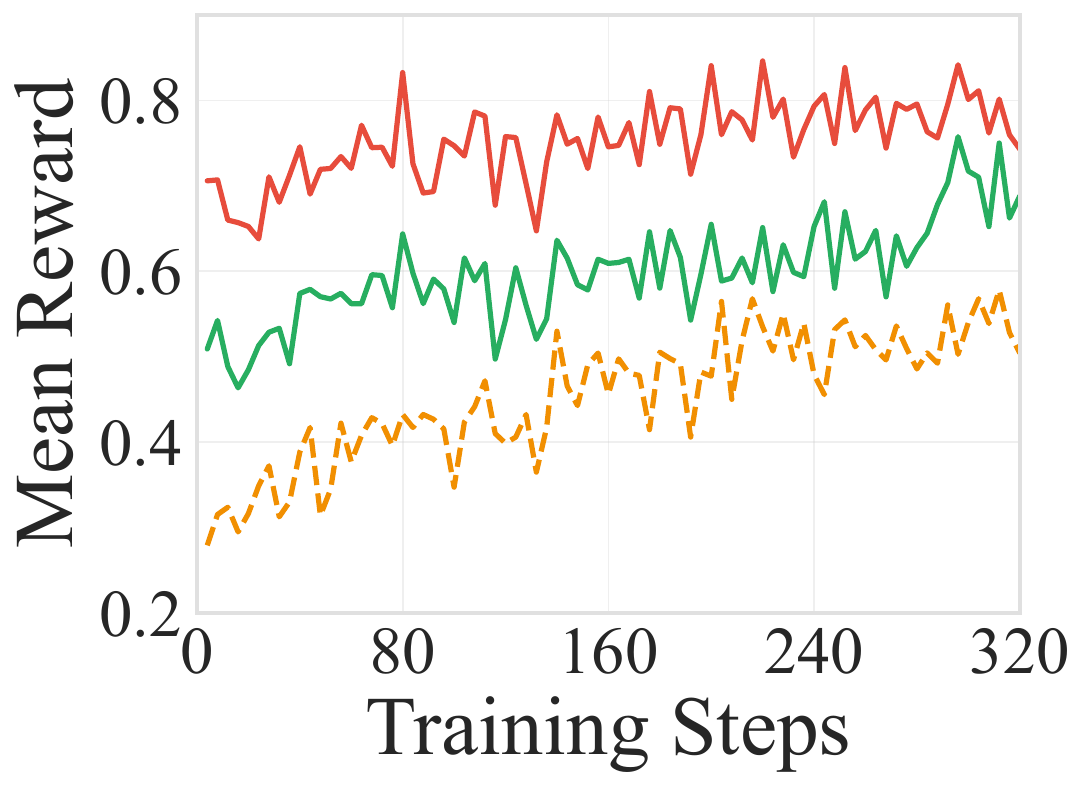}
        \caption{Training Reward}
        \label{fig:training_reward}
    \end{subfigure}
    \hfill
    \begin{subfigure}[t]{0.23\textwidth}
        \centering
        \includegraphics[width=\textwidth]{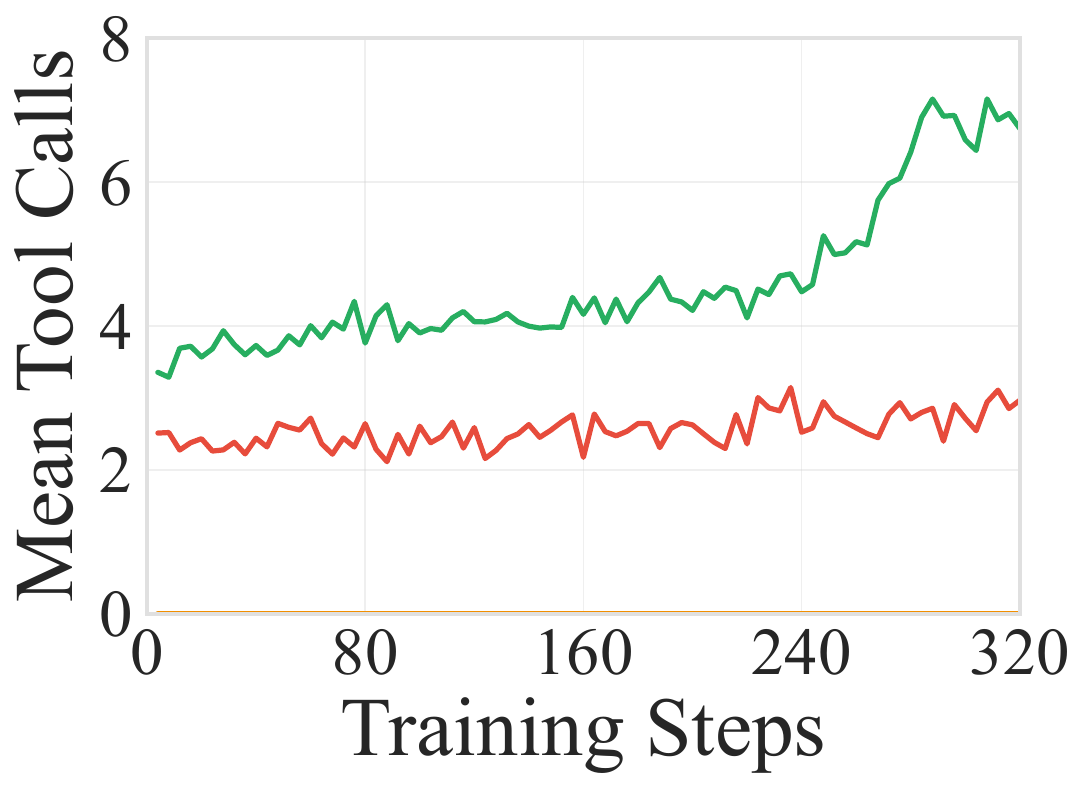}
        \caption{Num. of Tool Calls}
        \label{fig:training_tool_calls}
    \end{subfigure}
    \caption{Training dynamics of Qwen3-VL-8B-Instruct under different RL strategies. (a,b) Inference w/ Tools (solid) and Inference w/o Tools (dashed) scores on SpatialCLI-Bench and MindCube. (c) Training reward over RL training. (d) Number of executed tool calls per trajectory over RL training. We compare RL w/ Tools + SFT, RL w/ Tools w/o SFT, and RL w/o Tools.}
    \label{fig:training_dynamics}
\end{figure}

\begin{figure}[t]
    \centering
    \begin{subfigure}[t]{0.235\textwidth}
        \centering
        \includegraphics[width=\textwidth]{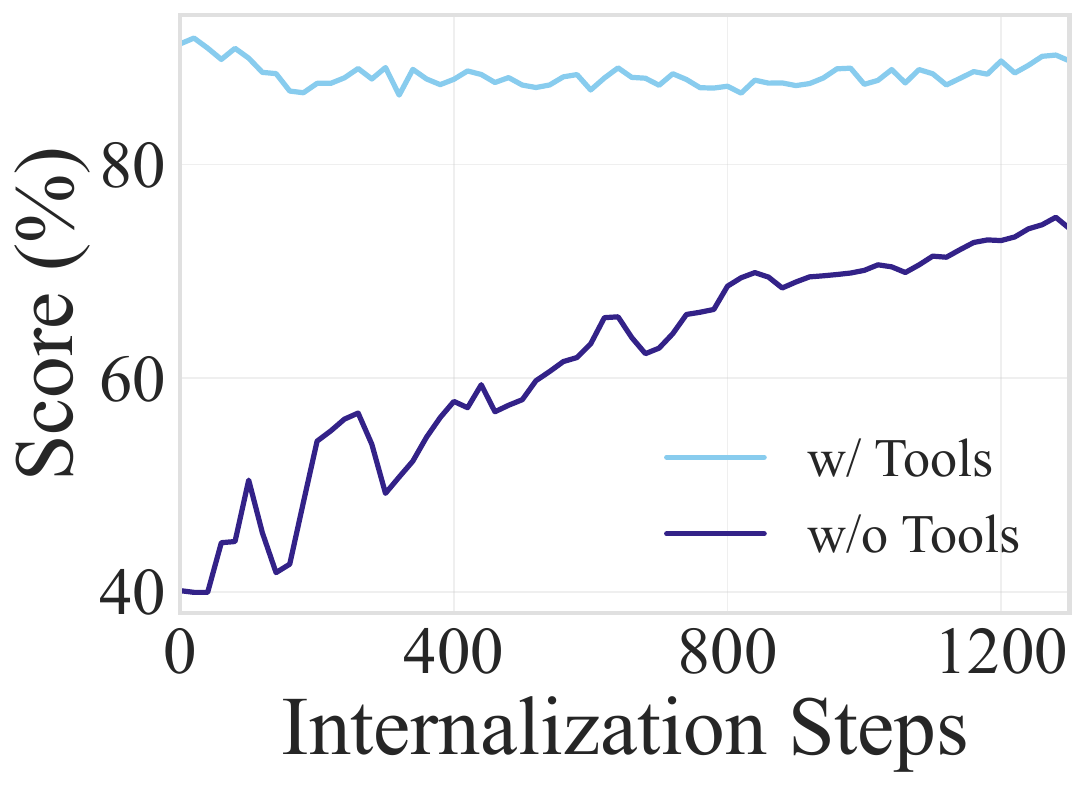}
        \caption{Performance vs. Steps}
        \label{fig:internalization_scores}
    \end{subfigure}
    \hfill
    \begin{subfigure}[t]{0.235\textwidth}
        \centering
        \includegraphics[width=\textwidth]{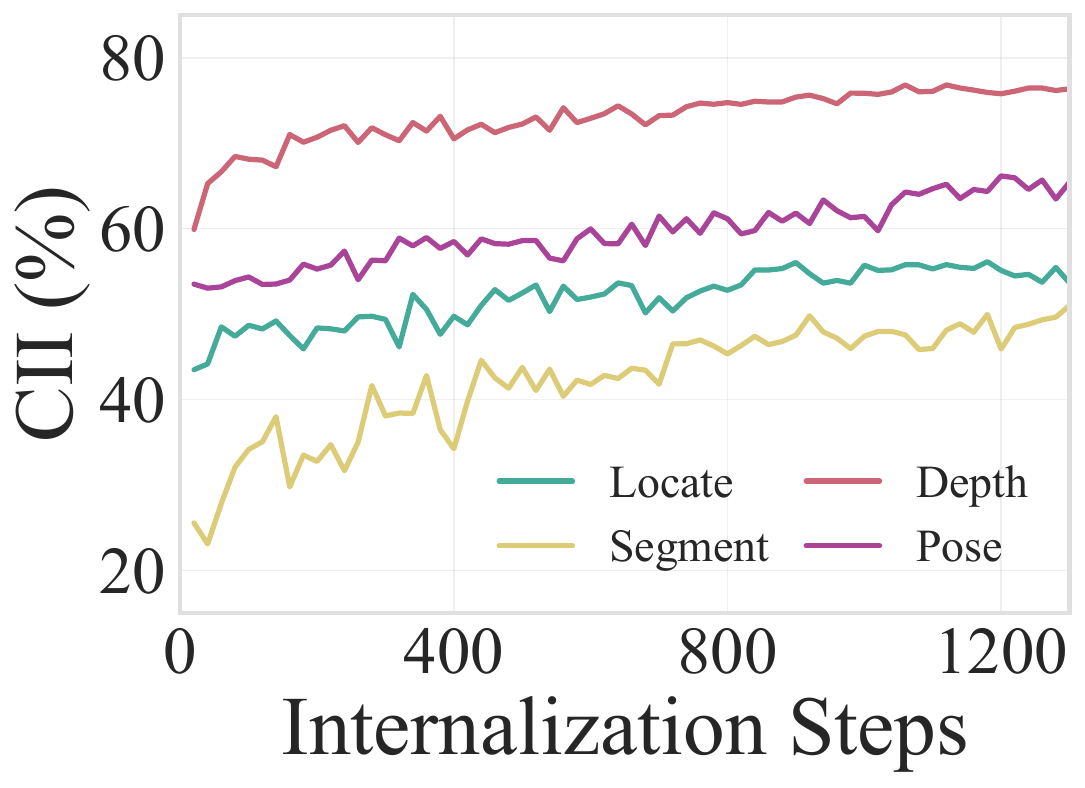}
        \caption{CII vs. Steps}
        \label{fig:internalization_index}
    \end{subfigure}
    \hfill
    \begin{subfigure}[t]{0.235\textwidth}
        \centering
        \includegraphics[width=\textwidth]{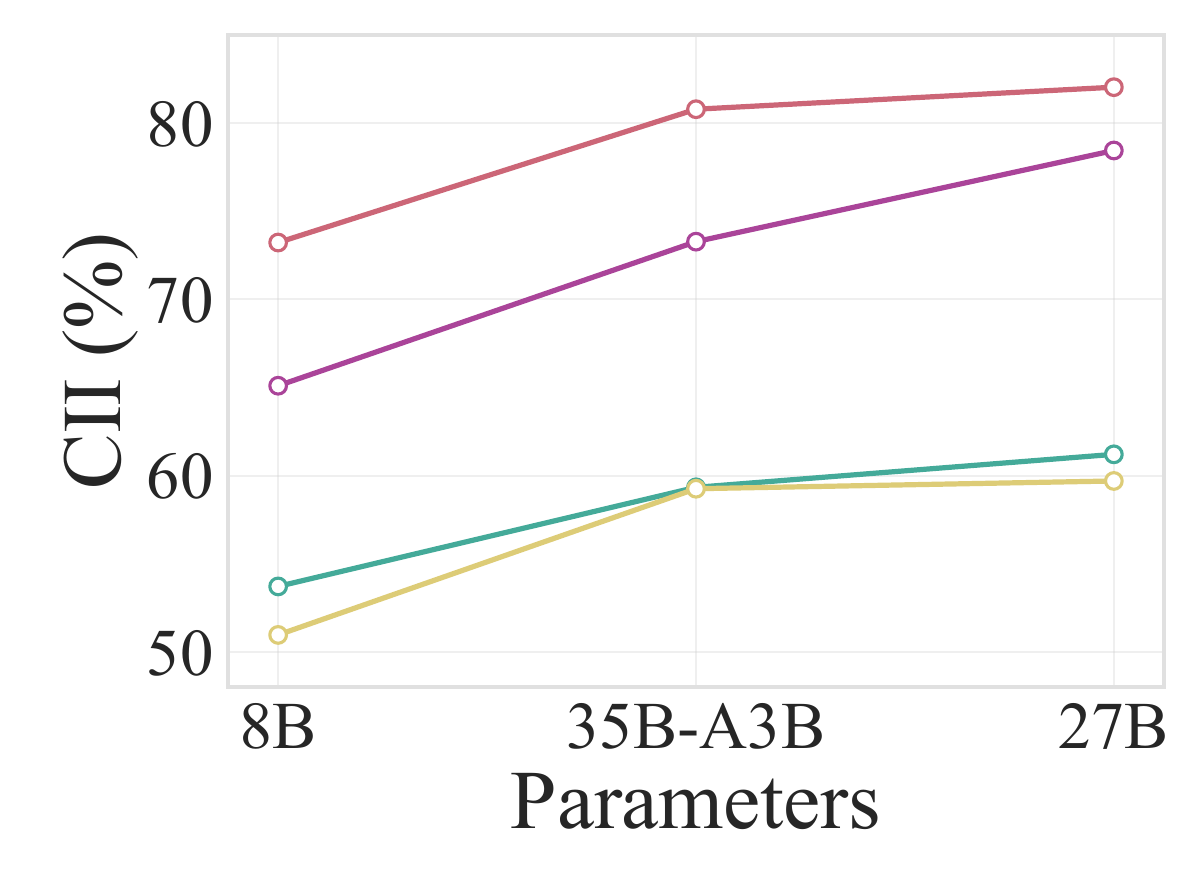}
        \caption{CII vs. Capacity}
        \label{fig:cii_scaling}
    \end{subfigure}
    \hfill
    \begin{subfigure}[t]{0.235\textwidth}
        \centering
        \includegraphics[width=\textwidth]{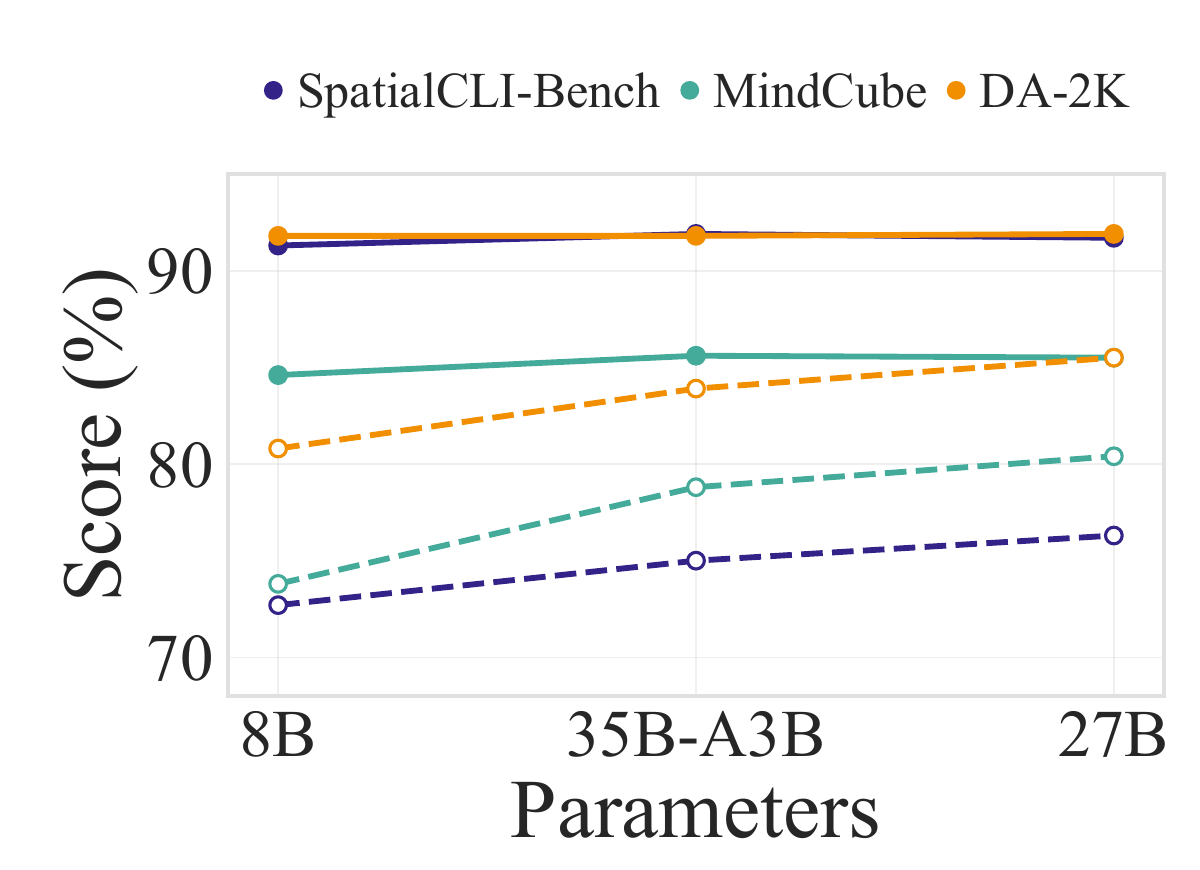}
        \caption{Performance vs. Capacity}
        \label{fig:benchmark_scaling}
    \end{subfigure}
    \caption{Capability internalization across training-data exposure and model capacity. (a) SpatialCLI-Bench scores under inference w/ Tools and w/o Tools over internalization-stage training steps. (b) Capability-specific CII over the same process. (c) Capability-specific CII across model capacities. (d) SpatialCLI-Bench, MindCube, and DA-2K scores under inference w/o Tools (dashed) and w/ Tools (solid) across model capacities.}
    \label{fig:capability_internalization}
\end{figure}

\subsection{Analysis}

In this subsection, we analyze RL training dynamics, the progression of capability internalization with increasing training data, and its scaling behavior across model capacities.
To quantify internalization, we use the Capability Internalization Index (CII), which measures how closely a model can reproduce the corresponding spatial-tool outputs without invoking the tools; higher values indicate stronger internalization, and the complete definition is provided in Appendix~E.

\paragraph{Training Dynamics.}
Figure~\ref{fig:training_dynamics} compares RL w/ Tools + SFT, RL w/ Tools w/o SFT, and RL w/o Tools.
RL w/ Tools + SFT starts from a strong tool-use policy, remains stable on both evaluation benchmarks, and finishes with higher training reward and substantially better MindCube performance than RL w/ Tools w/o SFT.
RL w/ Tools w/o SFT expands from 3.36 to 6.74 executed tool calls per trajectory, whereas RL w/ Tools + SFT remains near 2.56 throughout training, indicating that SFT provides a stable tool-use policy before RL.
RL w/o Tools generally improves direct-answer performance, but its w/ Tools performance begins to decline after approximately 100 training steps, indicating degraded tool-use competence.
Moreover, its direct-answer performance reaches only 52.7, well below the 72.7 achieved through capability internalization.
This gap indicates that SpatialCLI uses the available training data substantially more effectively than direct SFT or RL.
Appendix~F.1 provides the complete comparison.

\paragraph{Data Scaling of Capability Internalization.}
Figures~\ref{fig:internalization_scores} and~\ref{fig:internalization_index} show that, as the model is exposed to more internalization data, its w/o Tools score rises from 40.1\% to 74.0\%, while the four-capability macro-average CII increases from 45.6 to 61.6; the two curves exhibit closely aligned upward trends.
This synchronized improvement establishes a clear data-scaling trend: as internalization data increases, both task performance and CII continue to rise, indicating genuine transfer of specialist perceptual capabilities rather than final-answer memorization.
Although the w/ Tools score temporarily declines, it recovers close to its starting level, indicating that Dual-View Capability Internalization restores the learned tool-use policy after transient interference and avoids catastrophic forgetting.
Finally, the narrowing performance gap shows that, with increasing internalization data, spatial tools shift from an external capability source required for solving the tasks to an optional augmentation of the model's native capabilities.

\paragraph{Internalization continues to scale with model capacity after tool-use performance saturates.}
Figure~\ref{fig:benchmark_scaling} reveals a clear separation between tool-use and internalization scaling: across the three model variants, w/ Tools scores vary by at most 1.0 point on each benchmark, whereas both larger variants substantially outperform SpatialCLI-8B w/o Tools.
The saturation of w/ Tools performance is consistent with tool use requiring a relatively compact invocation-and-integration policy; once this policy is learned, the shared external specialists supply most of the required perceptual capability and thereby compress capacity-dependent differences.
Internalization is less prone to saturation because the model must instead encode and reproduce multiple specialist capabilities in its own parameters; accordingly, Figure~\ref{fig:cii_scaling} shows that SpatialCLI-35B-A3B and SpatialCLI-27B substantially outperform SpatialCLI-8B across all four capability-specific CII values.
These results show that external tools reduce performance differences across model sizes, whereas capability internalization continues to scale along two complementary dimensions: data exposure determines how fully a given model learns from the specialists, and model capacity affects how much of their capability it can absorb.

\subsection{Ablation Studies}

Using Qwen3-VL-8B-Instruct, we ablate how tools expose perceptual evidence and how tool-use trajectories are converted into capability-internalization supervision.

\paragraph{Structured returns provide more effective tool evidence.}
Table~\ref{tab:tool_return_format} compares the current structured tool returns with two red-border visual-return variants for Locate and Segment.
The structured-plus-visual variant retains the structured coordinates and polygons while appending each visualization as a new image, whereas the visual-only variant returns the visualization without the corresponding structured result; Depth and Pose returns remain unchanged.

\begin{table}[ht]
\centering
\small
\setlength{\tabcolsep}{1.8pt}
\begin{tabular}{lccc}
\toprule
\multirow{2}{*}{\textbf{Tool Return}} &
\multirow{2}{*}{\textbf{SpatialCLI-Bench}} &
\multicolumn{2}{c}{\textbf{BOPASK}} \\
\cmidrule(lr){3-4}
& & \textbf{Traj.} & \textbf{ObjRrr} \\
\midrule
No Tools (Initial Model) & 35.3 & 25.8 & 13.3 \\
\midrule
Visual Only & 45.9 & 45.6 & 18.3 \\
Structured + Visual & 66.0 & 54.5 & 20.1 \\
Structured Only (Ours) & 66.5 & 54.3 & 20.6 \\
\bottomrule
\end{tabular}
\caption{Tool-return format ablation with Qwen3-VL-8B-Instruct. Locate and Segment return red-border visualizations, structured coordinates or polygons, or both.}
\label{tab:tool_return_format}
\end{table}

Visual Only improves over the initial model on all three evaluations, showing that tool-provided perceptual evidence remains useful even without structured Locate and Segment outputs.
Adding structured coordinates and polygons yields substantially stronger performance, including a 20.1-point gain on SpatialCLI-Bench over Visual Only and consistent improvements on BOPASK.
Structured Only performs comparably to Structured + Visual across all evaluations.
Thus, explicit coordinates and polygons are the primary source of the improvement, while rendering the same evidence as additional images introduces extra visual context without a consistent score benefit.
Structured returns can also be directly converted into textual capability-internalization supervision, making them easier for current VLMs to internalize than multimodal tool outputs such as annotated images.

\paragraph{Progressive verbalization and dual-view training are both necessary.}
Table~\ref{tab:internalization_ablation} compares different forms of capability-internalization supervision.
One-Pass Dual-View replaces Progressive Evidence-Grounded Trajectory Verbalization with a single pass over the complete trajectory while retaining the same dual-view training.

\begin{table}[ht]
\centering
\small
\setlength{\tabcolsep}{2.0pt}
\begin{tabular}{lcccc}
\toprule
\multicolumn{1}{c}{\multirow{2.5}{*}{\textbf{Internalization Variant}}} & \multicolumn{2}{c}{\textbf{w/o Tools}} & \multicolumn{2}{c}{\textbf{w/ Tools}} \\
\cmidrule(lr){2-3}\cmidrule(lr){4-5}
& \textbf{Score} & \textbf{Tokens} & \textbf{Score} & \textbf{Tokens} \\
\midrule
Qwen3-VL-8B-Instruct & 35.3 & 2738 & 66.5 & 949 \\
\midrule
Final Answer Only & 52.7 & 19 & 52.1 & 28 \\
CoT + Answer & 45.0 & 5646 & 42.2 & 4401 \\
Internalization View Only & 71.1 & 227 & 62.6 & 235 \\
Tool-Use View Only & 42.2 & 6049 & 89.0 & 2155 \\
One-Pass Dual-View & 64.5 & 451 & 90.2 & 2599 \\
Full Dual-View (Ours) & \textbf{72.7} & 274 & \textbf{91.3} & 2480 \\
\bottomrule
\end{tabular}
\caption{Capability-internalization ablation on SpatialCLI-Bench. One-Pass and Full Dual-View use single-pass and progressive trajectory verbalization, respectively. Tokens include model outputs and, w/ Tools, tool returns; the best scores are bold.}
\label{tab:internalization_ablation}
\end{table}

For the untrained Qwen3-VL-8B-Instruct, external tools improve the score while reducing trajectory length from 2,738 to 949 tokens, indicating that tool use can also lower decoding cost.
The Qwen3.5-397B-A17B case studies in Appendix~H exhibit the same qualitative trend, with tools enabling more direct and reliable problem solving.
Final Answer Only yields concise but limited direct answers and loses tool use, while CoT + Answer is verbose and weak, showing that neither outcome supervision nor ungrounded reasoning transfers the perceptual process.
Internalization View Only produces strong and compact w/o Tools reasoning but weakens tool use, whereas Tool-Use View Only preserves w/ Tools performance without internalizing the capability, demonstrating the complementary roles of the two views.
One-Pass Dual-View supports both modes, but Full Dual-View reaches the best w/o Tools and w/ Tools scores of 72.7 and 91.3 with shorter outputs, validating progressive evidence consolidation.
Appendix~F.3 further confirms that Full Dual-View retains the controlled calling behavior learned from the tool-use view.

\section{Conclusion}

General VLMs lack the fine-grained perceptual capabilities required for embodied and spatial reasoning and cannot readily learn them from specialist vision models.
SpatialCLI addresses this gap by connecting specialist models as runtime tools, learning agentic tool use, and converting successful trajectories into dual-view supervision for capability internalization.
Across embodied and spatial benchmarks and model scales, SpatialCLI consistently improves both w/ Tools and w/o Tools reasoning, showing that external specialist capabilities can augment inference and be transferred into the VLM without sacrificing tool use.

SpatialCLI remains limited by specialist-tool coverage and reliability, with its current scope restricted to structured perceptual outputs and perception-centric tasks.
Future work will extend internalization to multimodal outputs using unified multimodal models capable of both understanding and generation, and integrate VLA tools under VLM planning for joint perception and action.

\clearpage
\bibliographystyle{assets/plainnat}\bibliography{references}
\clearpage\appendix\renewcommand{\thesection}{\Alph{section}}
\section*{Appendix}

\section*{Table of Contents}
\startcontents[sections]
\printcontents[sections]{l}{1}{\setcounter{tocdepth}{2}}
\clearpage
\suppressfloats[t]

\section{Algorithm Pseudocode}
\label{sec:algorithm_pseudocode}

Algorithm~\ref{alg:spatialcli_training} summarizes the complete Call--Learn--Internalize procedure.
Here, $\operatorname{Interact}(\pi,x,\mathcal{U},B)$ runs the ReAct agent with tools $\mathcal{U}$ and call budget $B$, returning $\xi=(I,q,\tau,y)$; $\tau$ records each call's reasoning, action, and observation, while $\xi$ retains complete assistant responses, including terminal reasoning.
Tool outputs are context only, and $\operatorname{SFT}$ minimizes mean negative log-likelihood over model-generated tokens.

\begin{algorithm}[H]
\caption{SpatialCLI: Call--Learn--Internalize}
\label{alg:spatialcli_training}
\begin{algorithmic}[1]
\REQUIRE Initial policy $\pi_0$; teacher policy $\pi_{\mathrm{teach}}$; shared task pool $\mathcal{D}$; RL subset $\mathcal{D}_{\mathrm{RL}}$; spatial tools $\mathcal{U}$; executed-call budget $B$; Tool-Use View weight $\lambda$
\ENSURE Trained SpatialCLI policy $\pi_{\theta}$
\STATE \textbf{Call:} Register $\mathcal{U}=\{\textsc{Locate},\textsc{Segment},\textsc{Depth},\textsc{Pose}\}$ in the live agent loop.
\STATE Define $\operatorname{Interact}(\pi,x,\mathcal{U},B)$ for $x=(I,q,y^*)$ as follows; $y^*$ is not exposed to $\pi$:
\STATE \hspace{\algorithmicindent}Initialize the complete history $\mathcal{H}\leftarrow(\mathcal{P}_{\mathcal{U}},I,q)$, trace $\tau\leftarrow()$, and executed-call count $k\leftarrow0$, where $\mathcal{P}_{\mathcal{U}}$ is the shared agentic prompt with the registrations of $\mathcal{U}$.
\WHILE{$k<B$}
    \STATE Sample the next complete assistant response $r=(z,u)\sim\pi(\cdot\mid\mathcal{H})$.
    \IF{$u$ is a terminal answer $y$}
        \STATE Append $r$ to $\mathcal{H}$.
        \RETURN The serialized interaction $\xi=(I,q,\tau,y)$.
    \ENDIF
    \STATE Parse $u$ as the next tool call $a_{k+1}$, execute $o_{k+1}\leftarrow\mathcal{U}(a_{k+1})$, and set $k\leftarrow k+1$.
    \STATE Append $(z,a_k,o_k)$ to $\tau$ and append the complete response $r$, observation $o_k$, and budget notice for $B-k$ remaining calls to $\mathcal{H}$; use the budget-exhausted instruction when $B-k=0$.
\ENDWHILE
\STATE Sample a terminal response $r=(z_{\mathrm{term}},y)\sim\pi(\cdot\mid\mathcal{H})$ with tool calls disabled and append $r$ to $\mathcal{H}$.
\RETURN The serialized interaction $\xi=(I,q,\tau,y)$.
\STATE \textbf{Learn---Cold-Start SFT:} $\widetilde{\mathcal{D}}_{\mathrm{SFT}}\leftarrow
\{\operatorname{Interact}(\pi_{\mathrm{teach}},x,\mathcal{U},B):x=(I,q,y^*)\text{ is sampled from }\mathcal{D}\}$.
\STATE $\mathcal{D}_{\mathrm{SFT}}\leftarrow
\{\xi\in\widetilde{\mathcal{D}}_{\mathrm{SFT}}:\text{valid call format, successful tool execution, and }y=y^*\}$.
\STATE $\pi_{\mathrm{SFT}}\leftarrow\operatorname{SFT}(\pi_0,\mathcal{D}_{\mathrm{SFT}})$ on reasoning, tool-call, and final-answer tokens.
\STATE \textbf{Learn---Agentic RL:} Initialize $\pi_{\theta}\leftarrow\pi_{\mathrm{SFT}}$.
\FOR{each agentic RL update}
    \STATE Sample $x=(I,q,y^*)\sim\mathcal{D}_{\mathrm{RL}}$.
    \STATE Set $\theta_{\mathrm{old}}\leftarrow\theta$.
    \STATE Sample $\{\xi_i\}_{i=1}^{G}\sim\mu_{\theta_{\mathrm{old}}}(\cdot\mid I,q)$ through $\operatorname{Interact}(\pi_{\theta_{\mathrm{old}}},x,\mathcal{U},B)$.
    \FOR{$i=1,\ldots,G$}
        \STATE Set $R_i\leftarrow V(y_i,y^*)$ from the final answer of $\xi_i$.
    \ENDFOR
    \STATE Update $\theta$ by maximizing $\mathcal{J}_{\mathrm{GRPO}}(\theta)$ in Equation~\eqref{eq:appendix_interactive_grpo}.
\ENDFOR
\STATE Set $\pi_{\mathrm{RL}}\leftarrow\pi_{\theta}$.
\STATE \textbf{Internalize:} Collect successful $\xi=\operatorname{Interact}(\pi_{\mathrm{RL}},x,\mathcal{U},B)$ with $y=y^*$ for tasks $x\in\mathcal{D}$.
\STATE Initialize $\mathcal{D}_{\mathrm{internal}}\leftarrow\varnothing$ and $\mathcal{D}_{\mathrm{agentic}}\leftarrow\varnothing$.
\FOR{each collected $\xi=(I,q,\tau,y^*)$, where $\tau=((z_t,a_t,o_t)_{t=1}^{T})$}
    \STATE Set $e_{<1}\leftarrow\varnothing$.
    \FOR{$t=1,\ldots,T$}
        \STATE $e_t\leftarrow\Psi_{\mathrm{ext}}(I,q,e_{<t},z_t,a_t,o_t)$; set $e_{<t+1}\leftarrow(e_1,\ldots,e_t)$.
    \ENDFOR
    \STATE Set $E_\tau\leftarrow(e_1,\ldots,e_T)$ and $c\leftarrow\Phi_{\mathrm{verb}}(I,q,E_\tau,y^*)$.
    \STATE Add the direct-answer sample $((I,q),(c,y^*))$ to $\mathcal{D}_{\mathrm{internal}}$.
    \STATE Add the original interaction $\xi$ to $\mathcal{D}_{\mathrm{agentic}}$, masking all returned tool outputs from the loss.
\ENDFOR
\STATE Initialize $\pi_{\theta}\leftarrow\pi_{\mathrm{RL}}$ and minimize
$\mathcal{L}_{\mathrm{CI}}=\mathcal{L}_{\mathrm{internal}}+\lambda\mathcal{L}_{\mathrm{agentic}}$
on $\mathcal{D}_{\mathrm{internal}}$ and $\mathcal{D}_{\mathrm{agentic}}$.
\RETURN $\pi_{\theta}$
\end{algorithmic}
\end{algorithm}

\section{Detailed Experimental Settings}
\label{sec:experimental_settings}

\subsection{Detailed Training Settings}
\label{sec:appendix_training}

\paragraph{Initial Models.}
We conduct training on Qwen3.6-27B~\citep{qwen2026qwen35}, Qwen3.6-35B-A3B~\citep{qwen2026qwen35}, and Qwen3-VL-8B-Instruct~\citep{bai2025qwen3vl} as the initial models.
All final models in the SpatialCLI series are obtained from their respective initial models through the same sequence of Cold-Start SFT, agentic RL, and trajectory-guided capability internalization.

\paragraph{Training Data.}
All three stages draw their stage-specific samples from a shared pool of 37,000 training tasks: 5,000 from Vlaser~\citep{yang2025vlaser}, 10,000 from MindCube-Train~\citep{wang2025mindcube}, 10,000 from BOPASK-Trajectory, 10,000 from BOPASK-Object-Rearrangement~\citep{bhat2026bopask}, and 2,000 from RefSpatial~\citep{zhou2025roborefer}.
We select these sources for their complementary spatial supervision: Vlaser contributes embodied question answering and grounding, MindCube-Train contributes multi-view spatial reasoning, the two BOPASK subsets contribute interaction-centric trajectory planning and object rearrangement, and RefSpatial contributes diverse 2D and 3D spatial referring.
Together, they cover single- and multi-image inputs as well as localization, depth, pose, trajectory, and rearrangement reasoning.
For Cold-Start SFT, we use Qwen3.5-397B-A17B~\citep{qwen2026qwen35} as the teacher model, equip it with the live agent loop described above, and sample approximately 5,000 tasks from this pool for tool-interaction annotation.
After filtering invalid tool-call formats, failed tool executions, and incorrect final answers, we retain approximately 40\% as high-quality correct trajectories, yielding approximately 2,000 SFT trajectories.
Agentic RL samples approximately 10,000 tasks from the same shared pool while explicitly excluding all tasks used for Cold-Start SFT.
Capability internalization uses approximately 42,000 successful tool-use trajectories collected by rolling out SpatialCLI-RL on tasks from this pool.
Both the turn-wise evidence consolidator and the global trajectory verbalizer use Qwen3.5-397B-A17B, with the same sampling and maximum-generation-length parameters as the evaluation configuration in Section~\ref{sec:appendix_evaluation}.
Because trajectory verbalization succeeds for nearly all retained trajectories, each trajectory produces one Capability-Internalization View sample and one Tool-Use View sample, yielding approximately 84,000 view-specific training samples.
We ensure that the training data have no textual or visual overlap with any benchmark evaluated in this work and therefore pose no risk of text or image leakage into evaluation.

\paragraph{Agentic RL Objective.}
For each task $x=(I,q,y^*)\sim\mathcal{D}_{\mathrm{RL}}$, the old policy $\pi_{\theta_{\mathrm{old}}}$ samples a group of $G$ complete interactions $\{\xi_i\}_{i=1}^{G}$ through live interaction with the spatial tools, and a task verifier assigns each interaction the outcome reward $R_i=V(y_i,y^*)$.
Most training tasks are multiple-choice questions, for which a deterministic parser extracts the final answer choice and assigns reward $1$ if it matches the ground truth and $0$ otherwise.
For BOPASK-Trajectory and BOPASK-Object-Rearrangement, we directly use the per-sample point-set score defined in Equation~\eqref{eq:bopask_score} as the trajectory reward.
We maximize the GRPO~\citep{shao2024deepseekmath} objective using the asymmetric Clip-Higher bounds and token-level policy-gradient reduction from DAPO~\citep{yu2025dapo}:
\begin{equation}
\begin{aligned}
\mathcal{J}_{\mathrm{GRPO}}(\theta)
&=
\mathbb{E}_{x\sim\mathcal{D}_{\mathrm{RL}},\{\xi_i\}_{i=1}^{G}\sim\mu_{\theta_{\mathrm{old}}}(\cdot\mid I,q)}
\\[-0.2em]
&\quad
\Bigg[
\frac{1}{\sum_{i=1}^{G}|\mathcal{T}_i|}
\sum_{i=1}^{G}\sum_{\ell\in\mathcal{T}_i}
\min\!\left(
\rho_{i,\ell}(\theta)\hat{A}_{i,\ell},
\operatorname{clip}\!\left(
\rho_{i,\ell}(\theta),
1-\varepsilon_{\mathrm{low}},
1+\varepsilon_{\mathrm{high}}
\right)\hat{A}_{i,\ell}
\right)
\Bigg].
\end{aligned}
\label{eq:appendix_interactive_grpo}
\end{equation}
Here, $\mu_{\theta_{\mathrm{old}}}$ is the rollout distribution induced by the old policy and tool execution, and $\mathcal{T}_i$ contains only model-generated token positions in $\xi_i$---reasoning, tool calls, and the final answer---excluding returned tool outputs.
For model-generated token $w_{i,\ell}$ with interaction history $h_{i,\ell}$, the token-level importance ratio is $\rho_{i,\ell}(\theta)=\pi_{\theta}(w_{i,\ell}\mid I,q,h_{i,\ell})/\pi_{\theta_{\mathrm{old}}}(w_{i,\ell}\mid I,q,h_{i,\ell})$.
The trajectory-level reward is broadcast to every model-generated token in the same interaction. The resulting group-relative advantage is
\begin{equation}
\hat{A}_{i,\ell}=\frac{R_i-\operatorname{mean}(\{R_j\}_{j=1}^{G})}{\operatorname{std}(\{R_j\}_{j=1}^{G})},\qquad \ell\in\mathcal{T}_i,
\label{eq:appendix_interactive_grpo_advantage}
\end{equation}

\paragraph{Training Configurations.}
Tables~\ref{tab:sft_training_configuration} and~\ref{tab:rl_training_configuration} summarize the configurations shared by Cold-Start SFT and capability internalization, and the agentic RL configuration, respectively.
The three initial models use the same hyperparameters.
Cold-Start SFT and capability internalization both use full-parameter fine-tuning.
Capability internalization uses the same settings as Cold-Start SFT, with the Tool-Use View loss weight set to $\lambda=0.5$.
Training uses a fixed random seed of 42.

\begin{table}[ht]
\centering
\footnotesize
\setlength{\tabcolsep}{3.2pt}
\begin{tabular}{@{}>{\bfseries}p{0.22\columnwidth}p{0.68\columnwidth}@{}}
\toprule
Category & Configuration \\
\midrule
\multirow{5}{*}{Optimization} & Optimizer: AdamW \\
& Weight Decay: 0.1 \\
& Learning Rate: $2 \times 10^{-5}$ \\
& LR Scheduler: Cosine \\
& Warmup Steps: 10 \\
\midrule
\multirow{4}{*}{Training} & Fine-Tuning Method: Full-Parameter \\
& Batch Size: 64 \\
& Context Length: 32768 \\
& Epochs: 2 \\
\bottomrule
\end{tabular}
\caption{Training configuration shared by Cold-Start SFT and capability internalization.}
\label{tab:sft_training_configuration}
\end{table}

\begin{table}[ht]
\centering
\footnotesize
\setlength{\tabcolsep}{3.2pt}
\begin{tabular}{@{}>{\bfseries}p{0.22\columnwidth}p{0.68\columnwidth}@{}}
\toprule
Category & Configuration \\
\midrule
\multirow{8}{*}{Algorithm} & RL Algorithm: GRPO \\
& Group Size: 8 \\
& Clip Ratio Low: 0.20 \\
& Clip Ratio High: 0.28 \\
& Dual-Clip $C$: 3.0 \\
& Entropy Coef: 0 \\
& KL Loss Coef: 0 \\
& Loss Aggregation: Token Mean \\
\midrule
\multirow{7}{*}{Rollout} & Maximum Input Length: 32768 \\
& Maximum Output Length: 32768 \\
& Tool-Call Budget: 10 \\
& Temperature: 1.0 \\
& Top-P: 1.0 \\
& Top-K: $-1$ \\
& Rollout Precision: FP16 \\
\midrule
\multirow{7}{*}{Training} & Training Batch Size: 128 \\
& Mini Batch Size: 32 \\
& Optimizer: AdamW \\
& Learning Rate: $1 \times 10^{-6}$ \\
& Weight Decay: 0.01 \\
& Gradient Clip: 1.0 \\
& Training Precision: FP16 \\
\bottomrule
\end{tabular}
\caption{Agentic RL training configuration.}
\label{tab:rl_training_configuration}
\end{table}

\subsection{Detailed Evaluation Settings}
\label{sec:appendix_evaluation}

\paragraph{Evaluation Benchmarks.}
We evaluate SpatialCLI on SpatialCLI-Bench, MindCube~\citep{wang2025mindcube}, selected subsets of MMSI~\citep{yang2025mmsi}, DA-2K~\citep{yang2024depthanythingv2}, and selected subsets of BOPASK~\citep{bhat2026bopask}.
Specifically, we use Motion-Cam and Pos-Cam-Cam from MMSI and Trajectory and Object-Rearrangement from BOPASK.
For the two BOPASK subsets, we use the same continuous point-set score during agentic RL and evaluation.
Model outputs and references express 2D points on a $0$--$999$ image-coordinate scale; before scoring, we divide both coordinates by $1000$.
Let $\hat{\mathcal{P}}$ and $\mathcal{P}^*$ be the resulting predicted and reference point sets.
Their symmetric Chamfer distance and per-sample score are
\begin{equation}
\begin{split}
d_{\mathrm{CD}}(\hat{\mathcal{P}},\mathcal{P}^*)
&=
\frac{1}{2}
\left[
\frac{1}{|\hat{\mathcal{P}}|}
\sum_{\hat p\in\hat{\mathcal{P}}}
\min_{p^*\in\mathcal{P}^*}\|\hat p-p^*\|_2
+
\frac{1}{|\mathcal{P}^*|}
\sum_{p^*\in\mathcal{P}^*}
\min_{\hat p\in\hat{\mathcal{P}}}\|p^*-\hat p\|_2
\right],\\
s_{\mathrm{BOPASK}}(\hat{\mathcal{P}},\mathcal{P}^*)
&=
\max\left(
0,\,
1-\frac{d_{\mathrm{CD}}(\hat{\mathcal{P}},\mathcal{P}^*)}{0.15}
\right).
\end{split}
\label{eq:bopask_score}
\end{equation}
If an output contains no valid predicted points, its per-sample score is zero.
For each BOPASK subset, we report the mean per-sample score as a percentage.
Under w/o Tools inference, no tools are registered and no system prompt is inserted; under w/ Tools inference, all evaluated models are provided with the same four spatial tools used to train SpatialCLI.
We parse final answers with benchmark-specific deterministic parsers and treat malformed or missing final answers as incorrect.

\paragraph{Evaluation Configurations.}
Unless a benchmark requires a different official setting, we use temperature 1.0, top-$p$ 0.95, top-$k$ 20, min-$p$ 0.0, a presence penalty of 1.5, a repetition penalty of 1.0, and a maximum generation length of 40,960 tokens. All task-performance evaluations permit at most ten tool calls per sample.
We report the mean over three evaluation runs, without fixing evaluation seeds.

\subsection{Hardware and Software Environment}
\label{sec:hardware_software_environment}

All experiments were conducted on a cluster whose nodes were equipped with 184 Intel Xeon CPU cores, 1.8~TiB of system memory, and 16 PPU-ZW810E accelerators, each with approximately 96~GB of device memory.
Both SFT and RL used four nodes.
The software environment used Ubuntu 24.04.2 LTS, a CUDA 12.9-compatible toolchain, Python 3.12.3, PyTorch 2.9.0, Transformers 5.5.4, vLLM 0.18.0+ppu2.0.0, verl 0.9.0.dev0, and Ray 2.49.2.

\FloatBarrier
\section{SpatialCLI-Bench}
\label{sec:spatialcli_bench}

\subsection{Scope and Sample Format}

SpatialCLI-Bench is a 516-example English six-choice visual question answering benchmark for evaluating tool-grounded compositional spatial reasoning.
Each example contains one or two images, one question, and six answer choices with a unique correct answer.
Each example has a private executable oracle tool plan of at most five steps, which is used for construction and auditing but is not exposed in the public question.
Every example jointly evaluates at least two of three capability groups: grounding and 2D region relations (G), metric depth (D), and object pose or cross-view camera motion (P).

\subsection{Data Provenance and Composition}

Every example has a unique candidate ID and a unique upstream visual-source ID.
No complete visual unit is duplicated, every example is associated with exactly one visual source, and both images of a two-image example come from the same source.
The visual sources are MindCube~\citep{wang2025mindcube}, HOPE~\citep{tyree20226}, LINEMOD~\citep{hinterstoisser2012model}, YCB-V~\citep{xiang2017posecnn}, HANDAL~\citep{guo2023handal}, CA-1M~\citep{lazarow2024cubify}, OpenImages~\citep{kuznetsova2018open}, and RefSpatial-Blender~\citep{zhou2025roborefer}.
The 195 MindCube-derived examples reuse image data from the existing MindCubeBench evaluation set.
Table~\ref{tab:spatialcli_bench_sources} gives the complete composition by visual source.

\begin{table}[ht]
\centering
\small
\setlength{\tabcolsep}{4.0pt}
\begin{tabular}{@{}lrr@{}}
\toprule
\textbf{Visual Source} & \textbf{Count} & \textbf{Share} \\
\midrule
MindCube & 195 & 37.79\% \\
HOPE & 110 & 21.32\% \\
LINEMOD & 46 & 8.91\% \\
YCB-V & 5 & 0.97\% \\
HANDAL & 1 & 0.19\% \\
CA-1M & 116 & 22.48\% \\
OpenImages & 29 & 5.62\% \\
RefSpatial-Blender & 14 & 2.71\% \\
\bottomrule
\end{tabular}
\caption{Visual-source provenance of the 516 SpatialCLI-Bench examples.}
\label{tab:spatialcli_bench_sources}
\end{table}

The benchmark contains 243 single-image examples (47.09\%) and 273 two-image examples (52.91\%).

\subsection{Capability Composition and Oracle Plans}

Table~\ref{tab:spatialcli_bench_capabilities} summarizes the joint capability composition.
Grounding appears in 414 examples (80.23\%), depth in 403 (78.10\%), and pose in 402 (77.91\%), giving nearly balanced marginal coverage.
All two-image examples include P through cross-view camera-motion or pose evidence.

\begin{table}[ht]
\centering
\small
\setlength{\tabcolsep}{7.0pt}
\begin{tabular}{@{}lrr@{}}
\toprule
\textbf{Capability Combination} & \textbf{Count} & \textbf{Share} \\
\midrule
GD & 114 & 22.09\% \\
GP & 113 & 21.90\% \\
DP & 102 & 19.77\% \\
GDP & 187 & 36.24\% \\
\bottomrule
\end{tabular}
\caption{Joint capability composition of SpatialCLI-Bench.}
\label{tab:spatialcli_bench_capabilities}
\end{table}

The 516 private oracle plans contain 1,964 tool steps, averaging 3.81 steps per example.
There are 12 two-step, 156 three-step, 268 four-step, and 80 five-step plans.
Across all plans, \texttt{query\_segment}, \texttt{query\_depth}, \texttt{query\_pose}, and \texttt{query\_locate} are invoked 862, 471, 404, and 227 times, respectively.

\subsection{Multi-Stage Construction Pipeline}

\paragraph{Frontier Multimodal Model Pre-Annotation.}
A frontier multimodal model first builds a conservative structured inventory of uniquely referable visible entities for each candidate image or image pair.
The inventory records each entity's image index, visible description and attributes, a short tool query, and whether the entity is uncertain; rejected entities and their rejection reasons are recorded separately.
We use Gemini 3.1 Pro~\citep{google2026gemini31pro} for this stage and retain only inventories with at least two reliable entities, including reliable coverage of both images for a two-image candidate.

\paragraph{Specialist-Vision-Model Fine Annotation.}
Specialist vision models then collect capability-aligned evidence for up to four reliable entities.
Locate and Segment establish entity positions and regions; Depth is queried at a representative point derived from localization or segmentation; Pose estimates cross-view camera motion for two-image candidates or object orientation for up to two entities in single-image candidates.
Each response is stored with its exact tool arguments, and a candidate is rejected when a required service returns missing, explicit-error, or empty evidence.

\paragraph{Frontier Multimodal Model Fine Annotation.}
Conditioned on the entity inventory and collected specialist evidence, Gemini 3.1 Pro generates the English question, one correct answer, five plausible distractors, a private evidence-grounded rationale, and a minimal reproducible oracle plan.
The generation prompt requires every factual clause in the correct answer to be supported by specialist evidence and each distractor to be false for a specific evidence-grounded reason.
Deterministic validation then enforces six unique choices, valid image indices and step dependencies, the five-step budget, exact capability--plan correspondence, and an exact match between every oracle step and a collected tool call; the choices are shuffled deterministically only after validation.

\subsection{Independent Human Review and Filtering Yield}

The automated construction pipeline described above initially generates 720 candidate questions.
Human experts then answer each question independently without access to the answer generated by Gemini 3.1 Pro.
The independent review is conducted by two graduate students with research backgrounds in embodied intelligence.
The 516 candidates for which all independent human answers agree with the generated answer are retained in SpatialCLI-Bench, for an overall retention rate of 71.67\%; the remaining 204 candidates are rejected because of answer disagreement.
The supporting tool evidence and minimal executable tool plan are retained only as private audit metadata.

\section{Spatial Tool Interfaces and Implementation}
\label{sec:tool_interfaces}

\subsection{Coordinate and Serialization Convention}

Following the JSON grounding convention used by Qwen vision-language models~\citep{bai2025qwen3vl}, SpatialCLI represents points and boxes with \texttt{point\_2d} and \texttt{bbox\_2d}.
All coordinates are integers in $[0,999]$, with the origin at the upper-left corner.
We introduce \texttt{polygon\_2d} as a SpatialCLI-specific extension for segmentation masks.

A bounding box is serialized as $[x_{\min},y_{\min},x_{\max},y_{\max}]$.
A point is serialized as $[x,y]$.
A polygon is represented by an ordered list of boundary vertices, and multiple connected components are represented as a list of polygons.
The Segment tool extracts external mask contours and represents each connected component with at most 16 vertices, targeting a rasterized mask IoU of 0.97 whenever feasible within this vertex budget.
The polygon is rasterized at the original resolution whenever mask IoU is required.

\subsection{Spatial Tool Interfaces}

The agent sees only four model-level interfaces: \texttt{query\_locate}, \texttt{query\_segment}, \texttt{query\_depth}, and \texttt{query\_pose}; it does not select the underlying specialist vision models.
The service layer encapsulates six specialist vision models: Locate Anything~\citep{wang2026locateanything} and Grounding DINO~\citep{liu2023groundingdino} for fused localization, SAM 3~\citep{sam3} for segmentation, Depth Anything 3~\citep{depthanything3} for metric depth, Orient Anything V2~\citep{wang2026orientanythingv2} for object orientation, and VGGT~\citep{wang2025vggt} for multi-view camera motion.
All four interfaces share the XML-wrapped JSON calling protocol and accept an optional \texttt{image\_indices} field with one-based image or frame indices.
At runtime, only responses matched by the fixed regular expression for this protocol are treated as tool calls; all unmatched responses, including those with malformed tool-call syntax, are treated as final answers.
Their complete registrations are provided in Boxes~\ref{box:locate_tool}--\ref{box:pose_tool}.

\paragraph{Locate.}
Locate fuses detections from Locate Anything and Grounding DINO, while exposing only the fused boxes and center points to the agent.
Both backends process the same image--query pair in parallel.
Grounding DINO detections are filtered at a confidence threshold of 0.30 and suppressed using box NMS with an IoU threshold of 0.50.
For a Locate Anything box $b_i^L$, we select the unmatched Grounding DINO box $b_{j^*(i)}^D$ with the largest IoU.
If their IoU is at least 0.50, the cross-confirmed box is
\begin{equation}
b_i^F=\operatorname{round}\left(\frac{2b_i^L+b_{j^*(i)}^D}{3}\right),
\end{equation}
which gives Locate Anything twice the coordinate weight of Grounding DINO.
Unmatched predictions from either backend are retained to preserve recall.
The combined candidates are greedily deduplicated: a candidate $b_i$ is removed if a retained box $b_j$ satisfies
\begin{equation}
\operatorname{IoU}(b_i,b_j)\geq 0.75
\quad\text{or}\quad
\frac{|b_i\cap b_j|}{\min(|b_i|,|b_j|)}\geq 0.90.
\end{equation}
Here, $|b|$ denotes the area of box $b$.
The second criterion removes near-contained duplicates that may have a modest box IoU because of their different areas.
Backend identities, confidence scores, and cross-model IoUs remain internal.

\begin{spatialcliprompt}[label={box:locate_tool},colback=blue!5!white,colframe=blue!75!black]{Locate Tool Registration}
{
  "name": "query_locate",
  "description": "Locate every visible instance matching a short language description. Use this when exact object positions, centers, boxes, counts, or left/right order depend on reliable grounding. Query a concrete visual category and attributes only. For 'closest', 'farthest', 'leftmost', or 'second' questions, do not put that relation in the query: locate all instances of the base category, then compare their returned centers or query their depths. All benchmark geometry and tool coordinates use the same normalized 0-999 image space. The tool returns bbox_2d and point_2d in that space; pass point_2d directly to query_depth when needed. It may return zero, one, or several instances; do not assume the first instance is the only one. Prefer this tool for answers that require explicit path points, object markers, or boxes. Answer simple categorical left/right yes-no questions directly when visually clear. It cannot infer five-point grasp-finger geometry or projected 3D cuboid corners from a 2D box, so do not treat its center or box corners as those answers. Tools are optional. In a multi-image or sampled-video task, omit image_indices to use all images, or provide one or more 1-based image/frame numbers to select a subset.",
  "parameters": {
    "type": "object",
    "properties": {
      "query": {
        "type": "string",
        "description": "Concrete visual category/attributes, without positional or depth superlatives."
      },
      "image_indices": {
        "type": "array",
        "items": {"type": "integer", "minimum": 1},
        "minItems": 1,
        "uniqueItems": true,
        "description": "Optional 1-based image/frame numbers. Omit to use the only image in a single-image task or all images jointly in a multi-image task."
      }
    },
    "required": ["query"]
  }
}

Example:
<tool_call>
{"name": "query_locate", "arguments": {"query": "red mug", "image_indices": [1]}}
</tool_call>

Possible return:
{"count": 1, "result": [{"bbox_2d": [532, 470, 628, 675], "point_2d": [580, 572]}]}
\end{spatialcliprompt}

\paragraph{Segment.}
Segment uses a short text query to obtain SAM 3 masks and returns each reliable instance as a box, center point, and one or more polygonal connected components.
Predictions with confidence at most 0.30 are removed, after which confidence-ordered mask suppression removes a candidate whose mask IoU with any retained instance is at least 0.90.
For each retained mask, we extract pixel-level external contours, discard connected components smaller than $16$ pixels$^2$, and order the remaining components by area.
Because only external contours are serialized, holes inside a component are not represented explicitly.

Contour vertices are quantized into the shared integer coordinate space; adjacent duplicate vertices and an explicit closing vertex are removed.
For a closed contour $\Gamma_k$ with perimeter $L_k$, we apply Douglas--Peucker simplification with tolerance $\varepsilon_k(r)=rL_k$ over
\begin{equation}
r\in\{0\}\cup\operatorname{LogSpace}(10^{-5},0.25,72).
\end{equation}
Each candidate polygon $P$ is first quantized into the final integer coordinate space, mapped back to the original resolution, and rasterized as $\mathcal{R}(P)$.
Its fidelity to the corresponding mask component $M_k$ is
\begin{equation}
Q(P;M_k)=\operatorname{IoU}\left(M_k,\mathcal{R}(P)\right)
=\frac{|M_k\cap\mathcal{R}(P)|}{|M_k\cup\mathcal{R}(P)|}.
\end{equation}
We choose the candidate with the fewest vertices subject to
\begin{equation}
Q(P;M_k)\geq 0.97,\qquad 3\leq |P|\leq 16,
\end{equation}
breaking ties in favor of higher IoU.
We then greedily attempt to remove low-contribution vertices in ascending order of
\begin{equation}
a_v=\left|(p_v-p_{v-1})\times(p_{v+1}-p_{v-1})\right|,
\end{equation}
accepting a removal only if the rasterized IoU remains at least 0.97.
If no polygon reaches the target within the 16-vertex budget, the bounded candidate with the highest IoU is returned instead; thus 0.97 is a target rather than an unconditional guarantee.
The returned \texttt{point\_2d} is the center of the SAM 3 predicted box, not the mask centroid, and need not lie inside the mask.

\begin{spatialcliprompt}[label={box:segment_tool},colback=green!3!white,colframe=green!50!black]{Segment Tool Registration}
{
  "name": "query_segment",
  "description": "Segment every visible instance matching a short language description. Use this for exact object boundaries, occupied regions, shapes, containment, overlap, or placement areas; use query_locate instead when centers, counts, or simple left/right order are sufficient. Pass a concrete visual category/attributes only, without closest/leftmost/second relations, and do not pass a box. All benchmark geometry and tool coordinates use the same normalized 0-999 image space. Each returned instance contains bbox_2d, point_2d, and polygon_2d in that space. The point is the mask bounding-box center and is easiest to use for paths or object markers. polygon_2d is a list because one mask may have disconnected parts. Prefer this tool for object markers, occupied regions, and boundary-sensitive coordinate tasks. Do not use a mask center or polygon as five-point grasp-finger geometry or projected 3D cuboid corners. Answer categorical yes-no and relative-depth questions directly unless an exact boundary is genuinely required. Tools are optional. In a multi-image or sampled-video task, omit image_indices to use all images, or provide one or more 1-based image/frame numbers to select a subset.",
  "parameters": {
    "type": "object",
    "properties": {
      "query": {
        "type": "string",
        "description": "Short description of the target instances."
      },
      "image_indices": {
        "type": "array",
        "items": {"type": "integer", "minimum": 1},
        "minItems": 1,
        "uniqueItems": true,
        "description": "Optional 1-based image/frame numbers. Omit to use the only image in a single-image task or all images jointly in a multi-image task."
      }
    },
    "required": ["query"]
  }
}

Example:
<tool_call>
{"name": "query_segment", "arguments": {"query": "red mug"}}
</tool_call>

Possible return:
{"count": 1, "result": [{"bbox_2d": [532, 470, 628, 675], "point_2d": [580, 572], "polygon_2d": [[[548, 475], [605, 472], [623, 516], [609, 670]]]}]}
\end{spatialcliprompt}

\paragraph{Depth.}
Depth uses the DA3NESTED-GIANT-LARGE-1.1 metric backend to return camera-axis distance in meters at one or more selected image points.
Inference uses a processing resolution of 504 pixels, and each query reads its corresponding depth-map pixel directly without spatial smoothing.
Metric-scale alignment is made deterministic across workers.
The returned \texttt{depth\_m} is a monocular estimate of camera-axis $Z$-depth, rather than Euclidean range from the camera center or a physical sensor measurement.

\begin{spatialcliprompt}[label={box:depth_tool},colback=orange!5!white,colframe=orange!75!black]{Depth Tool Registration}
{
  "name": "query_depth",
  "description": "Query depth at one or more image points. Pass every point needed for a closer/farther, front/behind, or distance comparison in one call. Point coordinates are normalized to 0-999 and the number of points is unrestricted. For named objects, first use query_locate to obtain their center points, then query all centers together. Each result contains depth_m, an estimated camera-axis distance in meters; larger means farther. Monocular depth estimates can be noisy: answer directly when the relative depth is already visually clear, and call this tool only when depth is essential and genuinely ambiguous. In a multi-image or sampled-video task, omit image_indices to use all images, or provide one or more 1-based image/frame numbers to select a subset.",
  "parameters": {
    "type": "object",
    "properties": {
      "points": {
        "type": "array",
        "minItems": 1,
        "description": "One or more [x, y] points in normalized 0-999 coordinates.",
        "items": {
          "type": "array",
          "items": {"type": "number"},
          "minItems": 2,
          "maxItems": 2
        }
      },
      "image_indices": {
        "type": "array",
        "items": {"type": "integer", "minimum": 1},
        "minItems": 1,
        "uniqueItems": true,
        "description": "Optional 1-based image/frame numbers. Omit to use the only image in a single-image task or all images jointly in a multi-image task."
      }
    },
    "required": ["points"]
  }
}

Example:
<tool_call>
{"name": "query_depth", "arguments": {"points": [[320, 480], [700, 510]]}}
</tool_call>

Possible return:
{"result": [{"point_2d": [320, 480], "depth_m": 2.41}, {"point_2d": [700, 510], "depth_m": 4.12}]}
\end{spatialcliprompt}

\paragraph{Pose.}
Pose routes a named-object query to Orient Anything V2 and the exact query \texttt{camera motion} to the multi-view VGGT camera-pose backend.
For object orientation, Pose first invokes the fused Locate interface and enlarges every detected box by 10\% of its width and height on each side, with a minimum padding of two pixels.
The crop is resized with its aspect ratio preserved, padded to $518\times518$, and processed by Orient Anything V2, which predicts azimuth, elevation, and roll using 360, 180, and 360 discrete bins, respectively.
Only the horizontal orientation is exposed to the agent.
For azimuth $\varphi_{\mathrm{az}}$, its eight-direction sector is
\begin{equation}
\nu_{\mathrm{az}}=\left\lfloor
\frac{(\varphi_{\mathrm{az}}+22.5^\circ)\bmod 360^\circ}{45^\circ}
\right\rfloor.
\end{equation}
The sector is converted into the complementary fields \texttt{visible\_side} and \texttt{facing\_direction\_camera}; in the latter, \texttt{front} points into the image and \texttt{back} points toward the camera.

For camera motion, VGGT jointly estimates world-to-camera extrinsics $E_i=[R_i\mid t_i]$ for the selected views in their given order.
The camera center and the adjacent-view translation expressed in the source-camera frame are
\begin{equation}
C_i=-R_i^\top t_i,\qquad
\Delta_i=R_i(C_{i+1}-C_i),\qquad
\hat{\Delta}_i=\frac{\Delta_i}{\|\Delta_i\|_2}
\ \text{if }\|\Delta_i\|_2\geq 0.002.
\end{equation}
Before serializing $\hat{\Delta}_i$ as signed axes, we negate its $Y$ and $Z$ components to obtain the exposed convention of $+X$ right, $+Y$ up, and $-Z$ forward.
The relative view rotation is
\begin{equation}
R_{i\rightarrow i+1}=R_{i+1}R_i^\top.
\end{equation}
Translation and view rotation are summarized separately using these right-handed camera axes.
A translation norm below 0.002 is treated as the same position, while rotations below $2^\circ$ are suppressed.
For $n_{\mathrm{view}}$ selected images, the tool returns $n_{\mathrm{view}}-1$ summaries for adjacent pairs rather than all pairwise relations.
Only scale-independent motion directions and signed axes are exposed; metric translation magnitudes, continuous angles, and camera matrices remain internal.

\begin{spatialcliprompt}[label={box:pose_tool},colback=purple!5!white,colframe=purple!75!black]{Pose Tool Registration}
{
  "name": "query_pose",
  "description": "Query a named object's facing direction in one image, or camera motion across images. For an object, pass only a short visible description. The result returns visible_side (the object's side facing the camera) and facing_direction_camera, where front means deeper into the image, back means toward the camera, and left/right are the viewer's image sides. For viewpoint change, pass exactly query='camera motion'. Each result describes to_image relative to from_image. Use position only for questions about the camera's shooting location or movement; use view_rotation only for questions about where the view turns. direction directly matches ordinary words such as forward-right, left, up, or clockwise. position.axes uses signed right-handed coordinates where +X=right, +Y=up, and -Z=forward. view_rotation.axes is the signed camera-pose rotation and should be used directly only when answer choices explicitly mention positive/negative X/Y/Z axes; for ordinary turn words use view_rotation.direction. same-position means no reliable translation. Omit image_indices for image 2 relative to image 1; reverse them only when the question explicitly asks for image 1 relative to image 2. Use object mode only for object orientation and camera-motion mode only for actual viewpoint change; otherwise answer directly. In a multi-image or sampled-video task, omit image_indices to use all images, or provide one or more 1-based image/frame numbers to select a subset.",
  "parameters": {
    "type": "object",
    "properties": {
      "query": {
        "type": "string",
        "description": "Object description, or 'camera motion'."
      },
      "image_indices": {
        "type": "array",
        "items": {"type": "integer", "minimum": 1},
        "minItems": 1,
        "uniqueItems": true,
        "description": "Optional 1-based image/frame numbers. Omit to use the only image in a single-image task or all images jointly in a multi-image task."
      }
    },
    "required": ["query"]
  }
}

Object-orientation example:
<tool_call>
{"name": "query_pose", "arguments": {"query": "red mug", "image_indices": [1]}}
</tool_call>

Possible return:
{"result": [{"image_index": 1, "bbox_2d": [532, 470, 628, 675], "visible_side": "front-right", "facing_direction_camera": "back-left"}]}

Camera-motion example:
<tool_call>
{"name": "query_pose", "arguments": {"query": "camera motion", "image_indices": [1, 2]}}
</tool_call>

Possible return:
{"result": [{"from_image": 1, "to_image": 2, "position": {"direction": "forward-right", "axes": ["+X", "-Z"]}, "view_rotation": {"direction": "right", "axes": ["+Y"], "dominant_axis": "+Y"}}]}
\end{spatialcliprompt}

\paragraph{Deployment and Runtime.}
All six specialist-model services are jointly deployed on two GPUs.
Compared with the central VLM, these specialist backends are generally compact and introduce modest runtime overhead.
During long-running RL training, the observed mean latency per spatial-tool call was 2.916 seconds.

\paragraph{Cache Design.}
Each specialist backend uses a content-addressed two-tier cache: a bounded in-process least-recently-used (LRU) cache backed by a persistent disk cache under \texttt{outputs/cache/}.
Each key is a SHA-256 digest derived from the input-image content hashes rather than file paths, together with the request text or mode, the relevant model or checkpoint fingerprint, and all inference and post-processing settings that can affect the returned result.
Consequently, identical image content can be reused across path aliases, whereas a changed image, model checkpoint, query, backend, or output-affecting parameter produces a different key.
Lookup checks memory before disk and promotes a disk hit into the in-memory LRU.
On a miss, the service executes the specialist model and atomically persists the successful result; failed requests are not cached, and per-process locks prevent concurrent duplicate inference.

The cache granularity follows each backend's computation.
Depth caches the full metric-depth map, intrinsics, confidence, image size, and scale factor per image, allowing different queried points on the same image to reuse one forward pass.
Segment maintains both a two-entry image-state cache, which reuses the SAM~3 image encoding across text queries, and 128-entry result caches keyed by image, query, and segmentation settings.
Locate and Pose each maintain 128-entry result LRUs; their disk entries store the corresponding structured outputs, while Pose additionally persists raw camera matrices for camera-motion requests.
Pose keys distinguish the ordered image sequence, object-or-camera mode, selected camera backend, resolution, model fingerprints, and, for object orientation, the localization service.
The equal-weight macro-average of the reported Depth, Segment, Pose, and Locate rates is approximately 58.8\%; a request-weighted rate would additionally depend on the tool-call composition.
The bounded LRU capacities limit memory growth, while cache hits bypass repeated specialist forward passes and therefore add negligible accelerator compute beyond lookup and output serialization.
Table~\ref{tab:tool_cache_hit_rates} reports the observed cache hit rates for each spatial tool.

\begin{table}[ht]
\centering
\small
\setlength{\tabcolsep}{5pt}
\begin{tabular}{@{}lll@{}}
\toprule
\textbf{Tool or Mode} & \textbf{Specialist Backend / Scope} & \textbf{Hit Rate (\%)} \\
\midrule
Depth & Depth Anything 3 & $\approx 99.2$ \\
Segment & SAM 3 & $\approx 49.4$ \\
Pose & All Pose requests & $\approx 58.2$ \\
\quad Camera motion & VGGT, successful requests & $100.0$ \\
\quad Object orientation & Orient Anything V2 & $\approx 43.8$ \\
Locate & All Locate requests & $\approx 28.5$ \\
\midrule
Macro average & Four reported top-level paths & $\approx 58.8$ \\
\bottomrule
\end{tabular}
\caption{Observed cache hit rates for spatial-tool requests. Pose is also broken down by its camera-motion and object-orientation backends. The macro average uses the reported Depth, Segment, overall Pose, and Locate rates.}
\label{tab:tool_cache_hit_rates}
\end{table}

\section{Capability Internalization Metric}
\label{sec:cii_details}

\subsection{Evaluation Data and Capability Coverage}

The CII validation suite contains 1,000 held-out examples, balanced across five query types: 200 each for Locate, Segment, Depth, object orientation, and camera motion.
The last two query types correspond to the two modes of the Pose interface.
Each example contains the required image or image pair, a capability-specific request, and a structured reference output.

\subsection{Reference Construction and Evaluation Protocol}

Candidate examples are converted into capability-specific requests, and the corresponding registered specialist tool is executed to produce each structured reference output.
After the independent human verification described below, each retained tool output is serialized in the registered tool's JSON result format.
At evaluation time, the model receives only the image input and an instruction to return the requested tool format; no spatial tool is registered or executed.
We use temperature 0.0, with a maximum of 512 output tokens for Locate, Depth, and both Pose modes, and 4,096 tokens for Segment because polygon serialization is substantially longer.
A deterministic parser extracts the structured JSON response; a response from which no valid JSON object can be recovered receives zero similarity.

\subsection{Independent Human Verification}

Two human experts independently verify each capability-specific request and its tool-produced reference against the image input.
Neither expert can see the other's judgment, and each checks that the request is unambiguous and the specialist-tool output is valid and consistent with the visual input.
We retain a candidate only when both experts approve it; all other candidates are discarded.
This independent agreement rule filters ambiguous requests and unreliable tool outputs before CII evaluation.

\subsection{Similarity Functions and Aggregation}

We evaluate whether a model can reproduce spatial-tool outputs without executing external tools.
For a capability type $\kappa\in\{\mathrm{Locate},\mathrm{Segment},\mathrm{Depth}\}$, let $\hat{o}_i$ be the model prediction and $o_i^*$ the corresponding spatial-tool output.
For each capability, samples are indexed locally by $i=1,\ldots,N_\kappa$.
We define
\[
\mathrm{CII}_{\kappa}
=
\frac{100}{N_{\kappa}}
\sum_{i=1}^{N_{\kappa}}
s_{\kappa}(\hat{o}_i,o_i^*),
\]
where $N_{\kappa}$ is the number of samples for capability $\kappa$ and $s_{\kappa}\in[0,1]$ is a capability-specific similarity function.
Because the Pose interface supports both object orientation and camera motion, we evaluate these two modes separately and macro-average them as defined below.
Unless stated otherwise, a malformed prediction or one missing a required field receives zero similarity.

\paragraph{Locate.}
Let $\hat{\mathcal{B}}_i$ and $\mathcal{B}_i^*$ be the predicted and reference box sets.
We obtain a one-to-one Hungarian matching $\mathcal{M}_i^{\mathrm{box}}$ that maximizes total pairwise IoU and define
\[
s_{\mathrm{Locate}}(\hat{o}_i,o_i^*)
=
\frac{
\sum_{(m,n)\in\mathcal{M}_i^{\mathrm{box}}}
\mathrm{IoU}(\hat{b}_{i,m},b_{i,n}^*)
}{
\max(|\hat{\mathcal{B}}_i|,|\mathcal{B}_i^*|)
}.
\]
This denominator assigns zero contribution to every unmatched prediction or reference box.
If both sets are empty, the similarity is defined as one.

\paragraph{Segment.}
Let $\hat{\mathcal{S}}_i$ and $\mathcal{S}_i^*$ be the predicted and reference instance-mask sets after rasterizing each \texttt{polygon\_2d} at the original image resolution.
Using the IoU-maximizing Hungarian matching $\mathcal{M}_i^{\mathrm{mask}}$, we define
\[
s_{\mathrm{Segment}}(\hat{o}_i,o_i^*)
=
\frac{
\sum_{(j,k)\in\mathcal{M}_i^{\mathrm{mask}}}
\mathrm{IoU}(\hat{m}_{i,j},m_{i,k}^*)
}{
\max(|\hat{\mathcal{S}}_i|,|\mathcal{S}_i^*|)
}.
\]
As in Locate, unmatched instances contribute zero, and two empty sets receive similarity one.

\paragraph{Depth.}
Let $\mathcal{P}_i$ be the nonempty set of reference query points with valid metric-depth outputs, and let $\hat{d}_{i,p}$ and $d_{i,p}^*$ be the predicted and reference depths at point $p$.
Predicted entries are matched to reference queries by \texttt{point\_2d}.
If a required point is missing or duplicated, or if the prediction contains an unrequested point, the sample receives zero similarity; otherwise, using $\varepsilon_{\mathrm{d}}=10^{-6}$ for numerical stability, we compute
\[
\mathrm{AbsRel}_i
=
\frac{1}{|\mathcal{P}_i|}
\sum_{p\in\mathcal{P}_i}
\frac{|\hat{d}_{i,p}-d^*_{i,p}|}{\max(d^*_{i,p},\varepsilon_{\mathrm{d}})}.
\]
Because lower AbsRel indicates better depth prediction, we convert it to a bounded similarity with
\[
s_{\mathrm{Depth}}(\hat{o}_i,o_i^*)
=
\exp(-\mathrm{AbsRel}_i).
\]
Metric depth is evaluated directly without median or scale alignment.

\paragraph{Pose.}
The Pose interface has two modes whose outputs are discrete spatial directions rather than rotation matrices.
For an object-orientation query, we evaluate the canonical \texttt{facing\_direction\_camera} field.
We use the following circular angle mapping:
\[
\begin{array}{ll}
\texttt{front}:0 & \texttt{front-right}:\pi/4 \\
\texttt{right}:\pi/2 & \texttt{back-right}:3\pi/4 \\
\texttt{back}:\pi & \texttt{back-left}:5\pi/4 \\
\texttt{left}:3\pi/2 & \texttt{front-left}:7\pi/4
\end{array}
\]
Let $\hat{\phi}_i$ and $\phi_i^*$ be the predicted and reference angles, and define their circular distance as
\[
\Delta\phi_i
=
\min\left(
|\hat{\phi}_i-\phi_i^*|,
2\pi-|\hat{\phi}_i-\phi_i^*|
\right).
\]
The object-orientation similarity is
\[
s_{\mathrm{obj},i}
=
\max(0,\cos\Delta\phi_i).
\]
The auxiliary \texttt{visible\_side} field and the returned box are not included in this score; localization is evaluated separately by Locate CII.

For a camera-motion query, we evaluate translation and view rotation separately from the signed axes returned by \texttt{position.axes} and \texttt{view\_rotation.dominant\_axis}.
Let $\operatorname{vec}(\alpha)$ be the unit vector denoted by signed axis $\alpha$, where, for example, $\operatorname{vec}(+X)=(1,0,0)$ and $\operatorname{vec}(-Z)=(0,0,-1)$, and define
\[
\operatorname{dir}(\mathcal{A})
=
\frac{\sum_{\alpha\in\mathcal{A}}\operatorname{vec}(\alpha)}
{\left\|\sum_{\alpha\in\mathcal{A}}\operatorname{vec}(\alpha)\right\|_2}.
\]
This mapping is defined only for a nonempty signed-axis set with a nonzero vector sum.
Using $\mathcal{A}_i^{\mathrm{trans}}$ for the translation axes and the singleton $\mathcal{A}_i^{\mathrm{rot}}$ containing the dominant rotation axis, we define
\[
\begin{array}{rcl}
s_{\mathrm{trans},i}
&=&
\max\left(0,\operatorname{dir}(\hat{\mathcal{A}}_i^{\mathrm{trans}})^{\top}
\operatorname{dir}(\mathcal{A}_i^{\mathrm{trans},*})\right),
\\
s_{\mathrm{rot},i}
&=&
\max\left(0,\operatorname{dir}(\hat{\mathcal{A}}_i^{\mathrm{rot}})^{\top}
\operatorname{dir}(\mathcal{A}_i^{\mathrm{rot},*})\right),
\end{array}
\]
and
\[
s_{\mathrm{cam},i}
=
\frac{s_{\mathrm{trans},i}+s_{\mathrm{rot},i}}{2}.
\]
For Pose, an unrecognized direction label or an axis set for which $\operatorname{dir}(\mathcal{A})$ is undefined receives zero similarity.
Finally, to prevent the more frequent query mode from dominating the metric, we compute
\[
\begin{array}{rcl}
\mathrm{CII}_{\mathrm{obj}}
&=&\displaystyle\frac{100}{N_{\mathrm{obj}}}\sum_{i=1}^{N_{\mathrm{obj}}} s_{\mathrm{obj},i},
\\
\mathrm{CII}_{\mathrm{cam}}
&=&\displaystyle\frac{100}{N_{\mathrm{cam}}}\sum_{i=1}^{N_{\mathrm{cam}}} s_{\mathrm{cam},i}.
\end{array}
\]
\[
\mathrm{CII}_{\mathrm{Pose}}
=\frac{\mathrm{CII}_{\mathrm{obj}}+\mathrm{CII}_{\mathrm{cam}}}{2}.
\]
Here, $N_{\mathrm{obj}}>0$ and $N_{\mathrm{cam}}>0$ are the numbers of object-orientation and camera-motion samples, respectively.
We additionally report the two components separately.
The four-capability macro-average reported in the main paper is
\[
\mathrm{CII}_{\mathrm{macro}}
=
\frac{
\mathrm{CII}_{\mathrm{Locate}}
+\mathrm{CII}_{\mathrm{Segment}}
+\mathrm{CII}_{\mathrm{Depth}}
+\mathrm{CII}_{\mathrm{Pose}}
}{4}.
\]

All capability-specific similarities lie in $[0,1]$ and equal one for an exact match; lower values indicate greater divergence from the spatial-tool output.
Every capability-level CII and $\mathrm{CII}_{\mathrm{macro}}$ therefore lie in $[0,100]$, with a higher value indicating stronger internalization.

\section{Additional Experiments}
\label{sec:additional_experiments}

\subsection{Comparison with Direct Fine-Tuning}
\label{sec:direct_finetuning_comparison}

Table~\ref{tab:stage_ablation} compares conventional Direct fine-tuning with inference-time tool augmentation, agentic fine-tuning, and capability internalization on SpatialCLI-Bench using Qwen3-VL-8B-Instruct.
Every variant is evaluated both w/o Tools and w/ Tools.
For Direct variants, w/ Tools evaluation exposes the same spatial-tool interfaces at inference time, even though their training does not include explicit tool-interaction supervision.

\noindent\begin{minipage}{\columnwidth}
\centering
\small
\setlength{\tabcolsep}{3pt}
\begin{tabular*}{\columnwidth}{@{\extracolsep{\fill}}llcclcc@{}}
\toprule
\multirow{2}{*}{\textbf{Stage}} &
\multicolumn{3}{c}{\textbf{Conventional Direct Fine-Tuning}} &
\multicolumn{3}{c}{\textbf{Agentic / SpatialCLI}} \\
\cmidrule(lr){2-4}\cmidrule(lr){5-7}
& \textbf{Variant} & \textbf{w/o Tools} & \textbf{w/ Tools}
& \textbf{Variant} & \textbf{w/o Tools} & \textbf{w/ Tools} \\
\midrule
Inference Only & Initial Model & 35.3 & 66.5 & Initial Model & 35.3 & 66.5 \\
SFT & SFT w/o Tools & 41.3 & 40.3 & SFT w/ Tools & 41.1 & 86.4 \\
RL w/o SFT & RL w/o Tools w/o SFT & 52.7 & 68.2 & RL w/ Tools w/o SFT & 37.2 & 90.5 \\
RL + SFT & RL w/o Tools + SFT & 51.6 & 48.1 & RL w/ Tools + SFT & 40.1 & \textbf{91.3} \\
Internalization & -- & -- & -- & SpatialCLI-8B & \textbf{72.7} & \textbf{91.3} \\
\bottomrule
\end{tabular*}
\captionof{table}{Comparison with conventional Direct fine-tuning on SpatialCLI-Bench. Variant names specify training configurations, while w/o Tools and w/ Tools denote inference settings. SpatialCLI-8B is obtained by Dual-View Capability Internalization after RL w/ Tools + SFT.}
\label{tab:stage_ablation}
\end{minipage}

The Inference Only row uses the same initial checkpoint on both sides; its Inference w/ Tools result denotes enabling the spatial tools without fine-tuning.
Inference-time tool augmentation alone yields a clear improvement w/ Tools, showing that spatial tools effectively compensate for the fine-grained perceptual limitations of the initial model.
SFT w/ Tools learns the basic tool-interaction protocol, while RL w/ Tools w/o SFT optimizes tool selection and termination from task feedback without Cold-Start SFT.
SFT w/ Tools raises the w/ Tools score from 66.5 to 86.4, and RL w/ Tools + SFT further raises it to 91.3.
Although RL w/ Tools w/o SFT eventually reaches 90.5, the training dynamics in Figure~3 show that it does so with substantially longer and more tool-intensive trajectories.
Capability internalization then raises the w/o Tools score from 40.1 to 72.7 while preserving the 91.3 w/ Tools score reached by RL w/ Tools + SFT.
In contrast, SFT w/o Tools, RL w/o Tools w/o SFT, and RL w/o Tools + SFT improve the w/o Tools score but do not acquire a comparably strong tool-use policy.
The results therefore isolate complementary roles for the three stages: immediate perceptual support, stable tool-policy learning, and transfer of tool-supplied capabilities into w/o Tools inference.

\subsection{Tool-Set Ablation}
\label{sec:toolset_ablation}

To isolate the contribution of each spatial-tool group, we evaluate the initial Qwen3-VL-8B-Instruct checkpoint while varying only the registered tool set.
The task inputs, shared agentic prompt template, decoding configuration, and evaluation protocol remain identical across variants.
Each single-tool variant exposes only one of the Locate, Segment, Depth, or Pose interfaces, whereas All Tools exposes all four interfaces.
The evaluated benchmarks, subsets, metrics, and reporting order exactly match those in the main-paper overall results (Table~1).
Table~\ref{tab:toolset_ablation} reports the resulting benchmark scores.

\noindent\begin{minipage}{\columnwidth}
\centering
\small
\renewcommand{\arraystretch}{1.05}
\setlength{\tabcolsep}{2.0pt}
\begin{tabular*}{\columnwidth}{@{\extracolsep{\fill}}lcccccccc@{}}
\toprule
\multicolumn{1}{c}{\multirow{2.5}{*}{\textbf{Model / Tool Set}}} &
\multicolumn{1}{c}{\multirow{2.5}{*}{\shortstack{\textbf{SpatialCLI}\\\textbf{Bench}}}} &
\multicolumn{1}{c}{\multirow{2.5}{*}{\textbf{MindCube}}} &
\multicolumn{2}{c}{\textbf{MMSI}} &
\multicolumn{1}{c}{\multirow{2.5}{*}{\textbf{DA-2K}}} &
\multicolumn{2}{c}{\textbf{BOPASK}} &
\multicolumn{1}{c}{\multirow{2.5}{*}{\textbf{Avg.}}} \\
\cmidrule(lr){4-5}\cmidrule(lr){7-8}
& & & \textbf{Motion-Cam} & \textbf{Pos-Cam-Cam} & & \textbf{Traj.} & \textbf{ObjRrr} & \\
\midrule
Qwen3-VL-8B-Instruct & 35.3 & 29.3 & 27.0 & 25.8 & 68.1 & 25.8 & 13.3 & 35.7 \\
\hspace*{1em}+ Locate Only  & 31.8 & 32.4 & 24.3 & 30.1 & 46.6 & 53.7 & 19.3 & 34.9 \\
\hspace*{1em}+ Segment Only & 39.0 & 26.3 & 27.0 & 28.0 & 57.0 & 43.0 & \textbf{21.0} & 36.3 \\
\hspace*{1em}+ Depth Only   & 40.1 & 27.3 & 24.3 & 26.9 & \textbf{91.6} & 22.7 & 14.1 & 40.6 \\
\hspace*{1em}+ Pose Only    & 50.4 & 44.5 & 37.8 & 38.7 & 52.0 & 4.1 & 1.4 & 37.6 \\
\hspace*{1em}+ All Tools    & \textbf{66.5} & \textbf{47.2} & \textbf{41.9} & \textbf{39.8} & \textbf{91.6} & \textbf{54.3} & 20.6 & \textbf{56.7} \\
\bottomrule
\end{tabular*}
\captionof{table}{Tool-set ablation of the initial Qwen3-VL-8B-Instruct model. Single-tool variants expose only the named interface, while All Tools exposes Locate, Segment, Depth, and Pose. Avg. follows the benchmark-level macro-averaging used in main-paper Table~1. The best result in each column is bolded, including ties.}
\label{tab:toolset_ablation}
\end{minipage}

The single-tool variants exhibit clear capability-aligned specialization.
Depth alone matches All Tools on DA-2K (91.6); among the single-tool settings, Pose performs best on SpatialCLI-Bench, MindCube, and both MMSI subsets, while Locate and Segment are most effective on BOPASK-Trajectory and BOPASK-Object-Rearrangement, respectively.
No single tool is uniformly beneficial across benchmarks.
Jointly exposing all four tools raises the macro-average from 35.7 to 56.7 and achieves the best or tied-best result on six of the seven reported evaluations, supporting the complementarity of the four spatial interfaces.

\subsection{Tool-Use Behavior across Internalization Variants}
\label{sec:internalization_tool_behavior}

To complement the score and output-length ablation in main-paper Table~3, Table~\ref{tab:internalization_tool_behavior} reports the actual tool invocations made by each capability-internalization variant w/ Tools across all evaluated benchmarks.
We report the mean number of tool calls over all samples, the percentage of samples that invoke at least one tool, and the mean number of calls among those samples. To reveal each variant's unconstrained calling tendency, this diagnostic retains the maximum generation length of 40,960 tokens but does not apply the ten-call cap used in the task-performance evaluations.

\noindent\begin{minipage}{\columnwidth}
\centering
\small
\setlength{\tabcolsep}{8pt}
\begin{tabular}{@{}lccc@{}}
\toprule
\textbf{Internalization Variant} &
\textbf{Mean Tool Calls} &
\textbf{Calling Samples (\%)} &
\textbf{Calls per Calling Sample} \\
\midrule
Qwen3-VL-8B-Instruct & 2.00 & 99.3 & 2.02 \\
Final Answer Only & 0.00 & 0.0 & 0.00 \\
CoT + Answer & 0.01 & 0.4 & 1.95 \\
Internalization View Only & 20.70 & 40.3 & 51.35 \\
Tool-Use View Only & 1.28 & 81.0 & 1.58 \\
Full Dual-View & 1.27 & 81.5 & 1.56 \\
\bottomrule
\end{tabular}
\captionof{table}{Tool-use behavior of capability-internalization variants averaged across all benchmarks evaluated in this paper w/ Tools.}
\label{tab:internalization_tool_behavior}
\end{minipage}

The initial model calls tools on 99.3\% of samples, averaging 2.00 calls per sample.
Final Answer Only never calls a tool, while CoT + Answer does so on only 0.4\% of samples.
Internalization View Only calls tools on 40.3\% of samples, but averages 51.35 calls within those samples and 20.70 calls overall.
This excessive number of calls indicates unstable and repetitive tool use rather than effective interaction.
By contrast, Tool-Use View Only and Full Dual-View exhibit nearly identical controlled behavior: they call tools on 81.0\% and 81.5\% of samples and average 1.28 and 1.27 calls overall, respectively.
Thus, dual-view training preserves the tool-use behavior learned from the tool-use view while obtaining the score and compact w/o Tools reasoning benefits reported in main-paper Table~3.

\subsection{Sensitivity to the Tool-Use View Loss Weight}
\label{sec:lambda_sensitivity}

To study the trade-off controlled by the Tool-Use View loss weight in
$\mathcal{L}_{\mathrm{CI}}=\mathcal{L}_{\mathrm{internal}}+\lambda\mathcal{L}_{\mathrm{agentic}}$,
we compare $\lambda\in\{0.2,0.5,1.0,1.5\}$ using Qwen3-VL-8B-Instruct.
All variants use the same training data, optimization configuration, training steps, and checkpoint-selection protocol.
We evaluate the four-capability macro-average CII together with w/o Tools and w/ Tools performance on SpatialCLI-Bench, thereby measuring both capability internalization and tool-use retention.
Table~\ref{tab:lambda_sensitivity} reports the resulting sensitivity analysis.

\noindent\begin{minipage}{\columnwidth}
\centering
\small
\setlength{\tabcolsep}{8.0pt}
\begin{tabular}{llll}
\toprule
\multirow{2}{*}{$\boldsymbol{\lambda}$} &
\multirow{2}{*}{\textbf{CII Macro}} &
\multicolumn{2}{l}{\textbf{SpatialCLI-Bench}} \\
\cmidrule(lr){3-4}
& & \textbf{w/o Tools} & \textbf{w/ Tools} \\
\midrule
0.2 & 60.8 & 72.5 & 89.2 \\
\textbf{0.5 (Ours)} & 60.8 & 72.7 & 91.3 \\
1.0 & 60.5 & 72.3 & 91.4 \\
1.5 & 55.6 & 62.9 & 91.4 \\
\bottomrule
\end{tabular}
\captionof{table}{Sensitivity to the Tool-Use View loss weight $\lambda$ with Qwen3-VL-8B-Instruct. CII Macro averages Locate, Segment, Depth, and Pose, where Pose macro-averages object orientation and camera motion.}
\label{tab:lambda_sensitivity}
\end{minipage}

The results reveal a clear trade-off between capability internalization and tool-policy retention.
Reducing $\lambda$ to 0.2 leaves CII and w/o Tools performance comparable to $\lambda=0.5$, but lowers w/ Tools performance from 91.3 to 89.2, suggesting that underweighting the Tool-Use View weakens retention of the learned tool-use policy.
Increasing $\lambda$ to 1.0 maintains w/ Tools performance at 91.4 while slightly reducing CII and w/o Tools performance; further increasing it to 1.5 substantially lowers CII and w/o Tools performance to 55.6 and 62.9 without an additional w/ Tools gain.
Therefore, $\lambda=0.5$ provides the most balanced trade-off rather than maximizing either training view alone.

\section{Prompt Templates}
\label{sec:prompt_templates}

\subsection{Agentic Tool-Use Prompt Template}

The prompt injects the applicable tool registrations and examples into a shared calling protocol.
The interaction permits at most ten executed tool calls per sample.
After every tool result, SpatialCLI appends \texttt{[Tool-call budget: at most N more tool calls may be made.]}; when the budget reaches zero, it instead appends \texttt{[Tool-call budget exhausted: no more tool calls are allowed. Answer the question immediately using the available information.]}.
In thinking mode, the complete thinking content is retained in the multi-turn history and passed back to the model in subsequent turns.
The complete template is shown in Box~\ref{box:agentic_tool_prompt}.

\begin{spatialcliprompt}[label={box:agentic_tool_prompt},colback=cyan!5!white,colframe=cyan!70!black]{Agentic Tool-Use Prompt Template}
You have the following tools available.

Tool call format:
For each function call, return a JSON object with function name and arguments within <tool_call></tool_call> XML tags:

<tool_call>
{"name": <function-name>, "arguments": <args-json-object>}
</tool_call>

Tool list:
<tools>
[
  {tool_definitions}
]
</tools>

Examples:
{tool_examples}
\end{spatialcliprompt}

\subsection{Turn-Wise Evidence Consolidation Prompt}

This prompt implements the first stage of Progressive Evidence-Grounded Trajectory Verbalization by converting one newly completed tool interaction into an evidence--reasoning unit conditioned on the visual input, task instruction, and previously consolidated units, while withholding the correct answer.
Its rules are designed to preserve every exposed specialist output and explicit visual observation without promoting plans, uncertain guesses, or answer-option analysis to observed facts.
The tool catalog standardizes the interpretation of heterogeneous return fields, while calibrated wording and exact-value preservation retain the uncertainty and numerical fidelity of perceptual estimates.
The complete template is shown in Box~\ref{box:turn_evidence_prompt}.

\begin{spatialcliprompt}[label={box:turn_evidence_prompt},colback=orange!5!white,colframe=orange!75!black]{Turn-Wise Evidence Consolidation Prompt}
You are preparing progressive evidence-grounded capability-internalization training data. Consolidate evidence from one step of a successful visual tool-use trajectory. Given the visual task, the previously consolidated evidence-reasoning units, and exactly one newly completed interaction, verbalize all information exposed by the current tool result and record its logical update.

Be literal, detailed, and evidence-grounded. No correct answer is provided at this stage. Do not solve from an answer key. Put the structured result in the requested XML sections.

Rules:
1. Every perceptual claim must be traceable to the current tool result, an explicit visual observation written in the current original reasoning, or a previous unit. Do not turn a plan, a proposed tool action, answer-option analysis, or an uncertain guess into an observed fact.
2. Exhaustively verbalize every item and every field exposed by the current tool result. Preserve the returned count, every instance, image index, label, bounding box, point, polygon component and vertex, depth value, pose field, motion field, direction, axis, error, and other returned value. Do not shorten, sample, merge, or omit returned information merely because it is lengthy or irrelevant to the task. Express machine structures as readable natural-language statements rather than copying an unexplained machine serialization.
3. Preserve every explicit visual observation and absolute or relative visual-spatial relation from the current original reasoning, even when it is unrelated to the current task. This includes left/right, above/below, front/behind, near/far, overlap, containment, adjacency, occlusion, orientation, size, appearance, and count descriptions.
4. Use calibrated perceptual wording such as "is estimated at", "appears", "approximately", or "the result indicates" where appropriate, because tool outputs are perception estimates. Keep every returned numeric value exactly as given: do not round, alter, normalize again, or invent precision.
5. Explain how the new evidence confirms, revises, or extends prior evidence. Clearly resolve a prior visual estimate when stronger tool evidence supersedes it.
6. Do not repeat unchanged evidence already preserved in a previous unit, but never discard newly returned or newly stated visual information merely because it is task-irrelevant.
7. Empty evidence is valid only when the interaction failed or truly returned no information; preserve the failure or empty-result details themselves.

Return the structured result in this form:
<evidence_unit>
<new_evidence>
Complete readable natural-language statements verbalizing all returned information.
</new_evidence>
<logical_update>
Natural-language updates to the accumulated evidence and reasoning state.
</logical_update>
</evidence_unit>

Tool catalog (field semantics only; not sample evidence):
Perception-tool semantics

The catalog below explains how to interpret fields. It is not evidence for the current sample. A claim is supported only when the corresponding value actually appears in the current tool result or a previous consolidated unit.

1. query_locate — open-vocabulary 2D localization
- Input: a short visible object/category description and optional 1-based image_indices.
- Output: count and zero or more matched instances. Each instance may contain bbox_2d=[x1,y1,x2,y2] and point_2d=[x,y].
- Coordinates use the normalized 0–999 image plane with origin at the top-left: x increases to the right and y increases downward. Smaller y is higher; larger x is farther right.
- point_2d is the center used for geometric comparisons or as input to query_depth. bbox_2d describes 2D extent only.
- This tool does not establish metric depth, segmentation boundaries, object orientation, or camera motion.

2. query_segment — open-vocabulary instance segmentation
- Input: a short visible object/category description and optional 1-based image_indices.
- The exposed segmentation result contains bbox_2d, point_2d, and polygon_2d. point_2d is derived from bbox_2d rather than independently estimated from the mask area.
- bbox_2d is the bounding box associated with that segmentation instance/mask and normally encloses all returned polygon components. It is emitted as a separate result field; it is not recomputed as the mathematically tight bounding box of the simplified polygon vertices, so small boundary differences are expected.
- point_2d is the rounded, clamped center of bbox_2d: [(x1+x2)/2,(y1+y2)/2]. It is not the area centroid of the mask or polygon, and for a non-convex or disconnected mask it may lie outside the occupied mask region.
- bbox_2d, point_2d, and polygon_2d all use the normalized 0–999 image coordinates. polygon_2d is a list of simplified boundary components tracing the occupied image region; multiple disconnected components may belong to one mask.
- polygon_2d supports boundary, shape, overlap, containment, free-space, and placement reasoning. Neither the bbox center nor a polygon is a grasp pose, 3D cuboid, metric depth, or object orientation.

3. query_depth — metric camera-axis depth at image points
- Input: one or more normalized 0–999 points and optional 1-based image_indices.
- Output: one result per point with point_2d and depth_m.
- depth_m is estimated camera-axis distance in meters. Smaller depth_m means closer to the camera; larger means farther.
- Compare the returned numeric values explicitly when relative depth is relevant. Do not confuse depth_m with image-plane y, Euclidean object-to-object distance, or object size.
- Depth is an estimate and only supports claims at the queried points/images.

4. query_pose — object orientation or cross-view camera motion
- Object mode input: a short object description, optionally with image_indices. Output may include image_index, bbox_2d, visible_side, and facing_direction_camera.
- visible_side names the object's side seen by the camera. facing_direction_camera describes where the object's front points in camera coordinates; these are related but not interchangeable. For facing direction, front means deeper into the image, back means toward the camera, and left/right are the viewer's image sides.
- Camera-motion mode input uses query='camera motion'. Each result is directional from from_image to to_image.
- position.direction describes translation of the camera position. view_rotation.direction describes where the camera view turns. Never substitute one for the other.
- Signed camera axes are +X right, +Y up, and -Z forward. Use ordinary direction strings for ordinary-language questions and signed axes only when the task explicitly asks about axes.

For every tool: respect image_index/image_indices, zero-result and failure states. Never treat a tool description, an input query, or the model's pre-call guess as observed evidence.

<visual_inputs>
{VISUAL_INPUTS}
</visual_inputs>

<task_instruction>
{TASK_INSTRUCTION}
</task_instruction>

<previous_evidence_units>
{PREVIOUS_EVIDENCE_UNITS}
</previous_evidence_units>

<current_reasoning>
{CURRENT_REASONING}
</current_reasoning>

<current_tool_call>
{CURRENT_TOOL_CALL}
</current_tool_call>

<current_tool_result>
{CURRENT_TOOL_RESULT}
</current_tool_result>
\end{spatialcliprompt}

\subsection{Global Trajectory Verbalization Prompt}

This prompt implements the second stage by converting the ordered evidence--reasoning units into a single tool-free perceptual reasoning target conditioned on the correct final answer.
The answer is introduced only after turn-wise evidence consolidation, so it can guide the organization of already extracted evidence without influencing its collection.
The rules merge repeated or superseded statements in dependency order, prohibit unsupported additions and construction-process leakage, and require exact preservation of the final answer.
The complete template is shown in Box~\ref{box:global_verbalization_prompt}.

\begin{spatialcliprompt}[label={box:global_verbalization_prompt},colback=green!3!white,colframe=green!50!black]{Global Trajectory Verbalization Prompt}
You are preparing the final trajectory-verbalization target for capability internalization. Convert ordered, grounded visual evidence into a tool-free reasoning target for vision-language SFT. Given the visual task, ordered evidence-reasoning units, and the correct final answer, construct a perceptual reasoning chain explaining why the evidence supports that answer.

Put the structured result in the requested XML sections. The target must read as direct visual reasoning and must not mention tools, calls, observations, evidence-unit boundaries, confidence scores, prompts, or this conversion process.

Rules:
1. Use only entities, attributes, values, and relations present in the evidence units.
2. Merge evidence in dependency order and remove repeated or superseded statements, but do not discard an explicit visual or spatial fact merely because it is unnecessary for choosing the answer.
3. Preserve the concrete visual-spatial descriptions carried by the evidence units and include every essential inference needed to support all clauses of the answer.
4. Write as direct visual reasoning. Never mention tools, calls, results, evidence units, confidence scores, or this instruction.
5. Do not invent entities, attributes, measurements, or relations.
6. Put reasoning and answer in separate XML sections. Copy the provided final answer exactly.

Return the structured result in this form:
<internalization_target>
<reasoning_chain>
Complete, detailed, tool-free perceptual reasoning.
</reasoning_chain>
<final_answer>
Exact provided final answer.
</final_answer>
</internalization_target>

<visual_inputs>
{VISUAL_INPUTS}
</visual_inputs>

<task_instruction>
{TASK_INSTRUCTION}
</task_instruction>

<evidence_reasoning_units>
{EVIDENCE_REASONING_UNITS}
</evidence_reasoning_units>

<correct_final_answer>
{CORRECT_FINAL_ANSWER}
</correct_final_answer>
\end{spatialcliprompt}

% BEGIN GENERATED CASE STUDIES
% AUTO-GENERATED by scripts/generate_case_study.py. DO NOT EDIT.
\section{Case Studies}

We present 2 w/o Tools versus w/ Tools comparisons for Qwen3.5-397B-A17B. The first examines depth comparison and camera view rotation, whereas the second examines instance-level bounding boxes and camera translation. Together, they show how the agent decomposes a question into grounding, measurement, and cross-checking steps and uses tool evidence to revise unreliable visual judgments. Section H.3 separately compares two correct SpatialCLI-8B results after capability internalization: direct answering and tool-augmented answering.

\subsection{Case 1: Two-Image Camera Rotation and Depth}

The task input and ground truth are shown in Box H.1. In the comparison below, the direct answer infers that the beige receptacle is closer from apparent size and image position. It therefore selects C despite correctly identifying the rightward view rotation. The tool-augmented agent first grounds the two receptacles, queries two-image camera motion, and then passes the grounded centers to Depth. The measured depths, 0.928 m for the beige receptacle and 0.776 m for the blue one, overturn the initial visual guess. Combining the blue receptacle's right-side location with the rightward rotation returned by Pose yields D. The trajectory follows a ground--measure--cross-check--verify logic.

\begin{casestep}{gray}{Box H.1: Task (SpatialCLI-Bench ID 269)}
\begin{center}
\begin{minipage}[t]{0.48\linewidth}
\centering
\textbf{Image 1}\par\smallskip
\includegraphics[width=\linewidth]{Figures/case_study_qwen35_397b_id269_image1.png}
\end{minipage}\hfill
\begin{minipage}[t]{0.48\linewidth}
\centering
\textbf{Image 2}\par\smallskip
\includegraphics[width=\linewidth]{Figures/case_study_qwen35_397b_id269_image2.png}
\end{minipage}
\end{center}
\medskip
\raggedright
In the first image, there are two large plastic waste receptacles in the foreground: a beige one with a black bag and a blue one with a clear bag. Compare their distances from the camera to find the closer one. In the first image, is this closer receptacle located on the left or the right? Finally, when moving from the first image to the second image, what is the dominant direction of the camera's view rotation?\\Options:\\A: The blue receptacle is closer; it is located on the right; the camera rotated to the left.\\B: The beige receptacle is closer; it is located on the left; the camera rotated to the left.\\C: The beige receptacle is closer; it is located on the left; the camera rotated to the right.\\D: The blue receptacle is closer; it is located on the right; the camera rotated to the right.\\E: The blue receptacle is closer; it is located on the left; the camera rotated to the right.\\F: The beige receptacle is closer; it is located on the right; the camera rotated to the left.\\Before choosing, briefly state the relevant facts and verify every clause of the selected option against all evidence available during solving. For any depth observations, smaller values mean closer; do not retain an earlier visual guess after receiving stronger evidence. Please show your choice in the answer field with only the choice letter, e.g., \{"answer": "C"\}.
\end{casestep}

\begin{casestep}{gray}{Ground Truth}
Correct option: \textbf{D}.
\end{casestep}

\setlength{\columnsep}{10pt}
\begin{paracol}{2}
\switchcolumn[0]
\begin{casestep}{red}{w/o Tools (Incorrect)}
\textcolor{purple!75!black}{\texttt{\textless{}think\textgreater{}}}\par
\raggedright
The user wants me to analyze two images and answer a multi-part question.
\par
**Part 1: Compare distances of the two large plastic waste receptacles in the first image.**\\-   **Receptacle 1:** Beige/tan bin with a black bag. It's on the left side of the foreground.\\-   **Receptacle 2:** Blue bin with a clear/white bag. It's on the right side of the foreground.\\-   **Comparison:** Looking at the perspective and size, the beige bin appears larger and lower down in the frame relative to the blue bin next to it (though they are similar in height, the base of the beige bin is lower). Wait, let's look closer. The beige bin is clearly in front of the blue bin behind it (the tall blue one with the lid). But there is another blue bin to the right. Let's re-evaluate.\\-   Actually, looking at the first image:\\    -   There is a beige bin on the left.\\    -   Behind it is a tall blue recycling bin with a lid.\\    -   To the right of the beige bin is a small black cart.\\    -   To the right of the black cart is another blue bin with a clear liner.\\-   The prompt asks about "two large plastic waste receptacles in the foreground: a beige one with a black bag and a blue one with a clear bag."\\    -   Beige bin: Left foreground.\\    -   Blue bin (with clear bag): Right foreground.\\-   Which is closer? The beige bin's base is lower in the frame than the blue bin's base. The beige bin obscures part of the view towards the back left. The blue bin on the right is also quite close. Let's look at the perspective. The beige bin seems slightly closer to the camera lens because it feels more "in your face" and its bottom edge is lower. However, let's look at the second image.\\-   The second image shows *only* the blue bin with the recycling symbol. This is the bin that was on the right in the first image (blue bin, clear bag). Wait, let me check the details.\\    -   Image 1 Blue Bin (right): Blue, rectangular, clear/white bag.\\    -   Image 2 Blue Bin: Blue, rectangular, white logo (recycling symbol), clear/white bag. This matches the bin on the right in Image 1.\\-   In Image 2, this blue bin is now on the left side of the frame and very prominent. The camera has moved.\\-   Let's re-evaluate depth in Image 1. The beige bin is on the left. The blue bin is on the right. The beige bin looks slightly closer because it is larger in apparent scale relative to what a standard bin looks like, and its bottom is lower. BUT, let's look at the options.\\    -   Option A: Blue closer, right, rotate left.\\    -   Option B: Beige closer, left, rotate left.\\    -   Option C: Beige closer, left, rotate right.\\    -   Option D: Blue closer, right, rotate right.\\    -   Option E: Blue closer, left... (Blue is on the right in img 1).\\    -   Option F: Beige closer, right... (Beige is on the left in img 1).
\par
Let's look really closely at Image 1.\\-   The beige bin is clearly the closest object. It is large and in the immediate foreground.\\-   The blue bin on the right is also in the foreground but set slightly further back or just to the side. Wait, actually, looking at the floor tiles, the beige bin is positioned further "down" (closer to bottom edge) than the blue bin on the right. Therefore, the beige bin is closer.\\-   So, "The beige receptacle is closer". And "it is located on the left". This eliminates A, D, E, F.\\-   We are left with B and C.\\    -   B: Camera rotated to the left.\\    -   C: Camera rotated to the right.
\par
**Part 2: Determine camera rotation.**\\-   **Image 1 View:** Shows beige bin (left), black cart (center), blue bin (right). Background shows desks and cabinets.\\-   **Image 2 View:** Shows the blue bin (from the right side of Image 1) now on the *left* side of the frame. To its right are white cabinets.\\-   In Image 1, the blue bin is on the right. In Image 2, we are looking directly at the front/side of that same blue bin, and the objects that were to its right (the white cabinets/shelves) are now visible.\\-   Wait, let's look at the background of Image 1. To the right of the blue bin is a desk area.\\-   In Image 2, to the right of the blue bin are white cabinets. Where were these white cabinets in Image 1? They are likely the white structure behind/next to the blue bin. In Image 1, to the right of the blue bin, we see a desk. The white cabinets in Image 2 look like storage units.\\-   Let's re-orient.\\    -   Image 1: Left = Beige bin. Center = Black cart. Right = Blue bin. Far Right = Desk area.\\    -   Image 2: Left = Blue bin. Right = White cabinets.\\-   If I am looking at Image 1, and I want to see what is to the *right* of the scene (where the blue bin is), I would turn the camera to the **right**.\\-   Let's trace the objects.\\    -   Object X: Blue bin with clear bag. In Img 1, it's on the right. In Img 2, it's on the left/center. This implies the camera moved to the right, causing the object to shift left in the frame. Or the camera rotated right.\\    -   If I rotate right, objects on the right move towards the center/left. Yes.\\    -   Also, new objects appear on the right side. In Image 2, we see white cabinets on the right. In Image 1, looking past the blue bin, we see a desk area. Wait, let's look at the white cabinets in Image 1. Behind the black cart and blue bin is a white wall/pillar. To the right of that is a desk.\\    -   Let's look at Image 2 again. The white cabinets are tall, with handles. In Image 1, can we see these? Ah, look at the top right of Image 1. There is a desk. Underneath/behind it might be cabinets.\\    -   Actually, look at the transition.\\    -   Image 1: Beige bin (Left), Black Cart (Middle), Blue Bin (Right).\\    -   Image 2: Blue Bin (Left), White Cabinets (Right).\\    -   The camera has shifted focus from the group (Beige, Cart, Blue) to just the Blue bin and what is next to it.\\    -   The Blue bin was on the right. Now it is on the left. This means the camera panned/rotated to the **right**. By turning right, the rightmost object (Blue bin) moves towards the center/left of the view, and we see what is further to the right (the white cabinets).\\    -   So, rotation is to the right.
\par
**Conclusion:**\\-   Closer receptacle: Beige (based on position lower in frame and general perspective).\\-   Location: Left.\\-   Rotation: Right.
\par
This matches Option C.
\par
Let's double check "Closer".\\-   In Image 1, the beige bin is very large. The blue bin on the right is also large.\\-   Let's look at the floor. The bottom of the beige bin is cut off by the bottom edge of the photo? No, it's fully visible. The bottom of the blue bin is also visible.\\-   The beige bin's bottom is lower in the y-axis than the blue bin's bottom. In perspective projection, lower on the ground plane usually means closer.\\-   Therefore, Beige is closer.\\-   Beige is on the left.\\-   Camera moves from seeing [Beige, Cart, Blue] to seeing [Blue, Cabinets].\\-   This is a pan/rotation to the right.
\par
So, Option C: "The beige receptacle is closer; it is located on the left; the camera rotated to the right."
\par
Let's verify Option B just in case. "Camera rotated to the left". If I rotated left, I would see more of the beige bin and whatever is to its left (the office chair area). I would lose the blue bin on the right. But Image 2 focuses on the blue bin. So it must be a rotation to the right.
\par
Therefore, C is the correct answer.
\par
\textcolor{purple!75!black}{\texttt{\textless{}/think\textgreater{}}}
\par\medskip
\raggedright
The beige receptacle is visibly larger and positioned lower in the frame than the blue receptacle on the right, indicating it is closer to the camera. It is located on the left side of the first image. Comparing the two images, the first image shows the beige bin, a black cart, and a blue bin. The second image centers on the blue bin (which was on the right in the first image) and reveals white cabinets to its right. For the blue bin to shift from the right side of the view to the left side, and for new objects (cabinets) to appear on the right, the camera must have rotated to the right.
\par
\{"answer": "C"\}
\end{casestep}
\switchcolumn
\begin{casestep}{green}{w/ Tools (Correct)}
\textcolor{purple!75!black}{\texttt{\textless{}think\textgreater{}}}\par
\raggedright
The user wants me to analyze two images of an office setting, focusing on waste receptacles and camera movement.
\par
**Step 1: Identify the objects in the first image.**\\- There is a beige trash can with a black bag on the left side of the foreground.\\- There is a blue recycling bin with a clear bag on the right side of the foreground.\\- Behind the beige can, there is another blue bin (taller).
\par
**Step 2: Compare distances of the beige and blue receptacles in the first image.**\\- I need to determine which one is closer to the camera.\\- Visually, the beige can appears larger and lower in the frame, suggesting it might be closer. However, the blue bin on the right is also quite large.\\- Let's use `query\_locate` to get their bounding boxes and centers, then `query\_depth` to check their actual distances.
\par
**Step 3: Determine the location (left/right) of the closer receptacle.**\\- Once I know which is closer, I need to state if it's on the left or right.\\- Based on visual inspection:\\    - Beige can: Left side.\\    - Blue bin: Right side.
\par
**Step 4: Determine the camera's view rotation from image 1 to image 2.**\\- Image 1 shows the beige can and blue bin side-by-side with a cabinet behind them.\\- Image 2 shows primarily the blue bin (with the recycling symbol visible now) and white cabinets to its right. The beige can is no longer visible or is out of frame to the left.\\- This suggests the camera turned to the right to focus on the blue bin and the cabinets next to it. Or did it turn left?\\- Let's look at the background. In image 1, to the right of the blue bin, there are desks and shelves. In image 2, to the right of the blue bin, there are white cabinets. Wait, let's look closer.\\- In image 1, the blue bin is on the right. To its right, further back, is a desk area.\\- In image 2, the blue bin is on the left side of the frame. To its right are white cabinets.\\- Let's re-evaluate the scene.\\- Image 1: Beige bin (left), Blue bin (right). Behind them is a partition/cabinet edge.\\- Image 2: Blue bin is now on the left side of the image. White cabinets are to its right.\\- If the blue bin was on the right in image 1 and is on the left in image 2, the camera must have panned/rotated to the *right*. Imagine standing facing the bins. If you turn right, objects on your right move towards the center/left of your view.\\- Let's verify with `query\_pose`.
\par
**Plan:**\\1. Locate the beige trash can and the blue trash can in the first image.\\2. Get depth for both to confirm which is closer.\\3. Confirm left/right position based on location.\\4. Use `query\_pose` with "camera motion" to determine the rotation direction from image 1 to image 2.
\par
**Execution:**\\1. `query\_locate` for "beige trash can" and "blue trash can" in image 1.\\2. `query\_depth` on their centers.\\3. `query\_pose` for "camera motion" between image 1 and 2.
\par
Let's refine the queries.\\- Query 1: "beige trash can with black bag" in image 1.\\- Query 2: "blue trash can with clear bag" in image 1.\\- Query 3: "camera motion" from image 1 to 2.
\par
Wait, looking at image 2, the blue bin has a recycling symbol. In image 1, the blue bin on the right has a clear bag. Is it the same bin? Yes, likely. The bin on the left in image 2 looks like the bin on the right in image 1.\\In image 1, the blue bin is to the right of the beige bin.\\In image 2, the blue bin is to the left of the white cabinets.\\This implies a rotation to the right. If I look straight at the gap between them, then turn right, the blue bin moves to the left of my field of view.
\par
Let's get the data to be sure.
\par
\textcolor{purple!75!black}{\texttt{\textless{}/think\textgreater{}}}
\par\medskip
\textcolor{blue!75!black}{\texttt{\textless{}tool\_call\textgreater{}}}\par
{\ttfamily\raggedright\obeyspaces
\{\\\ \ "name":\allowbreak{} "query\_locate",\allowbreak{}\\\ \ "arguments":\allowbreak{} \{\\\ \ \ \ "query":\allowbreak{} "beige trash can with black bag",\allowbreak{}\\\ \ \ \ "image\_indices":\allowbreak{} [\\\ \ \ \ \ \ 1\\\ \ \ \ ]\allowbreak{}\\\ \ \}\allowbreak{}\\\}\allowbreak{}
}
\par
\textcolor{blue!75!black}{\texttt{\textless{}/tool\_call\textgreater{}}}
\par\medskip
\textcolor{blue!75!black}{\texttt{\textless{}tool\_call\textgreater{}}}\par
{\ttfamily\raggedright\obeyspaces
\{\\\ \ "name":\allowbreak{} "query\_locate",\allowbreak{}\\\ \ "arguments":\allowbreak{} \{\\\ \ \ \ "query":\allowbreak{} "blue trash can with clear bag",\allowbreak{}\\\ \ \ \ "image\_indices":\allowbreak{} [\\\ \ \ \ \ \ 1\\\ \ \ \ ]\allowbreak{}\\\ \ \}\allowbreak{}\\\}\allowbreak{}
}
\par
\textcolor{blue!75!black}{\texttt{\textless{}/tool\_call\textgreater{}}}
\par\medskip
\textcolor{blue!75!black}{\texttt{\textless{}tool\_call\textgreater{}}}\par
{\ttfamily\raggedright\obeyspaces
\{\\\ \ "name":\allowbreak{} "query\_pose",\allowbreak{}\\\ \ "arguments":\allowbreak{} \{\\\ \ \ \ "query":\allowbreak{} "camera motion",\allowbreak{}\\\ \ \ \ "image\_indices":\allowbreak{} [\\\ \ \ \ \ \ 1,\allowbreak{}\\\ \ \ \ \ \ 2\\\ \ \ \ ]\allowbreak{}\\\ \ \}\allowbreak{}\\\}\allowbreak{}
}
\par
\textcolor{blue!75!black}{\texttt{\textless{}/tool\_call\textgreater{}}}
\par\medskip
\textcolor{green!55!black}{\texttt{\textless{}tool\_response\textgreater{}}}\par
{\ttfamily\raggedright\obeyspaces
\{"count":\allowbreak{}1,\allowbreak{}"result":\allowbreak{}[\{"bbox\_2d":\allowbreak{}[203,\allowbreak{}176,\allowbreak{}418,\allowbreak{}872]\allowbreak{},\allowbreak{}"point\_2d":\allowbreak{}[310,\allowbreak{}524]\allowbreak{}\}\allowbreak{}]\allowbreak{}\}\allowbreak{}
\par
[Tool-call budget:\allowbreak{} at most 4 more tool calls may be made.]\allowbreak{}
}
\par
\textcolor{green!55!black}{\texttt{\textless{}/tool\_response\textgreater{}}}
\par\medskip
\textcolor{green!55!black}{\texttt{\textless{}tool\_response\textgreater{}}}\par
{\ttfamily\raggedright\obeyspaces
\{"count":\allowbreak{}1,\allowbreak{}"result":\allowbreak{}[\{"bbox\_2d":\allowbreak{}[606,\allowbreak{}289,\allowbreak{}940,\allowbreak{}997]\allowbreak{},\allowbreak{}"point\_2d":\allowbreak{}[773,\allowbreak{}643]\allowbreak{}\}\allowbreak{}]\allowbreak{}\}\allowbreak{}
\par
[Tool-call budget:\allowbreak{} at most 3 more tool calls may be made.]\allowbreak{}
}
\par
\textcolor{green!55!black}{\texttt{\textless{}/tool\_response\textgreater{}}}
\par\medskip
\textcolor{green!55!black}{\texttt{\textless{}tool\_response\textgreater{}}}\par
{\ttfamily\raggedright\obeyspaces
\{"result":\allowbreak{}[\{"from\_image":\allowbreak{}1,\allowbreak{}"to\_image":\allowbreak{}2,\allowbreak{}"position":\allowbreak{}\{"direction":\allowbreak{}"forward-right",\allowbreak{}"axes":\allowbreak{}["+X",\allowbreak{}"+Y",\allowbreak{}"-Z"]\allowbreak{}\}\allowbreak{},\allowbreak{}"view\_rotation":\allowbreak{}\{"direction":\allowbreak{}"right",\allowbreak{}"axes":\allowbreak{}["+X",\allowbreak{}"+Y",\allowbreak{}"+Z"]\allowbreak{},\allowbreak{}"dominant\_axis":\allowbreak{}"+Y"\}\allowbreak{}\}\allowbreak{}]\allowbreak{}\}\allowbreak{}
\par
[Tool-call budget:\allowbreak{} at most 2 more tool calls may be made.]\allowbreak{}
}
\par
\textcolor{green!55!black}{\texttt{\textless{}/tool\_response\textgreater{}}}
\par\medskip
\textcolor{purple!75!black}{\texttt{\textless{}think\textgreater{}}}\par
\raggedright
Now I have the locations of both trash cans in image 1:\\- Beige trash can: point\_2d [310, 524]\\- Blue trash can: point\_2d [773, 643]
\par
I need to query the depth at these points to determine which is closer to the camera. Let me also note that the beige one is on the left (x=310) and the blue one is on the right (x=773) in the image.
\par
The camera motion result shows view\_rotation direction is "right" with dominant\_axis "+Y".
\par
Let me query the depths to confirm which receptacle is closer.
\par
\textcolor{purple!75!black}{\texttt{\textless{}/think\textgreater{}}}
\par\medskip
\textcolor{blue!75!black}{\texttt{\textless{}tool\_call\textgreater{}}}\par
{\ttfamily\raggedright\obeyspaces
\{\\\ \ "name":\allowbreak{} "query\_depth",\allowbreak{}\\\ \ "arguments":\allowbreak{} \{\\\ \ \ \ "points":\allowbreak{} [\\\ \ \ \ \ \ [\\\ \ \ \ \ \ \ \ 310,\allowbreak{}\\\ \ \ \ \ \ \ \ 524\\\ \ \ \ \ \ ]\allowbreak{},\allowbreak{}\\\ \ \ \ \ \ [\\\ \ \ \ \ \ \ \ 773,\allowbreak{}\\\ \ \ \ \ \ \ \ 643\\\ \ \ \ \ \ ]\allowbreak{}\\\ \ \ \ ]\allowbreak{},\allowbreak{}\\\ \ \ \ "image\_indices":\allowbreak{} [\\\ \ \ \ \ \ 1\\\ \ \ \ ]\allowbreak{}\\\ \ \}\allowbreak{}\\\}\allowbreak{}
}
\par
\textcolor{blue!75!black}{\texttt{\textless{}/tool\_call\textgreater{}}}
\par\medskip
\textcolor{green!55!black}{\texttt{\textless{}tool\_response\textgreater{}}}\par
{\ttfamily\raggedright\obeyspaces
\{"result":\allowbreak{}[\{"point\_2d":\allowbreak{}[310.0,\allowbreak{}524.0]\allowbreak{},\allowbreak{}"depth\_m":\allowbreak{}0.928\}\allowbreak{},\allowbreak{}\{"point\_2d":\allowbreak{}[773.0,\allowbreak{}643.0]\allowbreak{},\allowbreak{}"depth\_m":\allowbreak{}0.776\}\allowbreak{}]\allowbreak{}\}\allowbreak{}
\par
[Tool-call budget:\allowbreak{} at most 1 more tool call may be made.]\allowbreak{}
}
\par
\textcolor{green!55!black}{\texttt{\textless{}/tool\_response\textgreater{}}}
\par\medskip
\raggedright
Based on the tool results:
\par
1.  **Distances**: The depth query shows the beige receptacle (point [310, 524]) is at **0.928m** and the blue receptacle (point [773, 643]) is at **0.776m**. Since 0.776 \textless{} 0.928, the **blue receptacle is closer**.\\2.  **Location**: In the first image, the beige can is at x=310 (left side) and the blue can is at x=773 (right side). Therefore, the closer receptacle (blue) is located on the **right**.\\3.  **Camera Motion**: The pose query for camera motion from image 1 to 2 indicates a `view\_rotation` direction of **right**.
\par
Matching these facts to the options:\\*   Closer: Blue\\*   Location: Right\\*   Rotation: Right
\par
This corresponds to Option D.
\par
\{"answer": "D"\}
\end{casestep}
\end{paracol}

% AUTO-GENERATED by scripts/generate_case_study.py. DO NOT EDIT.
\subsection{Case 2: Two-Image Camera Translation and Instance Segmentation}

The task input and ground truth are shown in Box H.2. In the comparison below, the direct answer approximately identifies the target sandal and its maximum vertical coordinate, but interprets the scale change in the second image as backward camera motion. It consequently predicts backward-left and selects C. The tool-augmented agent first obtains a coarse location and the camera motion from Locate and Pose. When Locate merges the pair into one box, the agent switches to Segment to recover two instances and compares their maximum horizontal and vertical coordinates. The instance-level result supports 856, while Pose returns forward-left, yielding E. This trajectory follows a coarse-grounding--granularity-check--instance-refinement--joint-decision logic.

\begin{casestep}{gray}{Box H.2: Task (SpatialCLI-Bench ID 323)}
\begin{center}
\begin{minipage}[t]{0.48\linewidth}
\centering
\textbf{Image 1}\par\smallskip
\includegraphics[width=\linewidth]{Figures/case_study_qwen35_397b_id323_image1.jpg}
\end{minipage}\hfill
\begin{minipage}[t]{0.48\linewidth}
\centering
\textbf{Image 2}\par\smallskip
\includegraphics[width=\linewidth]{Figures/case_study_qwen35_397b_id323_image2.jpg}
\end{minipage}
\end{center}
\medskip
\raggedright
Consider the two white sandals with pearl straps in the first image. First, determine which of these two sandals has a bounding box that extends further to the right (has the larger maximum x-coordinate). What is the maximum y-coordinate of that same sandal's bounding box, and what is the direction of the camera's translation from the first image to the second?\\Options:\\A: Maximum y-coordinate: 747; Camera translation: backward-left\\B: Maximum y-coordinate: 747; Camera translation: forward-left\\C: Maximum y-coordinate: 856; Camera translation: backward-left\\D: Maximum y-coordinate: 747; Camera translation: forward-right\\E: Maximum y-coordinate: 856; Camera translation: forward-left\\F: Maximum y-coordinate: 856; Camera translation: forward-right\\Before choosing, briefly state the relevant facts and verify every clause of the selected option against all evidence available during solving. For any depth observations, smaller values mean closer; do not retain an earlier visual guess after receiving stronger evidence. Please show your choice in the answer field with only the choice letter, e.g., \{"answer": "C"\}.
\end{casestep}

\begin{casestep}{gray}{Ground Truth}
Correct option: \textbf{E}.
\end{casestep}

\setlength{\columnsep}{10pt}
\begin{paracol}{2}
\switchcolumn[0]
\begin{casestep}{red}{w/o Tools (Incorrect)}
\textcolor{purple!75!black}{\texttt{\textless{}think\textgreater{}}}\par
\raggedright
The user wants me to identify a specific sandal based on its bounding box and then determine camera movement.
\par
**1. Identify the target sandal:**\\- The prompt asks to consider the "two white sandals with pearl straps in the first image".\\- I need to find the bounding boxes for these two sandals.\\- Sandal 1 (left/back): This is the one further back, slightly to the left. Its strap has pearls.\\- Sandal 2 (right/front): This is the one closer to the front right. Its strap also has pearls. Wait, let's look closer.\\- In Image 1:\\    - There is a sandal in the foreground [143, 508, 765, 856]. This is the main white sandal facing somewhat left. It has a pearl strap.\\    - There is another sandal behind it/to the left [159, 492, 715, 747]. This one is partially visible, angled differently. Wait, let's re-evaluate.\\    - Let's look at the structure.\\    - One sandal is clearly in the front: [143, 508, 765, 856]. This is the right-most sandal of the pair. It has a square toe and a pearl strap.\\    - The other sandal is behind it: [159, 492, 715, 747]. This one is to the left and slightly behind.\\- The question asks: "determine which of these two sandals has a bounding box that extends further to the right (has the larger maximum x-coordinate)."\\    - Sandal 1 (front/right): Box is roughly [143, 508, 765, 856]. Max x is \textasciitilde{}765.\\    - Sandal 2 (back/left): Box is roughly [159, 492, 715, 747]. Max x is \textasciitilde{}715.\\    - Therefore, the sandal with the larger maximum x-coordinate is the front/right one: [143, 508, 765, 856].
\par
**2. Find the maximum y-coordinate of that sandal:**\\- The bounding box for the identified sandal is [143, 508, 765, 856].\\- The coordinates are usually [y\_min, x\_min, y\_max, x\_max] or [x\_min, y\_min, x\_max, y\_max]. Let's assume standard object detection format [ymin, xmin, ymax, xmax] or similar.\\- Let's look at the coordinates provided in the thought process.\\    - If the box is [143, 508, 765, 856], let's interpret.\\    - y\_min = 143? No, that's too high up. The shoe is lower down.\\    - Let's look at the image dimensions. It's a vertical phone photo.\\    - The shoes are at the bottom. So y-coordinates should be large (near the bottom).\\    - x-coordinates span the width.\\    - Let's re-examine the boxes.\\    - Sandal 1 (front): ymin \textasciitilde{} 508, xmin \textasciitilde{} 143, ymax \textasciitilde{} 856, xmax \textasciitilde{} 765.\\        - ymin=508 (mid-height), xmin=143 (left side), ymax=856 (bottom), xmax=765 (right side). This makes sense. The heel is around y=500-600. The toe is at the bottom y=856.\\    - Sandal 2 (back): ymin \textasciitilde{} 492, xmin \textasciitilde{} 159, ymax \textasciitilde{} 747, xmax \textasciitilde{} 715.\\        - ymin=492, xmin=159, ymax=747, xmax=715.\\    - Comparing max x: Sandal 1 (765) \textgreater{} Sandal 2 (715). So the target sandal is the front one.\\    - The question asks for the "maximum y-coordinate of that same sandal's bounding box".\\    - For the front sandal [143, 508, 765, 856], the max y is 856. Wait, let me check the coordinate order. Usually it's [ymin, xmin, ymax, xmax] or [xmin, ymin, xmax, ymax].\\    - Let's look at the options. Options have max y as 747 or 856.\\    - If the box is [143, 508, 765, 856], and assuming [ymin, xmin, ymax, xmax], then max y is 856.\\    - If the box is [159, 492, 715, 747], max y is 747.\\    - Since I identified the sandal extending further right as the front one, its max y should be 856.\\    - Let's double check which one extends further right.\\        - Front sandal: The heel strap goes back. The toe is forward. The rightmost part is likely the heel/strap area or the side of the foot. In image 1, the sandal in the foreground (closest to camera) has its heel towards the right side of the frame. The other sandal is behind it.\\        - Actually, looking closely at Image 1:\\            - There is a sandal lying flat: [159, 492, 715, 747]. This looks like the left shoe of the pair.\\            - There is a sandal standing/upright: [143, 508, 765, 856]. This looks like the right shoe of the pair.\\            - The upright sandal (right shoe) is closer to the right edge. Its bounding box x\_max would be larger.\\            - The flat sandal (left shoe) is further left.\\            - So the target is the upright sandal.\\            - Its max y-coordinate is the bottom of the shoe. In image 1, the bottom of the sole is quite low. 856 seems correct for the bottom. 747 would be higher up, corresponding to the other sandal.\\            - So, Max y-coordinate = 856.
\par
**3. Determine camera translation:**\\- Compare Image 1 and Image 2.\\- Image 1: Close-up of the shoes. We see the headboard clearly. The shoes are large in the frame.\\- Image 2: Wide shot. The shoes are further away. We see the whole room, clothes rack, towels.\\- The camera has moved *backward* (away from the shoes) to capture more of the scene.\\- Also, the angle changes. In Image 1, we are looking slightly down and from the front-right. In Image 2, we are further back.\\- Let's look at the position of the shoes relative to the frame.\\    - Image 1: Shoes are central/bottom.\\    - Image 2: Shoes are central/bottom.\\- Wait, let's look at the options.\\    - Options involve "backward-left", "forward-left", "forward-right".\\    - "Backward" means moving away from the subject. Image 2 is definitely further away than Image 1. The shoes are much smaller. So "backward" is the primary movement.\\    - Is there a lateral movement?\\    - In Image 1, the black/white sneaker is on the far right. The white sandals are to its left.\\    - In Image 2, the white sandals are in the front center. The black/white sneaker is behind them.\\    - This suggests a significant change in arrangement or camera angle.\\    - Let's look at the background.\\    - Image 1: White headboard behind the shoes.\\    - Image 2: A room with a clothes rack and towels. The headboard is not immediately visible behind the shoes in the same way. Wait, looking at Image 2, the shoes are on a pink surface (Hello Kitty towel?). In Image 1, they are on the same pink surface.\\    - In Image 1, the background is a white padded headboard.\\    - In Image 2, the background is a room with a rack. The headboard is gone or further back? No, the surface they are on seems to be a bed or table.\\    - Actually, looking at Image 2, the shoes are on a pink towel/sheet. Behind them is a chair with towels draped over it.\\    - In Image 1, behind the shoes is a white curved object (headboard?).\\    - This implies the camera moved significantly.\\    - Let's reconsider the "translation".\\    - Image 1 is a close up. Image 2 is a wide shot. This is a "backward" movement (zoom out or physically move back).\\    - Let's look at the options again.\\        - A: backward-left\\        - B: forward-left\\        - C: backward-left\\        - D: forward-right\\        - E: forward-left\\        - F: forward-right\\    - Only A and C have "backward". Since Image 2 is clearly further away (objects are smaller, more context visible), the camera must have moved backward.\\    - So the answer must be A or C.\\    - Both A and C say "backward-left".\\    - Wait, let's check the Max Y coordinate again.\\    - If the answer is A, Max Y is 747.\\    - If the answer is C, Max Y is 856.\\    - I determined earlier that the sandal extending further right is the front one, with Max Y = 856.\\    - Let's re-verify the "further right" condition.\\    - Sandal 1 (flat/left): Box [159, 492, 715, 747]. Max X = 715.\\    - Sandal 2 (upright/right): Box [143, 508, 765, 856]. Max X = 765.\\    - 765 \textgreater{} 715. So Sandal 2 is the one.\\    - Max Y of Sandal 2 is 856.\\    - So the answer should be C.
\par
Let's double check the "backward-left" translation.\\- In Image 1, the camera is close.\\- In Image 2, the camera is further back. So "backward" is correct.\\- Why "left"?\\- In Image 1, the view is centered on the white sandals. The sneaker is cut off on the right.\\- In Image 2, the view includes the sneaker fully behind the sandals. The camera seems to have moved back and perhaps slightly left to center the composition differently? Or maybe the objects were rearranged.\\- Assuming the objects didn't move (which is a big assumption, but usually implied in these "camera translation" questions unless obvious), let's trace the perspective.\\- In Image 1, the white headboard is directly behind.\\- In Image 2, the background is a room. This suggests the "bed" in Image 1 might be different from the surface in Image 2, or the angle changed drastically.\\- Actually, looking closely at Image 2, the pink towel has "Relaxy" and "Love" written on it. In Image 1, we see "Relaxy" and "Love" on the pink towel under the shoes. So it's the same surface.\\- In Image 1, the white object behind is a headboard.\\- In Image 2, the background is different. Wait, looking at the top of Image 2, there are lights on the ceiling. In Image 1, there are lights at the top too.\\- It looks like Image 1 is a crop or a close-up taken on the bed. Image 2 is taken standing up further back.\\- If I move backward from the position in Image 1, I would see more of the room.\\- In Image 1, the sneaker is to the right. In Image 2, the sneaker is behind the sandals. This implies the sandals were moved or the camera angle changed significantly.\\- However, usually these questions assume a rigid scene. If the scene is rigid:\\    - Image 1: Sandals in front, sneaker to the right.\\    - Image 2: Sandals in front, sneaker behind.\\    - This implies the camera moved from a position where the sneaker was to the right, to a position where it is behind.\\    - If the sneaker is to the right in Image 1, and behind in Image 2, the camera must have moved to the *left*.\\    - Imagine the setup: Sandals at (0,0), Sneaker at (1, 0).\\    - Camera 1 (Image 1): Looking from front-right? No, if sneaker is to the right, camera is looking from front-leftish?\\    - Let's assume standard coordinates. X increases right. Y increases down.\\    - Image 1: Sandals center. Sneaker right edge.\\    - Image 2: Sandals center. Sneaker behind.\\    - This implies a rotation or a significant move.\\    - But let's look at the options again. "backward-left".\\    - If I am close (Image 1) and I move backward and left...\\    - Moving backward makes things smaller. Correct.\\    - Moving left shifts the view to the right. Objects on the right would move towards the center.\\    - In Image 1, the sneaker is on the far right.\\    - In Image 2, the sneaker is behind the sandals (more central).\\    - So the sneaker moved from right-edge to center. This happens if the camera pans/moves left.\\    - So "backward-left" fits perfectly.
\par
So, the logic holds:\\1.  Target sandal: The one with larger max x. That's the front/right sandal [143, 508, 765, 856]. Max x is 765. The other is [159, 492, 715, 747], max x 715.\\2.  Max y of target sandal: 856.\\3.  Camera translation: Backward (zoom out/move away) and Left (sneaker moves from right edge to center).
\par
Therefore, Option C: Maximum y-coordinate: 856; Camera translation: backward-left.
\par
Let's double check the bounding box values provided in the prompt's hidden ground truth or just infer them.\\The prompt asks me to determine the box.\\Box 1 (left sandal): Looks like `[159, 492, 715, 747]`.\\   - ymin: 492 (top of strap/heel area)\\   - xmin: 159 (toe area left) -- wait, the toe is left.\\   - ymax: 747 (bottom of heel/sole)\\   - xmax: 715 (right side of heel strap)\\Box 2 (right sandal): Looks like `[143, 508, 765, 856]`.\\   - ymin: 508 (top of strap)\\   - xmin: 143 (toe left)\\   - ymax: 856 (bottom of sole/toe)\\   - xmax: 765 (right side of heel)
\par
Wait, let's look at the image carefully.\\In Image 1:\\- There is a sandal lying somewhat flat/angled left: The toe is pointing towards bottom-left. The heel is towards top-right.\\- There is a sandal standing more upright: The toe is pointing towards bottom-left. The heel is towards top-right.\\Actually, they look like a pair.\\- Left/Back sandal: Toe is near x=160, y=700. Heel is near x=700, y=500.\\- Right/Front sandal: Toe is near x=140, y=850. Heel is near x=760, y=500.\\So the front sandal (Right/Front) has:\\- xmin \textasciitilde{} 143\\- ymin \textasciitilde{} 508 (top of the ankle strap loop)\\- xmax \textasciitilde{} 765 (right edge of the heel/strap)\\- ymax \textasciitilde{} 856 (bottom of the toe/sole)
\par
The back sandal (Left/Back) has:\\- xmin \textasciitilde{} 159\\- ymin \textasciitilde{} 492\\- xmax \textasciitilde{} 715\\- ymax \textasciitilde{} 747
\par
Comparing xmax: 765 (Front) \textgreater{} 715 (Back).\\So the target is the Front sandal.\\Its max y is 856.
\par
Camera movement:\\- Image 1 is close up. Image 2 is far away. -\textgreater{} Backward.\\- In Image 1, the sneaker is visible on the right edge.\\- In Image 2, the sneaker is behind the sandals.\\- If the camera moves left, the view shifts right. The sneaker (which was on the right) would move towards the center. This matches.\\- So "backward-left".
\par
Conclusion: Option C.
\par
\textcolor{purple!75!black}{\texttt{\textless{}/think\textgreater{}}}
\par\medskip
\raggedright
The two white sandals in the first image can be bounded as follows:\\- The sandal further back/left has a bounding box approximately `[159, 492, 715, 747]`. Its maximum x-coordinate is 715.\\- The sandal in the front/right has a bounding box approximately `[143, 508, 765, 856]`. Its maximum x-coordinate is 765.
\par
The sandal with the bounding box extending further to the right is the front one (max x = 765). The maximum y-coordinate of this sandal's bounding box is 856.
\par
Comparing the two images:\\- Image 1 is a close-up shot. Image 2 shows the same objects from a distance, revealing the entire room. This indicates a **backward** camera movement.\\- In Image 1, the black and white sneaker is visible on the far right edge. In Image 2, the sneaker is positioned behind the white sandals, more towards the center. This shift of background/side objects from the right edge towards the center implies the camera moved to the **left** (panning right relative to the scene, or physically moving left).
\par
Thus, the camera translation is backward-left.
\par
\{"answer": "C"\}
\end{casestep}
\switchcolumn
\begin{casestep}{green}{w/ Tools (Correct)}
\textcolor{purple!75!black}{\texttt{\textless{}think\textgreater{}}}\par
\raggedright
The user wants me to perform two tasks based on the provided images.
\par
**Task 1: Analyze the first image.**\\- Locate the "two white sandals with pearl straps".\\- Determine which one has a bounding box extending further to the right (larger max x-coordinate).\\- Find the maximum y-coordinate of that specific sandal's bounding box.
\par
**Task 2: Analyze the camera motion.**\\- Determine the direction of the camera's translation from the first image to the second image.
\par
**Step-by-step Plan:**
\par
1.  **Locate the sandals in Image 1:** I need to find the two white sandals. Looking at Image 1, there are two distinct white heeled sandals sitting on a pink cloth. They have pearl-like embellishments on the toe strap. One is slightly behind/left of the other. Wait, let's look closer.\\    - There is a pair of white strappy heels.\\    - One is more towards the left/back (further away).\\    - One is more towards the front/right (closer).\\    - Actually, looking at the layout: There is a sandal whose heel is visible on the right side and toe pointing left. And another sandal behind it. Let's use `query\_locate` to get precise bounding boxes for "white sandals with pearl straps" or similar. The prompt says "two white sandals with pearl straps". In Image 1, I see two white heels. One is clearly in the foreground, angled slightly right. The other is behind it. Let's get their locations.
\par
2.  **Determine the right-most sandal:** Compare the `bbox\_2d` values. Specifically, look at `x\_max` (the third value in `[x\_min, y\_min, x\_max, y\_max]`).\\    - Sandal A (foreground): likely has a larger x\_max because it's more to the right.\\    - Sandal B (background): likely has a smaller x\_max.
\par
3.  **Find max y-coordinate:** Once the right-most sandal is identified, take its `y\_max` (the fourth value in the bbox).
\par
4.  **Determine camera motion:** Use `query\_pose` with query "camera motion" comparing Image 1 to Image 2.\\    - Image 1 shows a close-up of the shoes on a bed/surface with a headboard behind.\\    - Image 2 shows the same shoes but from a different angle. The background has changed significantly. It looks like we moved back and possibly rotated. Wait, looking at Image 2, the shoes are in the foreground, but the background is a room with clothes hanging. This suggests a significant change in viewpoint or perhaps even a different setup, but the shoes look identical. Let's assume it's the same scene or a very similar setup.\\    - Actually, looking closely at Image 1 vs Image 2:\\        - Image 1: Shoes are on a pink "Hello Kitty" cloth. Behind is a white headboard.\\        - Image 2: Shoes are on the same pink cloth. Behind them are black and white sneakers. Further back is a room with clothes racks.\\        - This implies the camera moved. In Image 1, the camera is close, looking somewhat down/forward at the shoes against a wall/headboard. In Image 2, the camera is further back, showing more of the room. The shoes are facing the camera more directly in Image 2.\\        - Let's use the tool `query\_pose` to get the exact vector.
\par
**Execution:**
\par
*   **Step 1:** `query\_locate` for "white sandals with pearl straps" in image 1.\\*   **Step 2:** `query\_pose` for "camera motion" from image 1 to image 2.
\par
Let's refine the locate query. "white high heel sandals" or "white shoes with pearls".\\In Image 1, there are two white sandals.\\One is at roughly [150, 500, 700, 850] (guessing).\\The other is behind it.\\Let's just call `query\_locate` on "white sandals" for image 1.
\par
Then I will compare the x-coordinates.\\Then I will read the y-max.
\par
Then I will interpret the camera motion.\\Image 1 is a close up. Image 2 is further away. So "backward".\\In Image 1, the shoes are angled. In Image 2, they are facing front.\\Let's rely on the tool for the specific direction (e.g., backward-left, forward-right).
\par
Let's start the tool calls.
\par
\textcolor{purple!75!black}{\texttt{\textless{}/think\textgreater{}}}
\par\medskip
\textcolor{blue!75!black}{\texttt{\textless{}tool\_call\textgreater{}}}\par
{\ttfamily\raggedright\obeyspaces
\{\\\ \ "name":\allowbreak{} "query\_locate",\allowbreak{}\\\ \ "arguments":\allowbreak{} \{\\\ \ \ \ "query":\allowbreak{} "white sandals with pearl straps",\allowbreak{}\\\ \ \ \ "image\_indices":\allowbreak{} [\\\ \ \ \ \ \ 1\\\ \ \ \ ]\allowbreak{}\\\ \ \}\allowbreak{}\\\}\allowbreak{}
}
\par
\textcolor{blue!75!black}{\texttt{\textless{}/tool\_call\textgreater{}}}
\par\medskip
\textcolor{blue!75!black}{\texttt{\textless{}tool\_call\textgreater{}}}\par
{\ttfamily\raggedright\obeyspaces
\{\\\ \ "name":\allowbreak{} "query\_pose",\allowbreak{}\\\ \ "arguments":\allowbreak{} \{\\\ \ \ \ "query":\allowbreak{} "camera motion",\allowbreak{}\\\ \ \ \ "image\_indices":\allowbreak{} [\\\ \ \ \ \ \ 1,\allowbreak{}\\\ \ \ \ \ \ 2\\\ \ \ \ ]\allowbreak{}\\\ \ \}\allowbreak{}\\\}\allowbreak{}
}
\par
\textcolor{blue!75!black}{\texttt{\textless{}/tool\_call\textgreater{}}}
\par\medskip
\textcolor{green!55!black}{\texttt{\textless{}tool\_response\textgreater{}}}\par
{\ttfamily\raggedright\obeyspaces
\{"count":\allowbreak{}1,\allowbreak{}"result":\allowbreak{}[\{"bbox\_2d":\allowbreak{}[143,\allowbreak{}499,\allowbreak{}857,\allowbreak{}857]\allowbreak{},\allowbreak{}"point\_2d":\allowbreak{}[500,\allowbreak{}678]\allowbreak{}\}\allowbreak{}]\allowbreak{}\}\allowbreak{}
\par
[Tool-call budget:\allowbreak{} at most 4 more tool calls may be made.]\allowbreak{}
}
\par
\textcolor{green!55!black}{\texttt{\textless{}/tool\_response\textgreater{}}}
\par\medskip
\textcolor{green!55!black}{\texttt{\textless{}tool\_response\textgreater{}}}\par
{\ttfamily\raggedright\obeyspaces
\{"result":\allowbreak{}[\{"from\_image":\allowbreak{}1,\allowbreak{}"to\_image":\allowbreak{}2,\allowbreak{}"position":\allowbreak{}\{"direction":\allowbreak{}"forward-left",\allowbreak{}"axes":\allowbreak{}["-X",\allowbreak{}"+Y",\allowbreak{}"-Z"]\allowbreak{}\}\allowbreak{},\allowbreak{}"view\_rotation":\allowbreak{}\{"direction":\allowbreak{}"left",\allowbreak{}"axes":\allowbreak{}["+X",\allowbreak{}"-Y"]\allowbreak{},\allowbreak{}"dominant\_axis":\allowbreak{}"-Y"\}\allowbreak{}\}\allowbreak{}]\allowbreak{}\}\allowbreak{}
\par
[Tool-call budget:\allowbreak{} at most 3 more tool calls may be made.]\allowbreak{}
}
\par
\textcolor{green!55!black}{\texttt{\textless{}/tool\_response\textgreater{}}}
\par\medskip
\textcolor{purple!75!black}{\texttt{\textless{}think\textgreater{}}}\par
\raggedright
The query\_locate returned only one bounding box for both sandals combined: [143, 499, 857, 857]. This suggests the tool grouped them. I need to distinguish between the two individual sandals to find which has the larger maximum x-coordinate.
\par
Let me segment each sandal individually to get their separate bounding boxes.
\par
\textcolor{purple!75!black}{\texttt{\textless{}/think\textgreater{}}}
\par\medskip
\textcolor{blue!75!black}{\texttt{\textless{}tool\_call\textgreater{}}}\par
{\ttfamily\raggedright\obeyspaces
\{\\\ \ "name":\allowbreak{} "query\_segment",\allowbreak{}\\\ \ "arguments":\allowbreak{} \{\\\ \ \ \ "query":\allowbreak{} "left white sandal with pearl strap",\allowbreak{}\\\ \ \ \ "image\_indices":\allowbreak{} [\\\ \ \ \ \ \ 1\\\ \ \ \ ]\allowbreak{}\\\ \ \}\allowbreak{}\\\}\allowbreak{}
}
\par
\textcolor{blue!75!black}{\texttt{\textless{}/tool\_call\textgreater{}}}
\par\medskip
\textcolor{blue!75!black}{\texttt{\textless{}tool\_call\textgreater{}}}\par
{\ttfamily\raggedright\obeyspaces
\{\\\ \ "name":\allowbreak{} "query\_segment",\allowbreak{}\\\ \ "arguments":\allowbreak{} \{\\\ \ \ \ "query":\allowbreak{} "right white sandal with pearl strap",\allowbreak{}\\\ \ \ \ "image\_indices":\allowbreak{} [\\\ \ \ \ \ \ 1\\\ \ \ \ ]\allowbreak{}\\\ \ \}\allowbreak{}\\\}\allowbreak{}
}
\par
\textcolor{blue!75!black}{\texttt{\textless{}/tool\_call\textgreater{}}}
\par\medskip
\textcolor{green!55!black}{\texttt{\textless{}tool\_response\textgreater{}}}\par
{\ttfamily\raggedright\obeyspaces
\{"count":\allowbreak{}2,\allowbreak{}"result":\allowbreak{}[\{"bbox\_2d":\allowbreak{}[145,\allowbreak{}507,\allowbreak{}784,\allowbreak{}855]\allowbreak{},\allowbreak{}"point\_2d":\allowbreak{}[464,\allowbreak{}681]\allowbreak{},\allowbreak{}"polygon\_2d":\allowbreak{}[[[755,\allowbreak{}521]\allowbreak{},\allowbreak{}[651,\allowbreak{}525]\allowbreak{},\allowbreak{}[582,\allowbreak{}647]\allowbreak{},\allowbreak{}[471,\allowbreak{}718]\allowbreak{},\allowbreak{}[348,\allowbreak{}732]\allowbreak{},\allowbreak{}[292,\allowbreak{}699]\allowbreak{},\allowbreak{}[279,\allowbreak{}743]\allowbreak{},\allowbreak{}[156,\allowbreak{}771]\allowbreak{},\allowbreak{}[167,\allowbreak{}833]\allowbreak{},\allowbreak{}[428,\allowbreak{}840]\allowbreak{},\allowbreak{}[680,\allowbreak{}697]\allowbreak{},\allowbreak{}[680,\allowbreak{}744]\allowbreak{},\allowbreak{}[726,\allowbreak{}755]\allowbreak{},\allowbreak{}[782,\allowbreak{}630]\allowbreak{},\allowbreak{}[726,\allowbreak{}614]\allowbreak{}]\allowbreak{}]\allowbreak{}\}\allowbreak{},\allowbreak{}\{"bbox\_2d":\allowbreak{}[156,\allowbreak{}493,\allowbreak{}759,\allowbreak{}746]\allowbreak{},\allowbreak{}"point\_2d":\allowbreak{}[458,\allowbreak{}620]\allowbreak{},\allowbreak{}"polygon\_2d":\allowbreak{}[[[713,\allowbreak{}499]\allowbreak{},\allowbreak{}[538,\allowbreak{}539]\allowbreak{},\allowbreak{}[515,\allowbreak{}593]\allowbreak{},\allowbreak{}[544,\allowbreak{}561]\allowbreak{},\allowbreak{}[549,\allowbreak{}596]\allowbreak{},\allowbreak{}[359,\allowbreak{}663]\allowbreak{},\allowbreak{}[302,\allowbreak{}633]\allowbreak{},\allowbreak{}[161,\allowbreak{}735]\allowbreak{},\allowbreak{}[269,\allowbreak{}741]\allowbreak{},\allowbreak{}[292,\allowbreak{}693]\allowbreak{},\allowbreak{}[348,\allowbreak{}727]\allowbreak{},\allowbreak{}[415,\allowbreak{}725]\allowbreak{},\allowbreak{}[498,\allowbreak{}661]\allowbreak{},\allowbreak{}[521,\allowbreak{}679]\allowbreak{},\allowbreak{}[578,\allowbreak{}644]\allowbreak{},\allowbreak{}[588,\allowbreak{}574]\allowbreak{}]\allowbreak{},\allowbreak{}[[703,\allowbreak{}574]\allowbreak{},\allowbreak{}[697,\allowbreak{}569]\allowbreak{},\allowbreak{}[663,\allowbreak{}571]\allowbreak{},\allowbreak{}[647,\allowbreak{}574]\allowbreak{},\allowbreak{}[636,\allowbreak{}585]\allowbreak{},\allowbreak{}[628,\allowbreak{}607]\allowbreak{},\allowbreak{}[622,\allowbreak{}610]\allowbreak{},\allowbreak{}[626,\allowbreak{}619]\allowbreak{},\allowbreak{}[634,\allowbreak{}622]\allowbreak{},\allowbreak{}[667,\allowbreak{}607]\allowbreak{},\allowbreak{}[682,\allowbreak{}589]\allowbreak{},\allowbreak{}[699,\allowbreak{}582]\allowbreak{}]\allowbreak{},\allowbreak{}[[734,\allowbreak{}610]\allowbreak{},\allowbreak{}[734,\allowbreak{}613]\allowbreak{},\allowbreak{}[751,\allowbreak{}613]\allowbreak{},\allowbreak{}[753,\allowbreak{}610]\allowbreak{},\allowbreak{}[751,\allowbreak{}608]\allowbreak{},\allowbreak{}[745,\allowbreak{}608]\allowbreak{},\allowbreak{}[740,\allowbreak{}607]\allowbreak{}]\allowbreak{}]\allowbreak{}\}\allowbreak{}]\allowbreak{}\}\allowbreak{}
\par
[Tool-call budget:\allowbreak{} at most 2 more tool calls may be made.]\allowbreak{}
}
\par
\textcolor{green!55!black}{\texttt{\textless{}/tool\_response\textgreater{}}}
\par\medskip
\textcolor{green!55!black}{\texttt{\textless{}tool\_response\textgreater{}}}\par
{\ttfamily\raggedright\obeyspaces
\{"count":\allowbreak{}0,\allowbreak{}"result":\allowbreak{}[]\allowbreak{}\}\allowbreak{}
\par
[Tool-call budget:\allowbreak{} at most 1 more tool call may be made.]\allowbreak{}
}
\par
\textcolor{green!55!black}{\texttt{\textless{}/tool\_response\textgreater{}}}
\par\medskip
\textcolor{purple!75!black}{\texttt{\textless{}think\textgreater{}}}\par
\raggedright
The query\_locate returned only one bounding box [143,499,857,857] for "white sandals with pearl straps", which likely encompasses both sandals together. The query\_segment returned two instances, but looking at the bboxes:
\par
First instance: bbox\_2d=[145,507,784,855], so max\_x=784, max\_y=855\\Second instance: bbox\_2d=[156,493,759,746], so max\_x=759, max\_y=746
\par
Wait, let me re-read the segment results more carefully:
\par
Result 1: bbox\_2d=[145,507,784,855], point\_2d=[464,681]\\Result 2: bbox\_2d=[156,493,759,746], point\_2d=[458,620]
\par
So for Result 1: min\_x=145, min\_y=507, max\_x=784, max\_y=855\\For Result 2: min\_x=156, min\_y=493, max\_x=759, max\_y=746
\par
The sandal with the larger maximum x-coordinate is Result 1 (max\_x=784 vs 759).\\The maximum y-coordinate of that sandal (Result 1) is 855.
\par
But wait, the options mention 747 or 856. Let me check if there's a rounding or off-by-one issue. The bbox format might be [x\_min, y\_min, x\_max, y\_max] or the values might need adjustment.
\par
Looking at the original query\_locate result: bbox\_2d=[143,499,857,857], which has max\_y=857. But that was for both sandals combined.
\par
From the segment results:\\- First sandal: max\_y=855\\- Second sandal: max\_y=746
\par
The first sandal has max\_x=784, second has max\_x=759. So the first sandal extends further right.\\Its max\_y is 855, which is closest to 856 in the options.
\par
Camera motion from image 1 to image 2: "forward-left" according to query\_pose.
\par
So the answer should be: max\_y=856 (approximately), camera translation=forward-left.
\par
That matches option E.
\par
\textcolor{purple!75!black}{\texttt{\textless{}/think\textgreater{}}}
\par\medskip
\raggedright
\{"answer": "E"\}
\end{casestep}
\end{paracol}

% AUTO-GENERATED by scripts/generate_case_study.py. DO NOT EDIT.
\subsection{Case 3: SpatialCLI-8B after Capability Internalization}

The task input and ground truth are shown in Box H.3. The full-width traces below present, in order, the tool-free Qwen3-VL-8B-Instruct baseline, SpatialCLI-8B with tools, and SpatialCLI-8B without tools. The baseline correctly identifies the yellow ball as farther right and the camera translation as forward-left, but confuses the first-image depth relation while repeatedly reasoning from cross-image scale changes; it therefore predicts that the yellow ball is closer and selects F. SpatialCLI-8B with tools first grounds the objects and then queries depth and camera motion. The returned horizontal centers are 329 and 644, the depths are 0.351 m and 0.637 m, and the translation is forward-left, so every clause supports E. Its final cross-image Locate call is a redundant verification after the answer is already determined and does not change the conclusion. Without tools, SpatialCLI-8B produces internalized estimates of 0.362 m and 0.615 m and the same forward-left motion, yielding E directly. This example qualitatively illustrates agreement between internalized direct answering and runtime tool verification on the same task.

\begin{casestep}{gray}{Box H.3: Task (SpatialCLI-Bench ID 176)}
\begin{center}
\begin{minipage}[t]{0.48\linewidth}
\centering
\textbf{Image 1}\par\smallskip
\includegraphics[width=\linewidth]{Figures/case_study_spatialcli_8b_id176_image1.jpg}
\end{minipage}\hfill
\begin{minipage}[t]{0.48\linewidth}
\centering
\textbf{Image 2}\par\smallskip
\includegraphics[width=\linewidth]{Figures/case_study_spatialcli_8b_id176_image2.jpg}
\end{minipage}
\end{center}
\medskip
\raggedright
In the first image, which object is located further to the right between the clear plastic bottle and the yellow ball, which of the two is closer to the camera, and in which direction did the camera translate to capture the second image?\\Options:\\A: The yellow ball is further to the right, the clear plastic bottle is closer to the camera, and the camera translated backward-left.\\B: The clear plastic bottle is further to the right, the clear plastic bottle is closer to the camera, and the camera translated forward-left.\\C: The clear plastic bottle is further to the right, the yellow ball is closer to the camera, and the camera translated forward-right.\\D: The yellow ball is further to the right, the clear plastic bottle is closer to the camera, and the camera translated forward-right.\\E: The yellow ball is further to the right, the clear plastic bottle is closer to the camera, and the camera translated forward-left.\\F: The yellow ball is further to the right, the yellow ball is closer to the camera, and the camera translated forward-left.\\Before choosing, briefly state the relevant facts and verify every clause of the selected option against all evidence available during solving. For any depth observations, smaller values mean closer; do not retain an earlier visual guess after receiving stronger evidence. Please show your choice in the answer field with only the choice letter, e.g., \{"answer": "C"\}.
\end{casestep}

\begin{casestep}{gray}{Ground Truth}
Correct option: \textbf{E}.
\end{casestep}

\begin{casestep}{red}{Qwen3-VL-8B-Instruct w/o Tools (Incorrect)}
\raggedright
To determine the correct option, I will analyze the spatial relationships and camera movement between the two images.
\par
**Step 1: Analyze relative positions in the first image.**\\In the first image, the clear plastic bottle is on the left side of the frame, and the yellow ball is to its right. The bottle is positioned closer to the foreground (and thus the camera) than the ball, as it appears larger and more detailed. Therefore:\\- The yellow ball is further to the right.\\- The clear plastic bottle is closer to the camera.
\par
**Step 2: Analyze camera translation from the first to the second image.**\\In the first image, the bottle is prominent in the foreground. In the second image, the bottle is now on the far right edge of the frame and is much smaller, indicating it has been moved farther away from the camera's viewpoint. The yellow ball, which was to the right of the bottle in the first image, is now more centrally located and appears larger than before, suggesting it has also moved closer to the camera and/or the camera has moved toward it.
\par
The key observation for camera movement is that the bottle has moved from a dominant foreground position to a background position on the right. This implies the camera has moved backward (away from the bottle) and to the left (to reposition the bottle to the right edge of the frame). The yellow ball, being to the right of the bottle in the first image, would then appear to have moved to the left relative to the bottle's new position, but since the camera moved left, the ball appears to shift to the right relative to the scene. However, the ball is now more central and larger, so the camera must have translated forward (toward the ball) and to the left (to bring the ball into the center).
\par
Wait -- let me reconsider this carefully.
\par
Actually, looking at the second image, the yellow ball is now more central, and the bottle is on the far right. The bottle is smaller in the second image, meaning the camera is farther from it. The ball is larger in the second image, meaning the camera is closer to it. The ball was to the right of the bottle in the first image, and now it is more central. So, the camera has moved to the left (so the bottle moves to the right edge) and forward (so the ball becomes larger). Therefore, the camera translated forward-left.
\par
This matches option E: "The yellow ball is further to the right, the clear plastic bottle is closer to the camera, and the camera translated forward-left."
\par
Wait -- I need to verify the "further to the right" clause. In the first image, the bottle is on the left, the ball is on the right. So the ball is further to the right. That part is correct.
\par
The bottle is closer to the camera -- correct.
\par
Camera translated forward-left -- correct, because the bottle moves to the right edge (camera moved left) and gets smaller (camera moved back), but the ball gets larger (camera moved forward). Wait, if the camera moves forward, the ball should get larger, and if it moves left, the bottle should move to the right edge. But in the second image, the bottle is on the right edge and smaller, and the ball is larger and more central. So yes, the camera moved forward and to the left.
\par
Therefore, option E is correct.
\par
Let me check other options:
\par
A: Says camera translated backward-left -- but if camera moved backward, the bottle should appear even smaller, which it does, but the ball should appear smaller too, which it doesn't. Also, if camera moved backward-left, the ball would be moving to the left, not becoming more central. So A is wrong.
\par
B: Says clear plastic bottle is further to the right -- no, in the first image, the bottle is on the left.
\par
C: Says clear plastic bottle is further to the right -- again, no.
\par
D: Says camera translated forward-right -- if camera moved forward-right, the bottle would be more to the right, which it is, but the ball would be to the left, which it isn't; it's more central. Also, if camera moved forward-right, the bottle should appear smaller, which it does, but the ball should appear smaller or stay same size, not larger. So D is wrong.
\par
F: Says yellow ball is closer to the camera -- yes, it appears larger -- but the camera translated forward-left, not forward-left with the ball being closer? No, F says "the yellow ball is closer to the camera" which is true, but then says "the camera translated forward-left" -- which is also true, but it doesn't specify the bottle's position correctly? Wait, F says "the yellow ball is further to the right" -- which is true, and "the yellow ball is closer to the camera" -- which is true, and "the camera translated forward-left" -- which is true. But F doesn't say anything about the bottle being closer -- which is false. The bottle is not closer, it's farther. So F is incorrect because it incorrectly states that the yellow ball is closer to the camera? No, it says "the yellow ball is closer to the camera" -- which is true -- but then it doesn't contradict the bottle being farther. Wait, F says: "The yellow ball is further to the right, the yellow ball is closer to the camera, and the camera translated forward-left." This is factually correct based on the evidence. But wait -- the question asks for the object that is further to the right and which is closer to the camera. The bottle is not closer, the ball is. So F is partially correct, but it doesn't address the bottle's position correctly? No, F doesn't say the bottle is closer, it only says the yellow ball is closer. But in the question, we are asked to identify which of the two is closer. The answer must include both parts. Option F omits the bottle's position, but the question requires us to state which of the two is closer. Since F says "the yellow ball is closer to the camera," that's correct, and the other clauses are also correct. But let's compare F with E.
\par
E says: "The yellow ball is further to the right, the clear plastic bottle is closer to the camera, and the camera translated forward-left."
\par
But in reality, the clear plastic bottle is NOT closer to the camera in the second image -- it is farther away. So E is incorrect.
\par
F says: "The yellow ball is further to the right, the yellow ball is closer to the camera, and the camera translated forward-left."
\par
This is correct. The yellow ball is further to the right (true), the yellow ball is closer to the camera (true), and the camera translated forward-left (true).
\par
But wait -- the question asks: "which of the two is closer to the camera". It doesn't specify which one to pick, but in context, it's asking for the correct statement. Both E and F have the same first two clauses? No, E says the bottle is closer, F says the ball is closer.
\par
Since the bottle is farther, F is correct.
\par
But let's double-check the camera movement.
\par
In the first image, the bottle is large and in the foreground. In the second image, the bottle is small and on the right edge. So the camera moved backward (away from the bottle) and to the left (so the bottle moves to the right edge). The ball is larger and more central, so the camera moved forward (toward the ball) and to the left (so the ball moves to the center).
\par
So the camera moved forward and to the left.
\par
Now, the bottle is farther, not closer.
\par
So E is wrong because it says the bottle is closer.
\par
F is correct: the yellow ball is further to the right, the yellow ball is closer to the camera, and the camera translated forward-left.
\par
Is there any option that says the yellow ball is further to the right, the yellow ball is closer to the camera, and the camera translated forward-left?
\par
Yes, that's F.
\par
But let's look at the options again.
\par
F: The yellow ball is further to the right, the yellow ball is closer to the camera, and the camera translated forward-left.
\par
This is correct.
\par
But is there a mistake? Let me see if the bottle is indeed farther.
\par
In the first image, the bottle is very close to the camera. In the second image, the bottle is much smaller and on the right edge, so it's farther away. Yes.
\par
So F is correct.
\par
But earlier I thought E was correct, but that was wrong because it said the bottle is closer.
\par
So F is the correct choice.
\par
\{"answer": "F"\}
\end{casestep}

\begin{casestep}{green}{SpatialCLI-8B w/ Tools (Correct)}
\textcolor{purple!75!black}{\texttt{\textless{}think\textgreater{}}}\par
\raggedright
The user wants me to analyze two images and answer three specific questions:\\1.  **Which object is further to the right in the first image?** (Clear plastic bottle vs. Yellow ball)\\2.  **Which object is closer to the camera in the first image?** (Clear plastic bottle vs. Yellow ball)\\3.  **In which direction did the camera translate to capture the second image?**
\par
Let's break this down step-by-step.
\par
**Step 1: Analyze the first image for horizontal position (left/right).**\\- I need to locate the "clear plastic bottle" and the "yellow ball".\\- Looking at Image 1, the clear plastic bottle is on the left side, very prominent in the foreground.\\- The yellow ball is further back and towards the right side of the image relative to the bottle.\\- Let's use `query\_locate` to get precise coordinates if needed, but visually it seems quite clear. The bottle is central-left, the ball is right-ish. Wait, let's look closer.\\- In Image 1:\\    - Bottle: Occupies the left/center foreground.\\    - Ball: Located to the right of the bottle.\\    - Therefore, the yellow ball is further to the right.
\par
**Step 2: Analyze the first image for depth (closer/further).**\\- I need to determine which is closer: the bottle or the ball.\\- Visually, the bottle is much larger and obscures the background. It is clearly in the immediate foreground.\\- The ball is smaller and sits further back on the patterned mat.\\- So, the clear plastic bottle is closer to the camera.\\- To be rigorous, I can use `query\_depth`. I will locate both objects first to get their centers, then query depth.
\par
**Step 3: Analyze camera motion between Image 1 and Image 2.**\\- I need to determine the translation direction.\\- Image 1 shows the bottle prominently on the left.\\- Image 2 shows the bottle cut off on the far right edge (partially visible) and a toothbrush on a cloth on the left. The yellow ball is now more central/left.\\- Let's look at the background features.\\    - Image 1: Person standing in the distance center-right. White cabinet/shelf unit on the right.\\    - Image 2: Glass door/window structure visible. Red box on the left. The person is still visible in the background but perspective has changed.\\- Let's use `query\_pose` with "camera motion" to get the exact translation vector/direction.
\par
**Plan:**\\1.  Locate "clear plastic bottle" and "yellow ball" in Image 1 to confirm positions and get points for depth.\\2.  Query depth for the centers of these two objects in Image 1 to confirm which is closer.\\3.  Query "camera motion" between Image 1 and Image 2.
\par
**Execution:**\\1.  `query\_locate` for "clear plastic bottle" and "yellow ball" in image 1.\\2.  `query\_depth` using the points obtained.\\3.  `query\_pose` for "camera motion".
\par
Let's refine the queries.\\- Query 1: `query\_locate` with query="clear plastic bottle", image\_indices=[1]\\- Query 2: `query\_locate` with query="yellow ball", image\_indices=[1]\\- Query 3: `query\_depth` using points from above.\\- Query 4: `query\_pose` with query="camera motion", image\_indices=[1, 2] (Wait, the tool description says "Omit image\_indices for image 2 relative to image 1". Since there are two images, I should probably just pass the query. Or maybe specify indices if needed. The description says "In a multi-image... omit image\_indices to use all images". So I'll omit it.)
\par
Let's start.
\par
\textcolor{purple!75!black}{\texttt{\textless{}/think\textgreater{}}}
\par\medskip
\textcolor{blue!75!black}{\texttt{\textless{}tool\_call\textgreater{}}}\par
{\ttfamily\raggedright\obeyspaces
\{\\\ \ "name":\allowbreak{} "query\_locate",\allowbreak{}\\\ \ "arguments":\allowbreak{} \{\\\ \ \ \ "query":\allowbreak{} "clear plastic bottle",\allowbreak{}\\\ \ \ \ "image\_indices":\allowbreak{} [\\\ \ \ \ \ \ 1\\\ \ \ \ ]\allowbreak{}\\\ \ \}\allowbreak{}\\\}\allowbreak{}
}
\par
\textcolor{blue!75!black}{\texttt{\textless{}/tool\_call\textgreater{}}}
\par\medskip
\textcolor{blue!75!black}{\texttt{\textless{}tool\_call\textgreater{}}}\par
{\ttfamily\raggedright\obeyspaces
\{\\\ \ "name":\allowbreak{} "query\_locate",\allowbreak{}\\\ \ "arguments":\allowbreak{} \{\\\ \ \ \ "query":\allowbreak{} "yellow ball",\allowbreak{}\\\ \ \ \ "image\_indices":\allowbreak{} [\\\ \ \ \ \ \ 1\\\ \ \ \ ]\allowbreak{}\\\ \ \}\allowbreak{}\\\}\allowbreak{}
}
\par
\textcolor{blue!75!black}{\texttt{\textless{}/tool\_call\textgreater{}}}
\par\medskip
\textcolor{green!55!black}{\texttt{\textless{}tool\_response\textgreater{}}}\par
{\ttfamily\raggedright\obeyspaces
\{"count":\allowbreak{}1,\allowbreak{}"result":\allowbreak{}[\{"bbox\_2d":\allowbreak{}[127,\allowbreak{}0,\allowbreak{}531,\allowbreak{}834]\allowbreak{},\allowbreak{}"point\_2d":\allowbreak{}[329,\allowbreak{}417]\allowbreak{}\}\allowbreak{}]\allowbreak{}\}\allowbreak{}
}
\par
\textcolor{green!55!black}{\texttt{\textless{}/tool\_response\textgreater{}}}
\par\medskip
\textcolor{green!55!black}{\texttt{\textless{}tool\_response\textgreater{}}}\par
{\ttfamily\raggedright\obeyspaces
\{"count":\allowbreak{}1,\allowbreak{}"result":\allowbreak{}[\{"bbox\_2d":\allowbreak{}[560,\allowbreak{}283,\allowbreak{}729,\allowbreak{}414]\allowbreak{},\allowbreak{}"point\_2d":\allowbreak{}[644,\allowbreak{}348]\allowbreak{}\}\allowbreak{}]\allowbreak{}\}\allowbreak{}
}
\par
\textcolor{green!55!black}{\texttt{\textless{}/tool\_response\textgreater{}}}
\par\medskip
\textcolor{purple!75!black}{\texttt{\textless{}think\textgreater{}}}\par
\raggedright
The clear plastic bottle center is at [329, 417] and the yellow ball center is at [644, 348]. Comparing x-coordinates: 644 \textgreater{} 329, so the yellow ball is further to the right.
\par
Now I need to determine which object is closer to the camera in image 1. Let me query the depth of both points.
\par
\textcolor{purple!75!black}{\texttt{\textless{}/think\textgreater{}}}
\par\medskip
\textcolor{blue!75!black}{\texttt{\textless{}tool\_call\textgreater{}}}\par
{\ttfamily\raggedright\obeyspaces
\{\\\ \ "name":\allowbreak{} "query\_depth",\allowbreak{}\\\ \ "arguments":\allowbreak{} \{\\\ \ \ \ "points":\allowbreak{} [\\\ \ \ \ \ \ [\\\ \ \ \ \ \ \ \ 329,\allowbreak{}\\\ \ \ \ \ \ \ \ 417\\\ \ \ \ \ \ ]\allowbreak{},\allowbreak{}\\\ \ \ \ \ \ [\\\ \ \ \ \ \ \ \ 644,\allowbreak{}\\\ \ \ \ \ \ \ \ 348\\\ \ \ \ \ \ ]\allowbreak{}\\\ \ \ \ ]\allowbreak{},\allowbreak{}\\\ \ \ \ "image\_indices":\allowbreak{} [\\\ \ \ \ \ \ 1\\\ \ \ \ ]\allowbreak{}\\\ \ \}\allowbreak{}\\\}\allowbreak{}
}
\par
\textcolor{blue!75!black}{\texttt{\textless{}/tool\_call\textgreater{}}}
\par\medskip
\textcolor{green!55!black}{\texttt{\textless{}tool\_response\textgreater{}}}\par
{\ttfamily\raggedright\obeyspaces
\{"result":\allowbreak{}[\{"point\_2d":\allowbreak{}[329.0,\allowbreak{}417.0]\allowbreak{},\allowbreak{}"depth\_m":\allowbreak{}0.351\}\allowbreak{},\allowbreak{}\{"point\_2d":\allowbreak{}[644.0,\allowbreak{}348.0]\allowbreak{},\allowbreak{}"depth\_m":\allowbreak{}0.637\}\allowbreak{}]\allowbreak{}\}\allowbreak{}
}
\par
\textcolor{green!55!black}{\texttt{\textless{}/tool\_response\textgreater{}}}
\par\medskip
\textcolor{purple!75!black}{\texttt{\textless{}think\textgreater{}}}\par
\raggedright
From the tool calls:
\par
- In image 1:\\    - The clear plastic bottle center is at [329, 417].\\    - The yellow ball center is at [644, 348].\\    - Comparing x-coordinates: 644 \textgreater{} 329, so the **yellow ball is further to the right**.\\    - Depth values (smaller = closer): bottle depth\_m = 0.351, ball depth\_m = 0.637. So the **clear plastic bottle is closer to the camera**.
\par
- Camera motion from image 1 to image 2:\\    - We need to determine the translation direction.\\    - Looking at the images visually:\\        - Image 1 shows the bottle very large in the foreground left, and the ball mid-ground right.\\        - Image 2 shows the bottle mostly out of frame or just the edge on the far right (actually, looking closely at image 2, the bottle is barely visible on the right edge), and the ball is now more central-left. The red box which was not clearly visible or was further back/left is now visible on the left.\\        - Actually, let's re-examine the scene layout.\\        - Image 1: Bottle is close, bottom-left/center. Ball is behind it, to the right.\\        - Image 2: The view has shifted. The ball is now in the upper-left quadrant relative to the new frame. A red object (looks like a toy bus) is visible on the far left. The white cloth with green dots is in the foreground left.\\        - Let's look at the background floor/wall features.\\        - In image 1, there is a person standing in the background hallway.\\        - In image 2, the same person is visible but lower in the frame, suggesting the camera moved forward and tilted down or moved position such that the perspective changed.\\        - Wait, let's look at the options provided. They mention "forward-left", "backward-left", "forward-right".\\        - Let's use the `query\_pose` tool for camera motion between image 1 and image 2.
\par
Let me call query\_pose with 'camera motion' for images 1 and 2.
\par
\textcolor{purple!75!black}{\texttt{\textless{}/think\textgreater{}}}
\par\medskip
\textcolor{blue!75!black}{\texttt{\textless{}tool\_call\textgreater{}}}\par
{\ttfamily\raggedright\obeyspaces
\{\\\ \ "name":\allowbreak{} "query\_pose",\allowbreak{}\\\ \ "arguments":\allowbreak{} \{\\\ \ \ \ "query":\allowbreak{} "camera motion",\allowbreak{}\\\ \ \ \ "image\_indices":\allowbreak{} [\\\ \ \ \ \ \ 1,\allowbreak{}\\\ \ \ \ \ \ 2\\\ \ \ \ ]\allowbreak{}\\\ \ \}\allowbreak{}\\\}\allowbreak{}
}
\par
\textcolor{blue!75!black}{\texttt{\textless{}/tool\_call\textgreater{}}}
\par\medskip
\textcolor{green!55!black}{\texttt{\textless{}tool\_response\textgreater{}}}\par
{\ttfamily\raggedright\obeyspaces
\{"result":\allowbreak{}[\{"from\_image":\allowbreak{}1,\allowbreak{}"to\_image":\allowbreak{}2,\allowbreak{}"position":\allowbreak{}\{"direction":\allowbreak{}"forward-left",\allowbreak{}"axes":\allowbreak{}["-X",\allowbreak{}"+Y",\allowbreak{}"-Z"]\allowbreak{}\}\allowbreak{},\allowbreak{}"view\_rotation":\allowbreak{}\{"direction":\allowbreak{}"left",\allowbreak{}"axes":\allowbreak{}["+X",\allowbreak{}"-Y",\allowbreak{}"-Z"]\allowbreak{},\allowbreak{}"dominant\_axis":\allowbreak{}"-Y"\}\allowbreak{}\}\allowbreak{}]\allowbreak{}\}\allowbreak{}
}
\par
\textcolor{green!55!black}{\texttt{\textless{}/tool\_response\textgreater{}}}
\par\medskip
\textcolor{purple!75!black}{\texttt{\textless{}think\textgreater{}}}\par
\raggedright
Let me verify the facts:
\par
1. **Which object is further to the right?**\\   - Clear plastic bottle center: [329, 417]\\   - Yellow ball center: [644, 348]\\   - The x-coordinate of the yellow ball (644) is greater than the x-coordinate of the bottle (329), so the yellow ball is further to the right.
\par
2. **Which object is closer to the camera?**\\   - Bottle depth: 0.351 m\\   - Ball depth: 0.637 m\\   - Smaller depth means closer, so the clear plastic bottle (0.351 m) is closer to the camera than the yellow ball (0.637 m).
\par
3. **Camera translation direction:**\\   - The query\_pose result shows position.direction = "forward-left" when going from image 1 to image 2.
\par
So the correct option should state:\\- The yellow ball is further to the right $\checkmark$\\- The clear plastic bottle is closer to the camera $\checkmark$\\- Camera translated forward-left $\checkmark$
\par
This matches option E.
\par
\textcolor{purple!75!black}{\texttt{\textless{}/think\textgreater{}}}
\par\medskip
\textcolor{blue!75!black}{\texttt{\textless{}tool\_call\textgreater{}}}\par
{\ttfamily\raggedright\obeyspaces
\{\\\ \ "name":\allowbreak{} "query\_locate",\allowbreak{}\\\ \ "arguments":\allowbreak{} \{\\\ \ \ \ "query":\allowbreak{} "clear plastic bottle"\\\ \ \}\allowbreak{}\\\}\allowbreak{}
}
\par
\textcolor{blue!75!black}{\texttt{\textless{}/tool\_call\textgreater{}}}
\par\medskip
\textcolor{green!55!black}{\texttt{\textless{}tool\_response\textgreater{}}}\par
{\ttfamily\raggedright\obeyspaces
\{"count":\allowbreak{}2,\allowbreak{}"result":\allowbreak{}[\{"image\_index":\allowbreak{}1,\allowbreak{}"bbox\_2d":\allowbreak{}[127,\allowbreak{}0,\allowbreak{}531,\allowbreak{}834]\allowbreak{},\allowbreak{}"point\_2d":\allowbreak{}[329,\allowbreak{}417]\allowbreak{}\}\allowbreak{},\allowbreak{}\{"image\_index":\allowbreak{}2,\allowbreak{}"bbox\_2d":\allowbreak{}[946,\allowbreak{}249,\allowbreak{}998,\allowbreak{}519]\allowbreak{},\allowbreak{}"point\_2d":\allowbreak{}[972,\allowbreak{}384]\allowbreak{}\}\allowbreak{}]\allowbreak{}\}\allowbreak{}
}
\par
\textcolor{green!55!black}{\texttt{\textless{}/tool\_response\textgreater{}}}
\par\medskip
\textcolor{purple!75!black}{\texttt{\textless{}think\textgreater{}}}\par
\raggedright
In the first image:\\- The clear plastic bottle is located at point [329, 417] and the yellow ball is at point [644, 348]. Since x=644 \textgreater{} x=329, the yellow ball is further to the right.\\- Depth values: bottle depth\_m = 0.351, ball depth\_m = 0.637. Smaller depth means closer, so the bottle (0.351) is closer to the camera than the ball (0.637).\\- Camera motion from image 1 to image 2 is "forward-left" according to query\_pose result.
\par
Now verify each clause of option E:\\- "The yellow ball is further to the right": True (644 \textgreater{} 329 in x-coordinate).\\- "the clear plastic bottle is closer to the camera": True (0.351 \textless{} 0.637).\\- "and the camera translated forward-left": True (query\_pose confirms this direction).
\par
All clauses match evidence.
\par
\textcolor{purple!75!black}{\texttt{\textless{}/think\textgreater{}}}
\par\medskip
\raggedright
\{"answer": "E"\}
\end{casestep}

\begin{casestep}{cyan}{SpatialCLI-8B w/o Tools (Correct)}
\textcolor{purple!75!black}{\texttt{\textless{}think\textgreater{}}}\par
\raggedright
The clear plastic bottle is located at horizontal center x=331, while the yellow ball is at horizontal center x=644. Since 644 \textgreater{} 331, the yellow ball is positioned further to the right in the first image. Depth measurements show the bottle is at 0.362 meters from the camera and the yellow ball is at 0.615 meters; since a smaller depth value indicates closer proximity, the clear plastic bottle is closer to the camera than the yellow ball. Finally, analysis of the camera motion between the two images determines that the camera translated forward-left to capture the second image. These facts confirm that the yellow ball is further to the right, the clear plastic bottle is closer to the camera, and the camera moved forward-left.
\par
\textcolor{purple!75!black}{\texttt{\textless{}/think\textgreater{}}}
\par\medskip
\raggedright
\{"answer": "E"\}
\end{casestep}
% END GENERATED CASE STUDIES

\end{document}